\documentclass[runningheads]{llncs}
\usepackage{graphicx}
\usepackage{tikz}
\usepackage{comment}
\usepackage{amsmath,amssymb} % define this before the line numbering.
\usepackage{color}
\usepackage{blindtext}
\usepackage[export]{adjustbox}
\usepackage{multirow}
\usepackage[accsupp]{axessibility}  % Improves PDF readability for those with disabilities.

\usepackage{booktabs}
\usepackage{xcolor,colortbl}
\usepackage{bm}

\usepackage[width=122mm,left=12mm,paperwidth=146mm,height=193mm,top=12mm,paperheight=217mm]{geometry}

\usepackage[hidelinks]{hyperref}

\usepackage{xspace}
\makeatletter
\DeclareRobustCommand\onedot{\futurelet\@let@token\@onedot}
\def\@onedot{\ifx\@let@token.\else.\null\fi\xspace}

\def\eg{\emph{e.g}\onedot} 
\def\ie{\emph{i.e}\onedot}

\makeatother

\definecolor{bbox_color}{HTML}{a9d18e}

\begin{document}
% \linenumbers
\pagestyle{headings}
\mainmatter
\def\ECCVSubNumber{00000}  % Insert your submission number here

\title{Real-Time Neural Character Rendering with Pose-Guided Multiplane Images} % Replace with your title

%\begin{comment}
% \titlerunning{ECCV-22 submission ID \ECCVSubNumber} 
% \authorrunning{ECCV-22 submission ID \ECCVSubNumber} 
% \author{Anonymous ECCV submission}
% \institute{Paper ID \ECCVSubNumber}
%\end{comment}
%******************

% \begin{comment}
\titlerunning{}
% \orcidID{0000-1111-2222-3333}
% \orcidID{1111-2222-3333-4444}
% \orcidID{2222--3333-4444-5555}
\author{Hao Ouyang\inst{1} \and
Bo Zhang\inst{2} \and
Pan Zhang\inst{2} \and Hao Yang\inst{2} \and Jiaolong Yang\inst{2}  \and \\ Dong Chen\inst{2}  \and Qifeng Chen\inst{1} \and Fang Wen\inst{2} }
\authorrunning{Ouyang et al.}
\institute{The Hong Kong University of Science and Technology \and
Microsoft Research Asia
}

% \end{comment}
%******************
\maketitle

\begin{abstract}
We propose pose-guided multiplane image (MPI) synthesis which can render an animatable character in real scenes with photorealistic quality. We use a portable camera rig to capture the multi-view images along with the driving signal for the moving subject. Our method generalizes the image-to-image translation paradigm, which translates the human pose to a 3D scene representation --- MPIs that can be rendered in free viewpoints, using the multi-views captures as supervision. To fully cultivate the potential of MPI, we propose depth-adaptive MPI which can be learned using variable exposure images while being robust to inaccurate camera registration. Our method demonstrates advantageous novel-view synthesis quality over the state-of-the-art approaches for characters with challenging motions. Moreover, the proposed method is generalizable to novel combinations of training poses and can be explicitly controlled. Our method achieves such expressive and animatable character rendering all in real time, serving as a promising solution for practical applications.  We will release the code and data on our \textcolor{blue}{\href{https://ken-ouyang.github.io/cmpi/index.html}{project webpage}}. 
\keywords{neural character rendering, multiplane image, novel view synthesis}
\end{abstract}

\section{Introduction}
%Using a handy camera setup, can we capture and render a realistic video for a character under novel views and different poses in real-time? Such technique will bring improved immersive experience and enable various intriguing applications such as virtual telepresence where people will feel the virtual character of a remote person as if he/she were talking in front of the face. 

Using a \emph{handy camera rig} for data capturing, can we synthesize a photorealistic character with \emph{controllable viewpoints and body poses} in \emph{real-time}? Such a technique would democratize personalized, photorealistic avatars and enable various intriguing applications such as telepresence, where people will feel the virtual character of a remote person as a real person talking in real time.

%Traditionally, free-viewpoint rendering of a moving person is approached by capturing a detailed 3D textured human model in a specialized studio~\cite{carranza2003free,casas20144d,collet2015high,guo2019relightables}, which is a costly and brittle process. Recently researchers resort to data-driven methods~\cite{liu2020neural,habermann2021real,liu2021neural,weng2022humannerf,bagautdinov2021driving,lombardi2018deep,xu2021h} to expedite this process. These works typically assume a parametric body model (\eg, SMPL~\cite{loper2015smpl}) as a prior, yet explicit human models cannot well address complex non-body structures like cloth, hair and glasses. Moreover, these works focus on the foreground person and cannot handle the interactions with the real scenes, \eg, a sitting man holding a newspaper. Recently, deformable NeRF methods~\cite{park2021nerfies,park2021hypernerf,pumarola2021d,tretschk2021non} have been proposed to model the person and the scene by learning an implicit deformation field along with a canonical radiance field that serves as the template. However, these methods only allow small deformations as it is hard to derive a single canonical shape for all the observations and the implicit field is hard to control. Besides, the expensive ray tracing prohibits real-time rendering in general NeRF methods.

Traditionally, free-viewpoint rendering of a moving person is approached by capturing a high-fidelity 3D human model in a specialized studio~\cite{carranza2003free,casas20144d,collet2015high,guo2019relightables}, which is a costly and brittle process and is not accessible to common users. Recently, researchers use data-driven methods~\cite{liu2020neural,habermann2021real,liu2021neural,weng2022humannerf,bagautdinov2021driving,lombardi2018deep,xu2021h} to expedite this process. These methods focus on rendering and animating human actors but do not address the interactions between human and real scenes (\eg, a person sitting on a couch with arms on a table). Moreover, they cannot handle challenging motions such as finger movements. 
Recently, deformable NeRF methods~\cite{park2021nerfies,park2021hypernerf,pumarola2021d,tretschk2021non} have been proposed to model the person and the scene by learning an implicit deformation field along with a canonical radiance field that serves as the template. However, these methods can only handle small deformations as it is hard to model complex human motions with a single canonical representation. Besides, 
%the  implicit deformation field is hard to control,
%the expensive ray tracing prohibits real-time rendering for general NeRF  methods.
the expensive volumetric rendering process prohibits real-time video synthesis.

% In this work we aim to capture and render the character in an unconstrained  environment. To this end, we need to solve the following challenges: 1)  How can we capture the character using a cheap device while resolving the 3D motion ambiguity? Meanwhile we hope that some driving signal, \eg, 2D keypoints, can be easily obtained from the same setup.  2) What 3D aware representation is appropriate to offer sufficient expressivity to model non-rigid movement and complex appearance of the scene while being computationally efficient? Also, such representation should be amenable to the control signal conditioning.

%To model the character in an unconstrained environment, we propose a novel capture setup which comprises a portable capture rig and a fixed reference camera. Synchronized mobile phone cameras are mounted on the rig and the videographer slightly moves the camera rig to capture the light field of the scene. By doing so, we strike a balance between 3D sensing accuracy and hardware affordability since the rig movement greatly reduces the number of cameras needed. Note that we also require a fixed camera used to extract the driving keypoints that users can manipulate for character retargeting.

For character modeling in unconstrained environments, we propose a novel capture setup that comprises a portable capture rig and a fixed driving camera. Synchronized mobile phone cameras are mounted on the rig, and the videographer slightly moves the camera rig to capture the light field of the scene. The fixed camera is used to extract the driving keypoints that users can manipulate for character retargeting. With this handy setup, we strike the desired balance between 3D sensing accuracy and hardware affordability since the rig movement greatly reduces the number of cameras needed.

\begin{figure}[t]
\centering
\includegraphics[height=4.5cm]{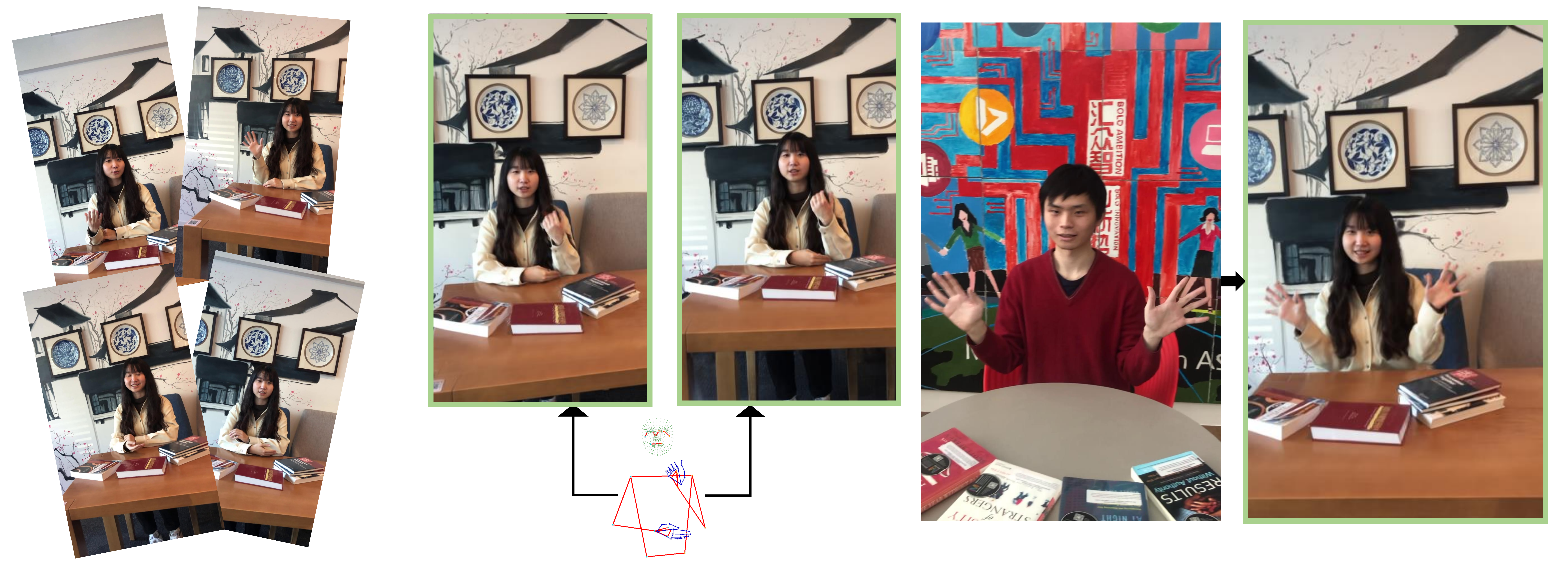}
\put(-347,-10){\footnotesize(a) Input images}
\put(-243,-10){\footnotesize(b) Pose guided}
\put(-247,-20){\footnotesize novel view synthesis}
\put(-122,-10){(c) Motion transfer}
\put(-248,23){\tiny 30FPS}
\put(-182,23){\tiny 30FPS}
\caption{\small Using images from a handy capture device, our method renders a photo-realistic character in real time. The character is animatable through motion transfer. }
\label{fig:teaser}
% \vspace{-2em}
\end{figure}
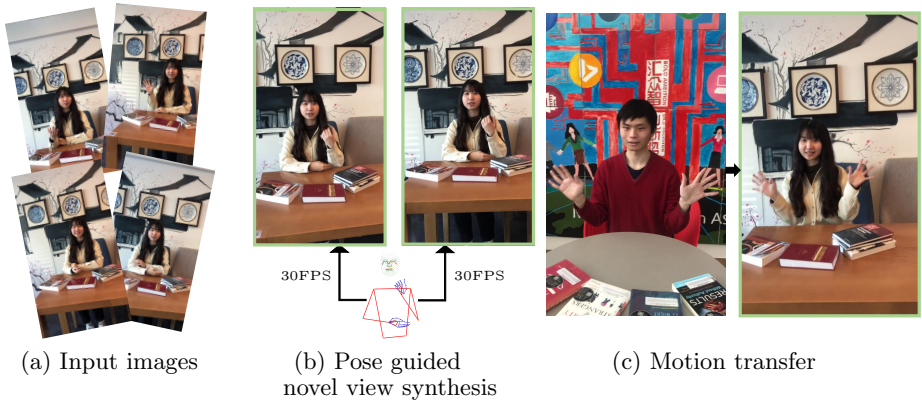

%We also introduce \emph{pose-guided multiplane images} for fast and controllable character rendering with high fidelity. Multiplane image (MPI) representation~\cite{zhou2018stereo} uses a set of parallel semi-transparent planes to approximate the light field and has shown compelling quality for complex scene modeling. Rather than performing optimization directly upon MPI~\cite{zhou2018stereo,flynn2019deepview,mildenhall2019local,wizadwongsa2021nex}, we propose to use a neural network to produce the planes with pose conditioning, and the whole framework can be regarded as a pose-to-MPI translation network. Since the 2D keypoints denote the pose in the driving camera and the predicted multiplane images lie in the frustum of the same camera, the two representations are spatially aligned and compatible for network processing. During inference, the keypoints serve as the driving signal, and one can obtain the corresponding character in novel views by rendering the predicted MPI from the target viewpoint.

We also introduce \emph{pose-guided multiplane images} for fast and controllable character rendering with high fidelity. The multiplane image (MPI) representation~\cite{zhou2018stereo} uses a set of parallel semi-transparent planes to approximate the light field and has shown compelling quality for complex scene modeling. Rather than performing optimization directly upon MPI~\cite{zhou2018stereo,flynn2019deepview,mildenhall2019local,wizadwongsa2021nex}, we propose to use a neural network to produce the planes with pose conditioning, and the whole framework is essentially a pose-to-MPI translation network. We use 2D keypoints on the image plane of the driving camera to define the pose, and predict the multiplane images in the frustum of the same camera. Hence, the two representations are spatially aligned and compatible for network processing. During inference, the keypoints serve as the driving signal, and one can obtain the corresponding character in novel views by rendering the predicted MPI from the target viewpoint.

%Our method brings several benefits. 1) We do not assume a human model or a shared canonical template but learn the character modeling purely in a data-driven manner, which offers improved flexibility to characterize the subject as well as complex interactions with the scene. In particular, we see substantial improvement for gesture modeling, which is challenging in prior arts. 2) Taking advantage of the inductive bias of convolutional neural networks, our method demonstrates improved generalization ability. Instead of memorizing the scene, the network trained from large data is equipped with generative power and can hallucinate plausible outputs for diverse poses. 3) The MPI rendering is blazingly fast, and the network can synthesize $640\times 360$ resolution videos in real-time.

Our method brings several benefits. 1) Since we do not assume a human model or a canonical template but learn character modeling purely in a data-driven manner, our method offers improved flexibility to characterize the subject as well as complex interactions with the scene. In particular, we observe substantial improvement for gesture modeling, which is challenging for prior arts. 2) Taking advantage of the inductive bias of convolutional neural networks, our method demonstrates improved generalization ability. Instead of memorizing the scene, the network trained with large data can hallucinate plausible outputs for diverse poses. 3) The MPI rendering is blazingly fast, and our network can synthesize videos of $640\times 360$ resolution in real time.

%We further propose several techniques to achieve better MPI synthesis. As opposed to placing the planes equally in disparity, we propose adaptive planes whose positions are jointly optimized during training, which considerably improves the modeling quality because the planes are now densely placed near the real surfaces of the scene. To compensate for the exposure mismatch of cameras, we also introduce to learn exposure code for each camera. Moreover, when dealing with long video sequences (\eg, $>4$k frames), we observe unsatisfactory camera poses estimation using conventional structure-from-motion (SfM) pipeline~\cite{schonberger2016structure}, which leads to blurry results. To solve this, we online refine the camera poses using the gradient of the static background pixels.

We further propose several techniques to achieve better MPI synthesis. As opposed to evenly placing the planes in disparity space, we propose adaptive planes whose positions are jointly optimized during training, which considerably improves the modeling quality because the planes are now densely placed near the real surfaces of the scene. To compensate for the exposure mismatch among different cameras, we also introduce a learnable exposure code for each camera. Moreover, when dealing with long video sequences (\eg, $>4$k frames), we observe unsatisfactory camera pose estimation using conventional structure-from-motion (SfM) pipeline~\cite{schonberger2016structure}, which leads to blurry results. To solve this, we refine the camera poses during training using the gradients of the static background pixels.

%We demonstrate that our pose-guided MPI quantitatively and qualitatively outperforms state-of-the-art approaches including Video-NeRF~\cite{li2021neural}, Nerfies~\cite{park2021nerfies}, HyperNeRF~\cite{park2021hypernerf}, and NeX~\cite{wizadwongsa2021nex} on novel view synthesis for characters with complex motions. Moreover, the character rendered by our method can be explicitly controlled and we show photorealistic results of character reenactment, as an example in Fig.~\ref{fig:teaser}. The whole rendering framework runs in real-time and is scalable to high resolutions. 

We demonstrate that our pose-guided MPI quantitatively and qualitatively outperforms state-of-the-art approaches including Video-NeRF~\cite{li2021neural}, Nerfies~\cite{park2021nerfies}, HyperNeRF~\cite{park2021hypernerf}, and NeX~\cite{wizadwongsa2021nex} on novel view synthesis for characters with complex motion. Moreover, the character rendered by our method can be explicitly controlled, and we achieve photorealistic results of character reenactment, as illustrated in Fig.~\ref{fig:teaser}. The whole rendering framework runs in real time and is scalable to high resolutions.

% Propose conditional MPI for dynamic modelling.  There are two primary advantages of conditional MPI. 1). Both fast in inference novel condition and novel view 2). In comparison with implicit methods, it enables the explicit control. In the next section, we further extend the structure to dynamic scene for human body control. 

% Our primary contribution are: 
% 1). A novel structure of conditional multi-plane images which achieve reliable high-quality novel view synthesis results on static and dynamic scenes which enables for fast inference of novel conditions (dynamic scenes) and real-time rendering of novel views. 
% 2). A practical solution for capturing and rendering real world controllable human avatar including the head, body skeleton and hands using key points as guidance and a demonstration that successfully transfer motion between different people for novel view synthesis.  
% 3). A bag of modules including learnable planes, learnable poses and adaptive exposure adjustment which may improve the quality of general MPI training. 

\section{Related Work} 
\subsubsection{Neural character rendering~} 
Traditional graphics pipelines~\cite{carranza2003free,casas20144d,collet2015high,guo2019relightables} require a well-orchestrated studio with a dense array of cameras to build the mesh for the characters. In the past years, deep generative neural networks have been introduced to synthesize photorealistic  characters~\cite{tewari2020state,tewari2021advances}. A popular synthesis paradigm is image-to-image translation~\cite{isola2017image,wang2018high}, which learns the mapping from certain representations, such as joint heatmap~\cite{ma2017pose,aberman2019deep,balakrishnan2018synthesizing,zhang2020cross,zhou2021cocosnet}, rendered skeleton~\cite{chan2019everybody,pumarola2018unsupervised,si2018multistage,shysheya2019textured,wang2018video}, and depth map~\cite{martin2018lookingood}, to real images of the character. These works have certain generalization ability to novel poses and have shown compelling rendering quality even for complex clothing and in-the-wild scenes. However, they cannot guarantee view consistency as they learn the generation in 2D screen space. More recent methods attempt to solve this by leveraging a human model, \eg, SMPL model~\cite{loper2015smpl}. One line of works~\cite{liu2020neural,sarkar2020neural,bagautdinov2021driving} unwraps the body mesh to 2D UV space where the network learns texture synthesis and then translates the rendered textured mesh to images. Meanwhile, another line of works improves view consistency by learning the deformation to a canonical 3D space enforced by the SMPL model~\cite{habermann2021real}. In comparison, our method does not assume an explicit human model and hence can model complicated finger motions as well as the character's interactions with the scene. Our method generalizes the image translation but the output representation we adopt ensures multi-view consistency.

\subsubsection{Neural scene representation~} Instead of rendering with a black-box image translation process, recent works turn to using neural networks to model some intrinsic aspects of the scene followed by a physics-based differentiable renderer. One recent notable work is to model the scene as a neural radiance field (NeRF) whose color and volumetric density across the continuous coordinates are parameterized by the multilayer perceptron (MLP). The NeRF representation ensures that the images rendered at different views have coherent geometry, and the volumetric rendering is able to produce realistic outputs with stunning details. 

Several NeRF variants~\cite{pumarola2021d,park2021nerfies,park2021hypernerf,tretschk2021non,gao2021dynamic,li2021neural} have been proposed to handle dynamic scenes. One category of these works is deformation-based~\cite{pumarola2021d,park2021nerfies,park2021hypernerf,tretschk2021non} which optimizes a deformation field that warps each observed point to a canonical NeRF. These approaches have shown impressive quality even for in-the-wild scenes, but only simple deformation can be modeled as it is difficult to find a template that accounts for all the observations. To render characters with NeRF~\cite{liu2021neural,weng2022humannerf,xu2021h,peng2021animatable}, there are works that use the SMPL model to enforce the canonical NeRF, but they may suffer from the limited modeling capability of the parametric human model. Some methods modulate NeRF with additional conditioning~\cite{li2021neural,zhang2021editable} and achieve enhanced expressivity to model dramatic topology change. Nonetheless, the majority of these methods are designed for replaying the dynamic scene, and it is hard to generalize to novel character poses. This is because the point-wise MLPs in NeRF models do not leverage a larger context for generative modeling. In contrast, we employ a convolutional neural network which leverages a large image context for hallucination, and thus our method is more amenable to animation and generalizes better. Moreover, our method renders characters much faster than NeRF-based approaches.

\subsubsection{Multiplane images representation~} 
The MPI representation uses a stack of RGB$\alpha$ layers arranged at various depths to approximate the light field. Initially, the MPI is used for the stereo magnification problem~\cite{zhou2018stereo} where an MPI is predicted from a CNN given a stereo pair input. Later, a few works extend the MPI for view synthesis where images from a variable number of viewpoints can be accessed~\cite{flynn2019deepview,mildenhall2019local,wizadwongsa2021nex,srinivasan2019pushing,tucker2020single,li2021mine}. The multiplane images can be optimized in three ways: direct optimization~\cite{mildenhall2019local,wizadwongsa2021nex,srinivasan2019pushing}, using a CNN to learn gradient updates upon MPI~\cite{flynn2019deepview} or directly predicting the MPI using CNN~\cite{tucker2020single,li2021mine}. The main drawback of MPI methods is that the rendered view range is limited by the number of planes. To solve this, \cite{mildenhall2019local} proposes to use multiple MPIs to account for local light fields, which are further blended for the final output. MPI-based approaches have limited power for handling non-Lambertian surfaces and wide viewing angles, and they are outshined by the recent advances of NeRF-based methods. Very recently, NeX~\cite{wizadwongsa2021nex} tackles view-dependent modeling by representing the pixel color of MPI as a combination of spherical basis functions, and it excels over NeRF in visual quality with much faster rendering speed. However, no existing method ever studies the controllability of such a representation. Our approach is motivated by the recent success of MPI, and we demonstrate the great potential of this representation for animatable character rendering.  

\begin{figure}[!t]
\centering
\includegraphics[width=\linewidth]{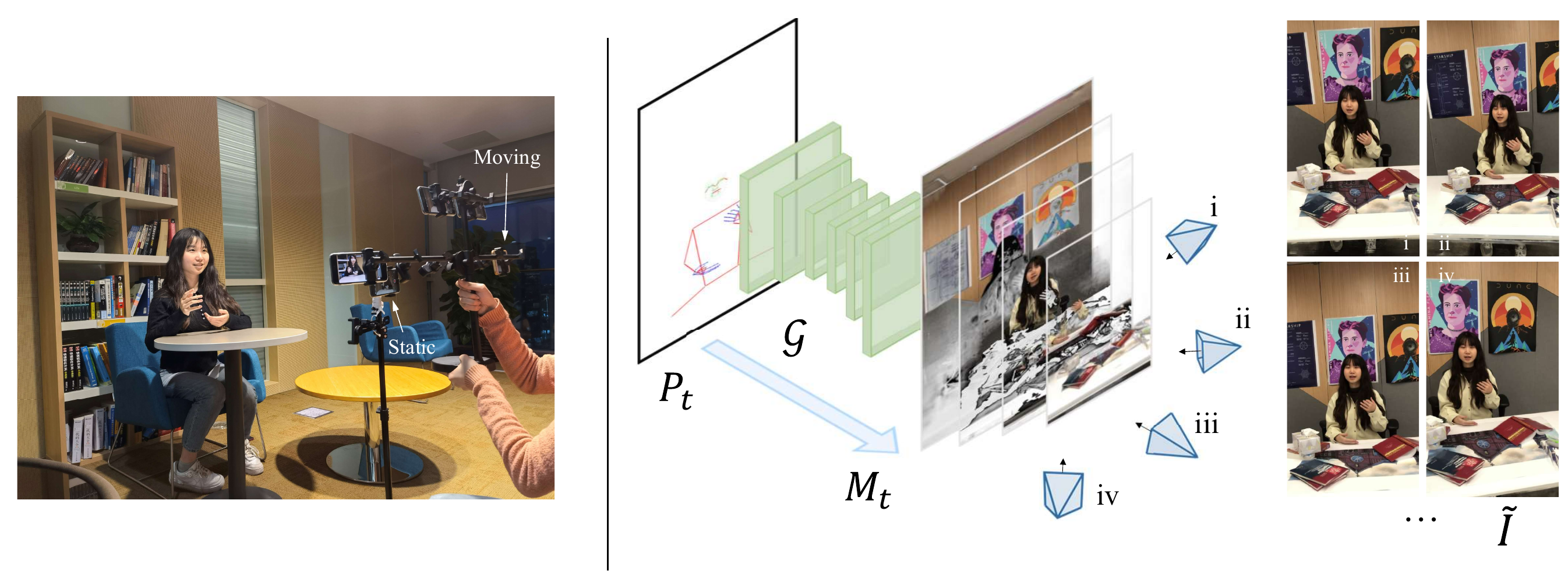}
\put(-290,1){\footnotesize (a)}
\put(-117,1){\footnotesize (b)}
%\put(-98,14){\footnotesize cameras}
\caption{Overview of our method.
% A 3D controllable avatar representation can be acquired after a close-form training procedure. 
% The 3D avatar learned using our framework can not only be rendered in novel views, but also be controlled by any given pose: 
(a) The device we use to capture data for building an animatable character in a real scene.
(b) The built character is controllable. Given any pose $\mathbf{P}_t$ at time step $t$, we feed its pose image into a pre-trained network $\mathcal{G}_\theta$ to acquire a 3D character represented in MPI ($\mathbf{M}_t$). The character has the same pose with $\mathbf{P}_t$, and can also be efficiently rendered in free views. }
\label{fig:pipeline}
\end{figure}

\section{Approach} 
We aim to render an animatable character that can be controlled by a driving input. To this end, we devise a portable data capture setup to ease the multi-view capture in open scenes (Section~\ref{sec:data_capture}). Once we finish the data capture for the moving character, we train a neural network conditioned on character poses to predict the multiplane images that explain the multi-view observations (Section~\ref{sec:mpi}). During inference, we can render realistic characters given a driving input (Section~\ref{sec:motion_transfer}). Next, we elaborate on these three parts respectively.

% We present our method in three parts: first, we describe our data capture process of how the multiview images set $I$ are collected and registered and condition set $C$ are achieved (Section~\ref{sec:data_capture}); then we discuss our conditional MPI respresentations which extend the original MPIs to fit to dynamic scenes (Section~\ref{sec:mpi}); and finally we describe how to apply motion transfer to generate MPIs on novel conditions for the proposed representation (Section~\ref{sec:motion_transfer}). 

% \section{Controllable human avatar} 
% \begin{figure}
% \centering
% \includegraphics[height=6.5cm]{eijkel2}
% \caption{We propose a cheap and practical methods for capturing multi-view video sequences with corresponding input keypoints.}
% \label{fig:data_capture}
% \end{figure}

\subsection{Data capture}
\label{sec:data_capture}
In our design, data capturing should meet the following requirements. First, the capture device should be portable, lightweight, and low-cost to benefit as many users as possible for their character creation. Second, the captured images for the moving subject should resolve most motion ambiguity. Otherwise, the character reconstruction from sparse views is highly ill-posed. Third, we need to use some driving signal handy for user control, and such signal should be readily obtained from the same capture setup.

Taking these into consideration, we propose a novel character capture setup as shown in Fig.~\ref{fig:pipeline} (a), which consists of a moving camera rig along with a static camera fixed on a tripod. We mount four smartphone cameras on the capture rig, which the videographer can hold and move to capture the subject. Such capturing manner is motivated by Nerfies~\cite{park2021nerfies} that uses a single moving camera for selfies. In our early attempts, we find a single camera does not suffice to resolve the ambiguity for complex non-rigid motion, whereas adding a few more cameras significantly improves the modeling quality. Also, the rig motion makes the capture to cover various combinations of viewpoints and character poses, which helps to reduce the number of multi-view cameras by requiring longer sequence capturing. On the other hand, we propose to use body keypoints as the driving signal for character animation. We believe manipulating 2D keypoints is easy for most users since people can extract such keypoints from their monocular videos to drive the virtual character. Hence, we also require a fixed camera, or ``driving camera'',  which is used to extract the driving keypoints. Compared to a specialized lab, it is much cheaper to build this capture setup. Once we capture the data, we use audio to synchronize the multi-view videos with the driving camera and run COLMAP~\cite{schonberger2016structure} to estimate the camera poses for each frame. 

% Our data capture device is composed of two basic parts: a static camera and a moving camera rig with four cameras as shown in Fig.~\ref{fig:data_capture}.  We call the static camera "driving camera" as we will extract the driving keypoints from the video captured by this camera. When capturing the new sequence, we slowly hang the rig around to capture more diverse multi-view images.  Note that the images captured by the camera rig is only used for supervision as the human pose extracted from these camera is entangled with the change of camera pose. In comparison to other settings such as static camera rigs with tons of cameras, our setup is cheap and portable which enable us to capture data for various scenes. For simplicity, we use 5 iPhone8 for capturing while different cameras or cellphones are also acceptable. After we capture the sequences, we use audio for synchronization and run COLMAP to estimate the camera pose for each frame. 

\subsection{Conditional MPI representation} 
\label{sec:mpi}

Our method builds on the multiplane image (MPI) scene representation, which consists of $D$ fronto-parallel planes, each with an associated $H \times W \times 4$ RGB$\alpha$ image. As illustrated in Fig.~\ref{fig:pipeline} (b), the multiplane images are scaled and positioned at different depths $d_1,...,d_D$ within the view frustum of the driving camera. Typically, the planes are placed equally in the depth space (for the bounded scene) or in the disparity space (for the unbounded scene).

While MPI-based methods have shown impressive quality in modeling static scenes from sparse views, its ability to model a moving character, especially those with complex motions, remains underexplored. To accomplish this, we formulate the character rendering by pose-guided MPI synthesis framework, which is illustrated in Fig. \ref{fig:pipeline} (b). Given the pose image $\mathbf{P}_t$ extracted from the driving camera at time $t$, we train a convolutional neural network $\mathcal{G}_\theta$ to translate this input to multiplane images $\mathbf{M}_t = \mathcal{G}_\theta(\mathbf{P}_t)$, using the supervision of the multi-view observations $\{\mathbf{I}_t^n\}_{n=1}^N$. Here, $N$ denotes the number of cameras we use for data capture. During training, we extract rich character pose information using~\cite{lugaresi2019mediapipe}, which includes facial landmarks, body keypoints, and finger keypoints that are extracted from the driving frames. Note that both $\mathbf{P}_t$ and  $\mathbf{M}_t$ are spatially aligned as they are viewed from the same driving camera; hence their mapping is naturally suitable for the CNN learning. 

The synthesized MPI can be rendered in the target view by compositing the colors along the rays. The implementation is efficient: the image planes are first warped towards the target view and then alpha-blended. Specifically, we refer to the RGB channels of the MPI as $\mathbf{C} = \{{ c}_1,...,{ c}_D\}$ and the corresponding alpha channels as $\mathbf{A} = \{{ \alpha}_1,...,{ \alpha}_D\}$. The MPI rendering can be formulated as
\begin{equation}
    \tilde{\mathbf{I}} = \mathcal{O}\big(\mathcal{W}(C), \mathcal{W}(A)\big) ,
    \label{eq:mpi_rendering}
\end{equation}
where the warping operator $\mathcal{W}$ warps the images via a homography function~\cite{zhou2018stereo} depending on the relative rotation $\mathbf{R}$ and translation $\mathbf{t}$ from the target to the source view and the layer depth $d_i$. Formally, the warping matrix is 
\begin{equation}
    \mathbf{K}_s (\mathbf{R} - \frac{\mathbf{t} \mathbf{n}^T}{d_i}) \mathbf{K}_t^{-1},
    \label{eq:homography_warp}
\end{equation}
where $\mathbf{n}$ is the normal vector; $\mathbf{K}_s$ and $\mathbf{K}_t$ respectively are intrinsic matrix of source camera and target camera.
Besides, $\mathcal{O}$ in Equation~\ref{eq:mpi_rendering} denotes the composition~\cite{porter1984compositing} of the warped images ($c_i'$ and $\alpha_i'$) from back to front, \ie,
\begin{equation}
    \mathcal{O}({C},{A}) =  \sum_{i=1}^{D} \Big({c}_i'{\alpha}_i' \prod_{j=i+1}^{D} (1-\alpha_j') \Big).
    \label{eq:composite}
\end{equation}
The above rendering process is fully differentiable, so the MPI synthesis network can be trained end-to-end using 2D supervision.

This pose-guided MPI enjoys many features of CNNs and shows various advantages over implicit approaches. First of all, we do not explicitly model the geometry of the scene or assume a template for all the character motions, so our model is more expressive and can better fit challenging motions and delicate details, as proved in our view synthesis experiments. Second, compared to implicit neural representation, our method can generalize better thanks to the inductive bias of CNNs. Intuitively, it seems that our approach is good at hallucinating plausible outputs, even for unseen poses. Third, manipulating the keypoints is more straightforward for character animation instead of using latent code or body parameters. Finally, the MPI is inferred with a single feed-forward pass of the network, and its rendering is also fast. 

To cultivate the full potential of the above framework and achieve state-of-the-art quality, we introduce the following key components.

\subsubsection{Depth-adaptive MPI }
We argue that manually placing the multiplane images at fixed depth may not be optimal. Ideally, the planes should be distributed more densely around the real scene surfaces. Otherwise, some planes are wasted for modeling the vacant space. More importantly, in our scenario the bound of the scene cannot be reliably estimated because we have to mask out the moving foreground when computing the COLMAP. As a result, the MPIs initialized with a mistaken depth range may lead to scene clipping modeling.
 
In view of this, we propose to make the MPI depth as learnable parameters so that MPIs positions can be adaptive to the scene content. Formally, we refer to the initialized depth as $d_i^{init}$. During training, we learn the residual $\delta_i$ which is initialized with zeros, so the refined depth becomes $d_i = d_i^{init} + \delta_i$. As we know, the homography warping $\mathcal{W}$ in Equation~\ref{eq:mpi_rendering} is the function of $d_i$; hence the gradient can be back-propagated to update the depth as the training proceeds. 

However, one may notice that the depth refinement may alter the plane orderings, which will mislead the alpha composition (Equation~\ref{eq:composite}) that renders the planes from back to front. Therefore, we need to enforce the depth order to be unchanged in the depth refinement. To achieve this, we clamp the value of  $\delta_i$ once we find the plane shift causes the crossing over the adjacent planes. During training, the depth refinement uses $0.1$x learning rate compared to the network.

\subsubsection{Learning with variable exposure images}
There always exist exposure and color differences among cameras even if we employ cameras of the same type and manually choose the ISO and exposure time. To use different exposed images for our training, we assume that the MPI rendering models the true light intensity of the scene but we introduce learnable exposure coefficients to account for the exposure variance among cameras. Specifically, we adopt a linear exposure model~\cite{ouyang2021neural,abdelhamed2018high} which outputs the image as

\begin{equation}
    \hat{\mathbf{I}} =  clamp\big((\tilde{\mathbf{I}} + \bm{\beta}) \circ  \bm{\gamma} \big),
\end{equation}
where $clamp(\cdot)=\min(\max(\cdot,0),1)$ whereas $\bm{\beta}\in \mathbb{R}^3$ and $\bm{\gamma} \in \mathbb{R}^3$ are learnable exposure coefficients associated with each camera. Here, $\bm{\gamma}$ accounts for the exposure time whereas $\bm{\beta}$ is for compensating the shift of black level. 

\subsubsection{Learnable camera poses}
In our experiments, the model sometimes fails to reconstruct the static background because of the inaccurate camera pose estimation from SfM. The problem becomes even worse when dealing with long video sequences to capture more diverse character poses. 

To improve the robustness of inaccurate camera registration, we jointly refine the camera poses during training. The gradient through the homography warping can be used to update the camera poses for each frame. Note that the pose refinement can only leverage the gradient of a static background, whereas the MPI synthesis is updated using the whole image. Therefore, the optimization follows an alternative manner: we take two consecutive training steps to alternatively optimize the MPI synthesis network and the camera pose, with the loss computing over the whole image and the background, respectively. Since this may slow down the training speed, we only apply this strategy for sequences with blurry background reconstruction.

\subsubsection{Sharing textures for compact MPI} 
It is known that a large number of RGBA layers are helpful for high-fidelity modeling, but this brings huge memory costs when directly using the network for the MPI synthesis. To make the MPI more compact, we follow the strategy of NeX~\cite{wizadwongsa2021nex} and share the RGB textures for every $K$ layers. In this way, we reduce the output channels from $4D$ to $(D+3D/K$) without obvious degradation in visual quality.

\subsubsection{Losses}
To optimize our model, we feed input pose to generate the MPI and render an output image $\hat{\mathbf{I}}$ using the camera pose of the ground-truth image $\mathbf{I}$. We use three losses for reconstruction: mean square error between $\mathbf{I}$ and $\hat{\mathbf{I}}$ as $\mathcal{L}2$: $||\mathbf{I}-\hat{\mathbf{I}}||^2$, gradient difference along the width and height dimension as $\mathcal{L}_{grad}$: $||\nabla \mathbf{I}- \nabla \hat{\mathbf{I}}||^2$ and the perceptual loss of the difference between VGG features: $\mathcal{L}_{perceptual}$: $||{VGG}_F(\mathbf{I}) - VGG_F(\hat{\mathbf{I}})||_1$. In total we optimize:  
\begin{align}
\min_{\theta, d, R, t, \beta, \gamma} \mathcal{L}_2 + \lambda_1 \mathcal{L}_{grad} + \lambda_2 \mathcal{L}_{perceptual}.
\end{align}
Note that for $\mathcal{L}_2$ loss and the $\mathcal{L}_{grad}$, we apply a 10$\times$ weight on the foreground person using the object mask detected by \cite{he2016deep}.

\subsection{Motion transfer} 
\label{sec:motion_transfer}
Our pose-to-MPI translation network learns to generate 3D representation with a conditional pose. Given a trained model of a character, we can transfer the motion from the driving character and generate novel views. Since characters differ in height, limb length and body shape, it is not suitable to transfer the absolute pose directly from the driving character to the source character built by our method. Relative motion transfer, which keeps the physical characteristics of the source character, is desired. Our input pose comprises face keypoints, body keypoints, and finger keypoints. Denote the face landmarks of the driving character by $t$ and that of the source character by $s$. To transfer the relative motion, we find the landmarks $t'$ most similar to $s$ in the driving video. Thus, the transferred pose for the source character becomes  $s + t - t'$.  We treat the body landmarks as a tree structure for body motion transfer. We generate the transferred body keypoints by keeping the same limb length as the source body while utilizing the limb direction of the driving body. The tree root is the midpoint of the left shoulder and right shoulder. Finger motion transfer is similar to body transfer, except that the tree root is changed to the wrist. Please refer to the supplementary material for more details.

\section{Experiments}

\renewcommand{\arraystretch}{.5}

\subsection{Implementation details}
% For the output MPI, we set the number of alpha planes to 192 and every 12 alpha planes share a RGB texture plane and thus the output channel size is 240. We capture our video sequence in a vertical mode in 1080p for 1 minutes to 3 minutes for each person. We resize the video to 360x640 for evaluation and pad 180 pixels both in width and height to generate better MPI. For each scene, we choose 1/16 images as validation image. The learning rate is decayed from 1e-3 to 1e-4 in 500 epochs. The training takes 8~12 hours using four Nvidia v100 GPUs. 

The output MPI is composed of $192$ alpha layers, with every $12$ alpha layers sharing the same RGB texture layer, which leads to $192/12=16$ RGB texture layers. The total number of channels to output thus becomes $192+16\times3=240$. 
Our video sequences are all captured in 1080p resolution. The temporal length of each captured sequence lies between 1 to 3 minutes. For each sequence, 1/16 frames are selected as a validation set with the rest left for training. %The resolution of each MPI layer is $(360+2\times 180) \times (640 + 2 \times 180)$
We down-sample each video frame to $360\times 640$ for fast inference.
However, the resolution of each output MPI layer is larger than this to support rendering with wider view angles:
it is equal to padding 180 pixels to all four sides of the $360\times 640$ frame. 
We train the model using Adam\cite{kingma2014adam} optimizer and decay the learning rate from 1e-3 to 1e-4 in 500 epochs. The training takes 8 to 12 hours with four Tesla V100 GPUs.

\renewcommand{\arraystretch}{1.05}
\begin{table*} [t]
\scriptsize
\centering 
\begin{tabular}{@{\hspace{0.5mm}}l  @{\hspace{2mm}} c @{\hspace{0.5mm}} c @{\hspace{0.5mm}}  c @{\hspace{0.5mm}}c@{\hspace{2mm}} c @{\hspace{0.5mm}} c @{\hspace{0.5mm}} c @{\hspace{0.5mm}} c@{\hspace{2mm}}c @{\hspace{0.5mm}} c@{\hspace{0.5mm}} c @{\hspace{0.5mm}}c@{\hspace{2mm}} c @{\hspace{0.5mm}} c @{\hspace{0.5mm}} c @{\hspace{0.5mm}}} 
\toprule 
& \multicolumn{3}{c}{\textbf{Sequence 1 }}&& \multicolumn{3}{c}{\textbf{Sequence 2}}&& \multicolumn{3}{c}{\textbf{Sequence 3}} && \multicolumn{3}{c}{\textbf{ }} \\ 
& \multicolumn{3}{c}{\textbf{(755 images) }}&& \multicolumn{3}{c}{\textbf{(755 images)}}&& \multicolumn{3}{c}{\textbf{(755 images)}} && \multicolumn{3}{c}{\textbf{ MEAN}} \\ 
\cmidrule{2-4} \cmidrule{6-8} \cmidrule{10-12} \cmidrule{14-16}
  &  PSNR & SSIM & LPIPS && PSNR & SSIM& LPIPS&& PSNR&SSIM&LPIPS&& PSNR&SSIM&LPIPS\\ 
\midrule % In-table horizontal line
 NeX\cite{wizadwongsa2021nex} & 24.05 & 0.919 & 0.168 && 22.63 & 0.921 & 0.197 && 24.26 &  0.908 & 0.225 && 23.65 &0.916& 0.197 \\
Video-Nerf\cite{li2021neural} & 27.74 &  0.940 & 0.189&&  26.94 & 0.932 & 0.230 &&  27.20 & 0.931 & 0.221 && 27.30 & 0.934 & 0.213 \\
Nerfies\cite{park2021nerfies} & 27.91 &  0.938 & 0.180 && 26.90 & 0.922 & 0.180 && 26.84 & 0.932 & 0.191&& 27.22 & 0.931 & 0.187 \\ 
HyperNerf\cite{park2021hypernerf} & 28.10 & 0.946 & 0.162 && \textbf{27.13} & 0.942 & 0.178 && \textbf{27.23}  & 0.939 & 0.193 && \textbf{27.48}  & 0.942  & 0.178 \\
\midrule
Ours &  \textbf{28.20} &  \textbf{0.954} & \textbf{0.062} &&  26.72 & \textbf{0.957} & \textbf{0.082} && 26.95 & \textbf{0.945} & \textbf{0.092} && 27.29 & \textbf{0.952} & \textbf{0.079}\\

\bottomrule
\end{tabular}
\vspace{0.8mm}
\caption{{Quantitative comparisons on different datasets in terms of PNSR$\uparrow$, SSIM$\uparrow$, and LPIPS$\downarrow$. The best results are highlighted in \bf{bold}.} } 
\label{tab:metric} 
\end{table*}  

\renewcommand{\arraystretch}{.5}
\begin{figure*}[!ht]
    \centering 
    \small
    \begin{tabular}{@{}l@{\hspace{1.5mm}}c@{\hspace{0.5mm}}c@{\hspace{0.5mm}}c@{\hspace{0.5mm}}c@{\hspace{0.5mm}}c@{\hspace{0.5mm}}c@{}}
    \rotatebox{90}{{\qquad Scene 1}}&
     \includegraphics[trim={0 0 0 3cm},clip=true,width=0.15\linewidth]{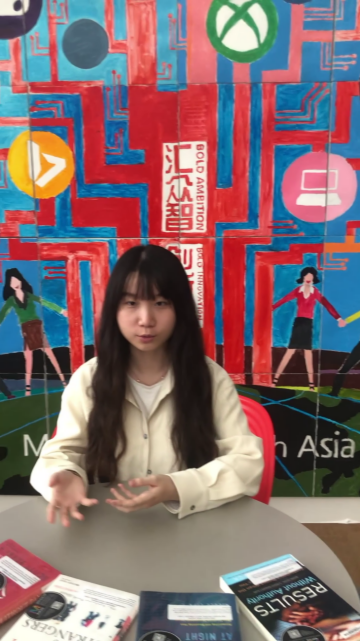}       &   
     \includegraphics[trim={0 0 0 3cm},clip=true,width=0.15\linewidth]{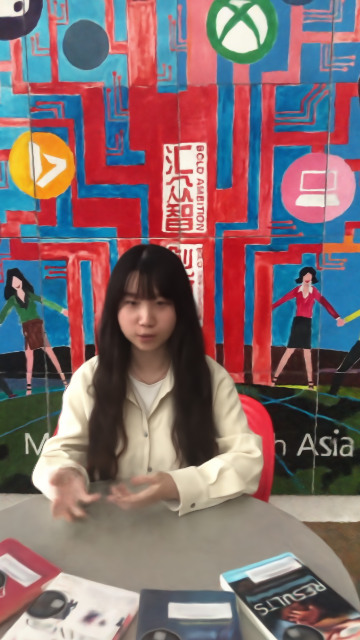}       &        
     \includegraphics[trim={0 0 0 3cm},clip=true,width=0.15\linewidth]{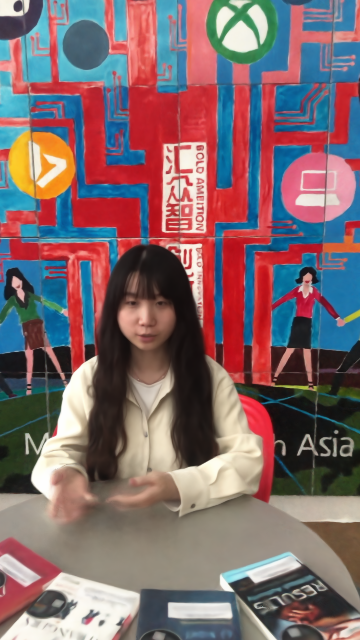}        &   
     \includegraphics[trim={0 0 0 3cm},clip=true,width=0.15\linewidth]{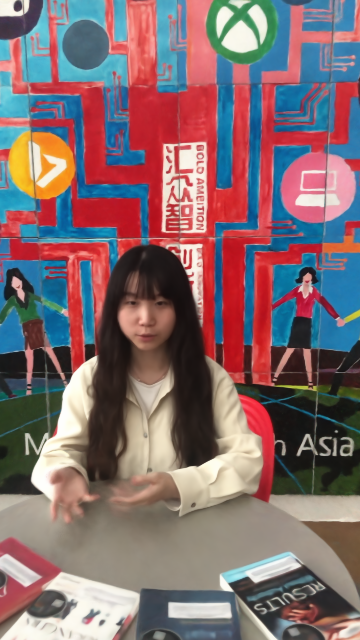} &   
     \includegraphics[trim={0 0 0 3cm},clip=true,width=0.15\linewidth]{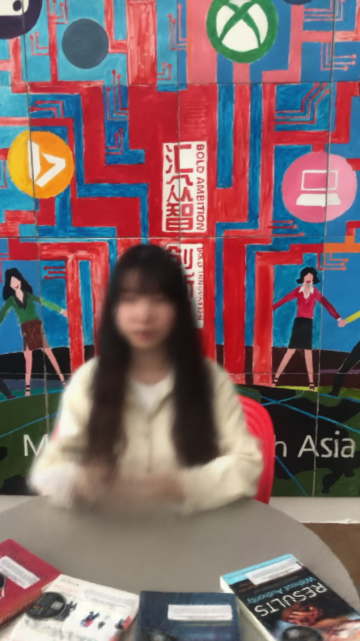} &
     \includegraphics[trim={0 0 0 3cm},clip=true,width=0.15\linewidth]{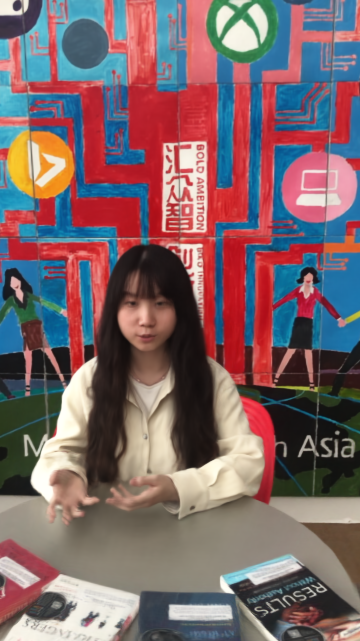}
     \\
    
    % \rotatebox{90}{{\qquad Scene 1}}&
    %  \includegraphics[trim={0 0 0 3cm},clip=true,width=0.15\linewidth]{figures/seq1/00011_gt.png}       &   
    %  \includegraphics[trim={0 0 0 3cm},clip=true,width=0.15\linewidth]{figures/seq1/00011_dnerf.png}       &        
    %  \includegraphics[trim={0 0 0 3cm},clip=true,width=0.15\linewidth]{figures/seq1/00011_nerfies.png}        &   
    %  \includegraphics[trim={0 0 0 3cm},clip=true,width=0.15\linewidth]{figures/seq1/00011_hypernerf.png} &   
    %  \includegraphics[trim={0 0 0 3cm},clip=true,width=0.15\linewidth]{figures/seq1/00011_nex.png} &
    %  \includegraphics[trim={0 0 0 3cm},clip=true,width=0.15\linewidth]{figures/seq1/00011_ours.png}
    %  \\

    % \rotatebox{90}{{\qquad Scene 1}}&
    %  \includegraphics[trim={0 0 0 3cm},clip=true,width=0.15\linewidth]{figures/seq1/00015_gt.png}       &   
    %  \includegraphics[trim={0 0 0 3cm},clip=true,width=0.15\linewidth]{figures/seq1/00015_dnerf.png}       &        
    %  \includegraphics[trim={0 0 0 3cm},clip=true,width=0.15\linewidth]{figures/seq1/00015_nerfies.png}        &   
    %  \includegraphics[trim={0 0 0 3cm},clip=true,width=0.15\linewidth]{figures/seq1/00015_hypernerf.png} &   
    %  \includegraphics[trim={0 0 0 3cm},clip=true,width=0.15\linewidth]{figures/seq1/00015_nex.png} &
    %  \includegraphics[trim={0 0 0 3cm},clip=true,width=0.15\linewidth]{figures/seq1/00015_ours.png}
    %  \\

    \rotatebox{90}{{\qquad Scene 1}}&
     \includegraphics[trim={0 0 0 3cm},clip=true,width=0.15\linewidth]{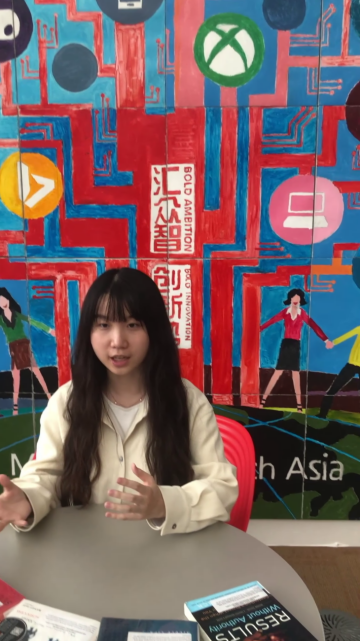}       &   
     \includegraphics[trim={0 0 0 3cm},clip=true,width=0.15\linewidth]{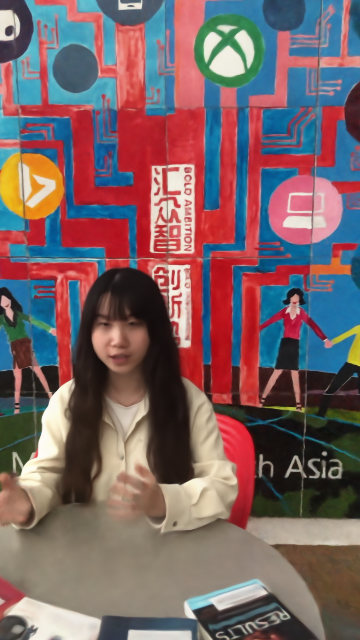}       &        
     \includegraphics[trim={0 0 0 3cm},clip=true,width=0.15\linewidth]{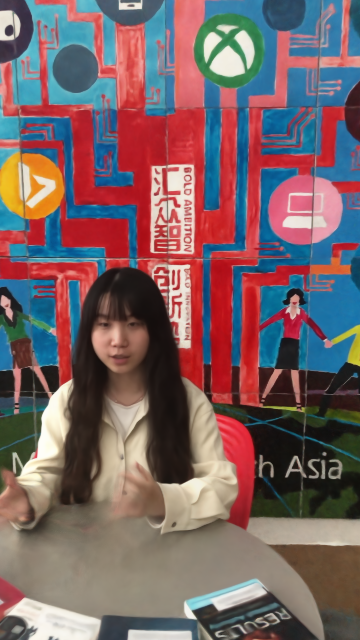}        &   
     \includegraphics[trim={0 0 0 3cm},clip=true,width=0.15\linewidth]{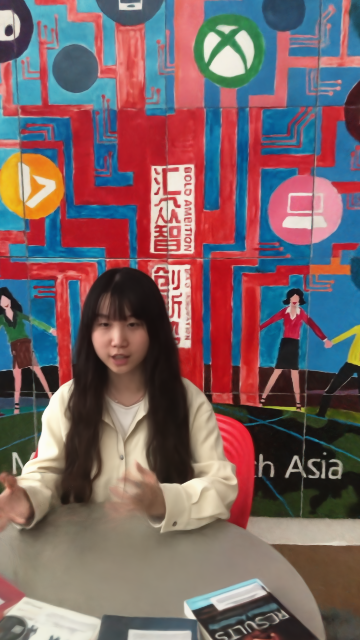} &   
     \includegraphics[trim={0 0 0 3cm},clip=true,width=0.15\linewidth]{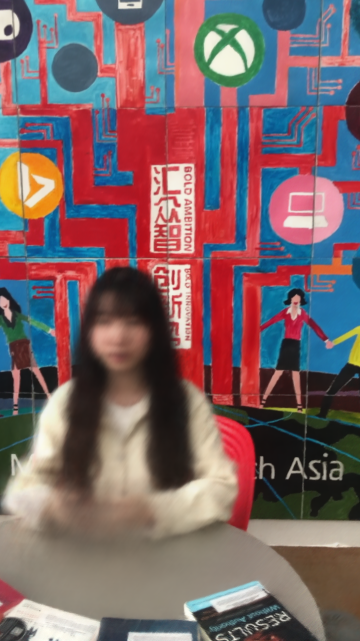} &
     \includegraphics[trim={0 0 0 3cm},clip=true,width=0.15\linewidth]{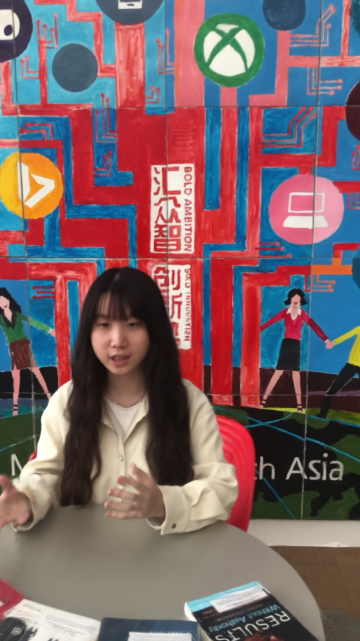}
     \\
     
     \rotatebox{90}{{\qquad Scene 2}}&
     \includegraphics[trim={0 0 0 3cm},clip=true,width=0.15\linewidth]{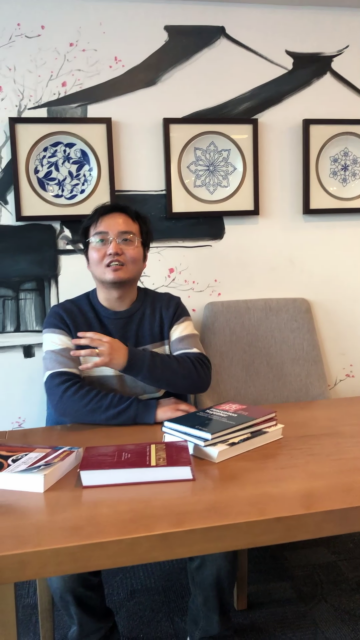}       &   
     \includegraphics[trim={0 0 0 3cm},clip=true,width=0.15\linewidth]{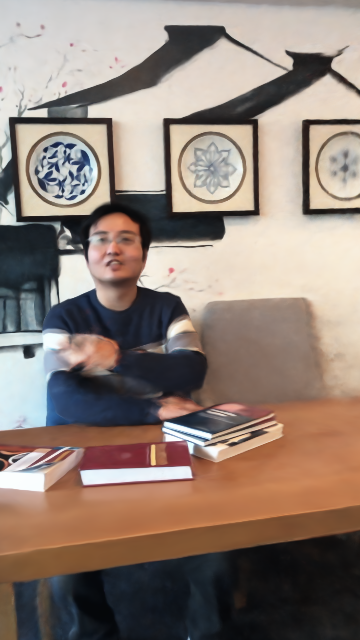}       &        
     \includegraphics[trim={0 0 0 3cm},clip=true,width=0.15\linewidth]{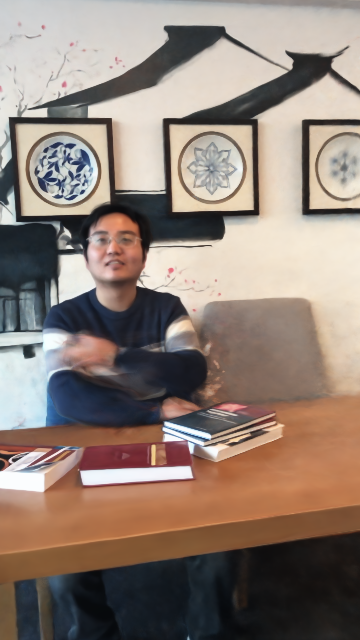}        &   
     \includegraphics[trim={0 0 0 3cm},clip=true,width=0.15\linewidth]{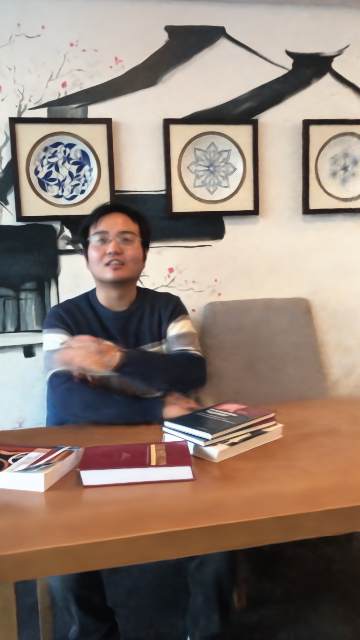} &   
     \includegraphics[trim={0 0 0 3cm},clip=true,width=0.15\linewidth]{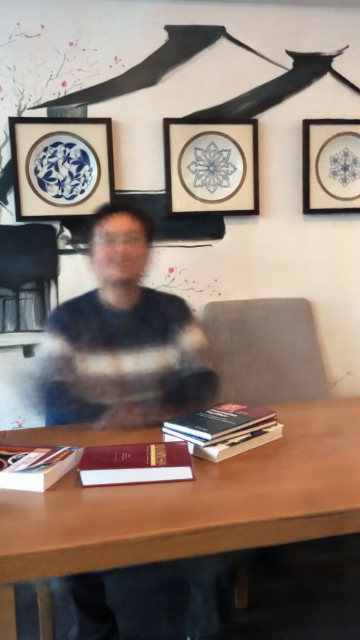} &
     \includegraphics[trim={0 0 0 3cm},clip=true,width=0.15\linewidth]{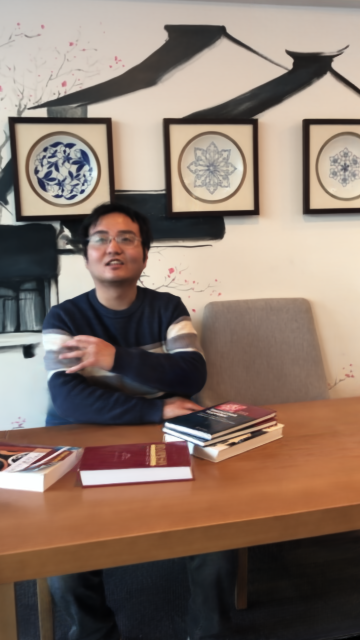}
     \\
     
     \rotatebox{90}{{\qquad Scene 2}}&
     \includegraphics[trim={0 0 0 3cm},clip=true,width=0.15\linewidth]{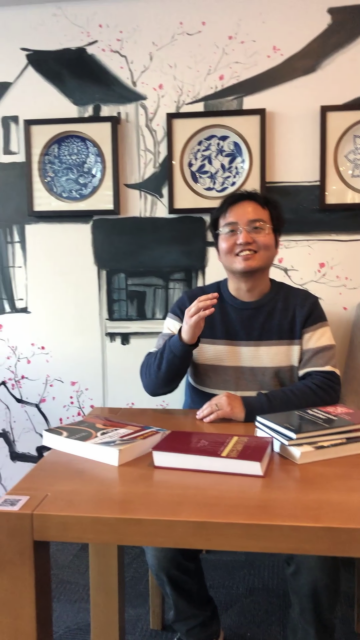}       &   
     \includegraphics[trim={0 0 0 3cm},clip=true,width=0.15\linewidth]{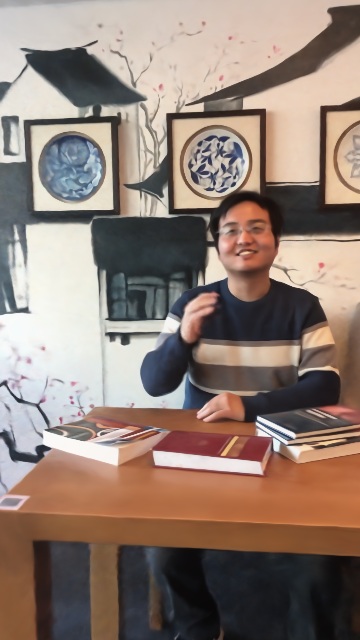}       &        
     \includegraphics[trim={0 0 0 3cm},clip=true,width=0.15\linewidth]{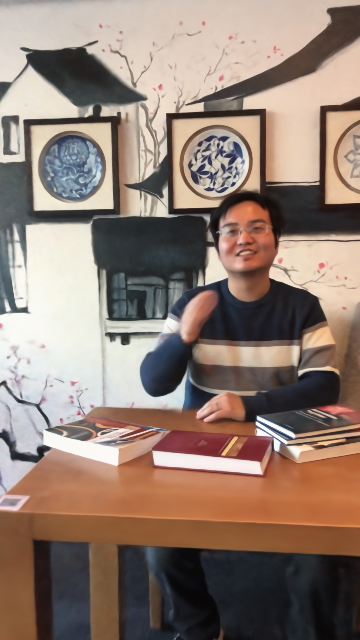}        &   
     \includegraphics[trim={0 0 0 3cm},clip=true,width=0.15\linewidth]{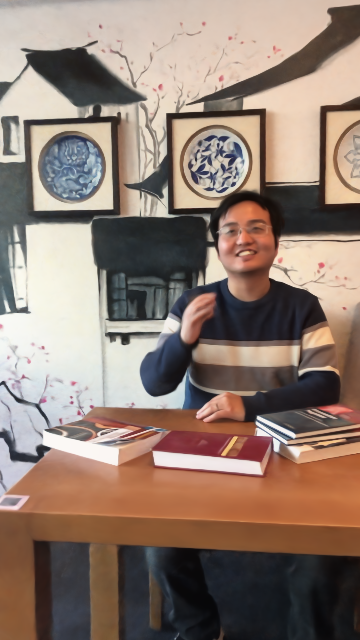} &   
     \includegraphics[trim={0 0 0 3cm},clip=true,width=0.15\linewidth]{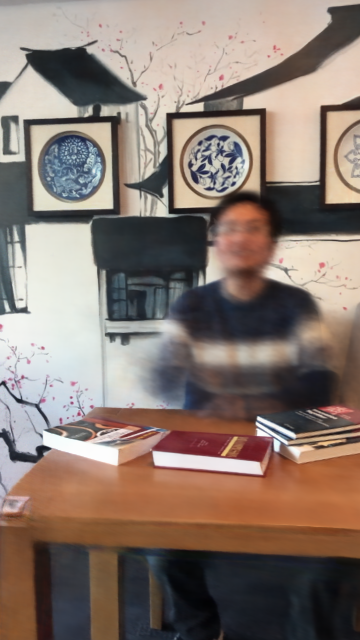} &
     \includegraphics[trim={0 0 0 3cm},clip=true,width=0.15\linewidth]{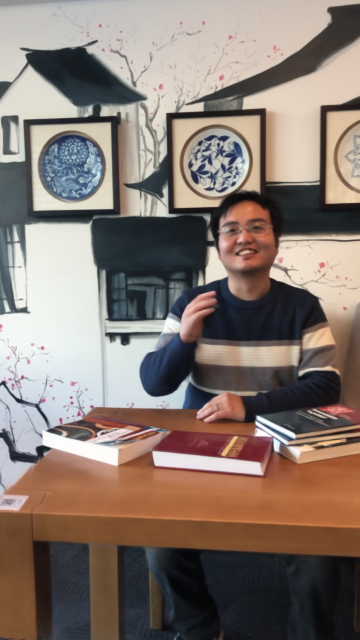}
     \\
     
     \rotatebox{90}{{\qquad Scene 3}}&
     \includegraphics[trim={0 0 0 3cm},clip=true,width=0.15\linewidth]{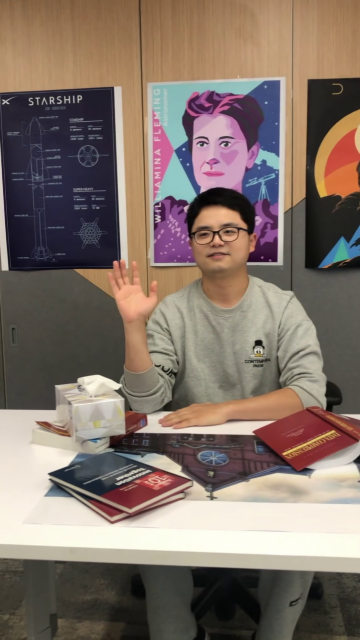}       &   
     \includegraphics[trim={0 0 0 3cm},clip=true,width=0.15\linewidth]{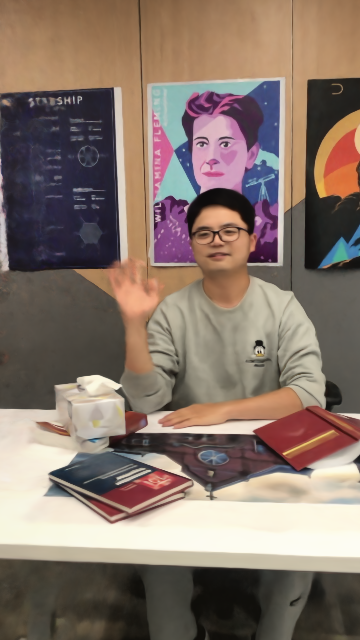}       &        
     \includegraphics[trim={0 0 0 3cm},clip=true,width=0.15\linewidth]{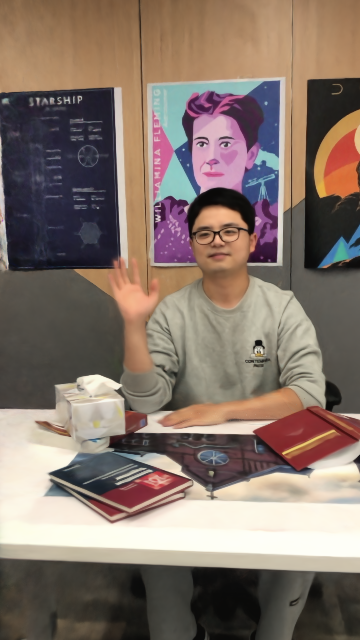}        &   
     \includegraphics[trim={0 0 0 3cm},clip=true,width=0.15\linewidth]{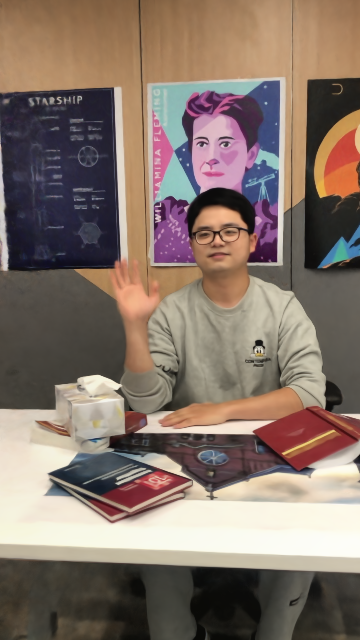} &   
     \includegraphics[trim={0 0 0 3cm},clip=true,width=0.15\linewidth]{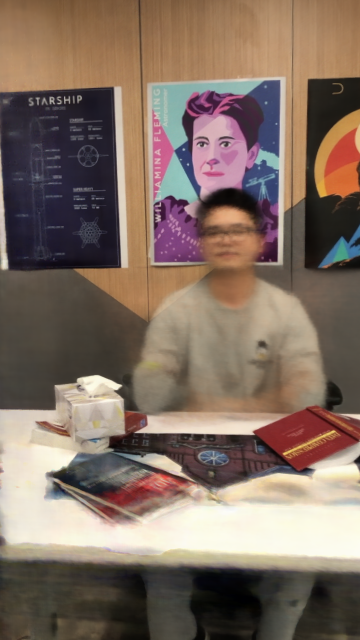} &
     \includegraphics[trim={0 0 0 3cm},clip=true,width=0.15\linewidth]{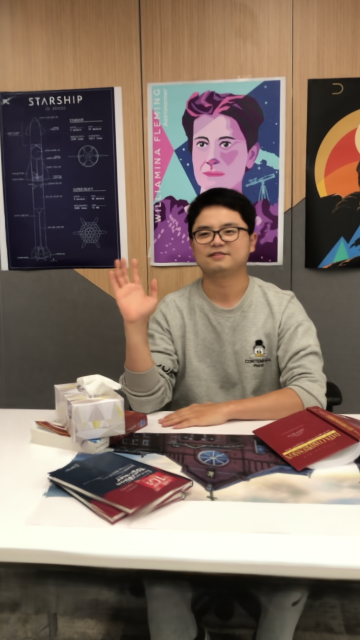}
     \\
     
     \rotatebox{90}{{\qquad Scene 3}}&
     \includegraphics[trim={0 0 0 3cm},clip=true,width=0.15\linewidth]{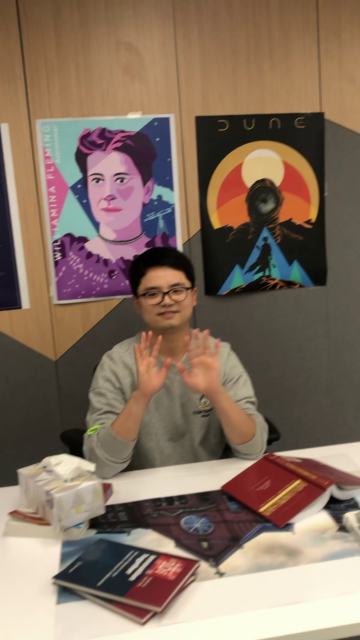}       &   
     \includegraphics[trim={0 0 0 3cm},clip=true,width=0.15\linewidth]{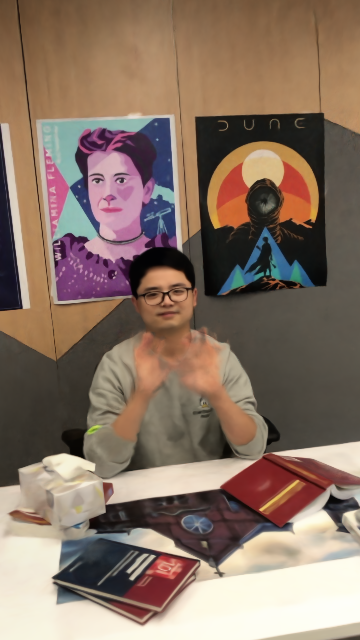}       &        
     \includegraphics[trim={0 0 0 3cm},clip=true,width=0.15\linewidth]{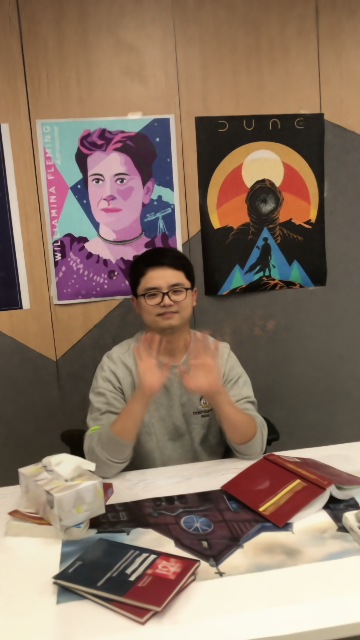}        &   
     \includegraphics[trim={0 0 0 3cm},clip=true,width=0.15\linewidth]{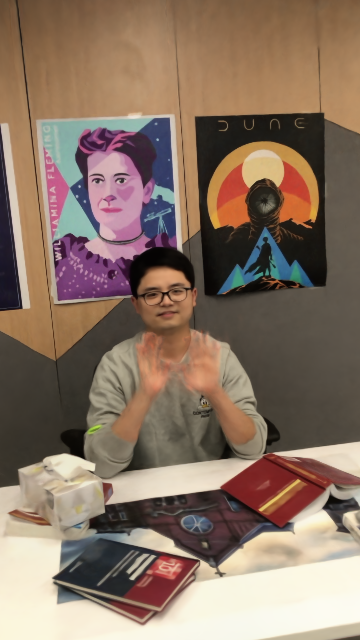} &   
     \includegraphics[trim={0 0 0 3cm},clip=true,width=0.15\linewidth]{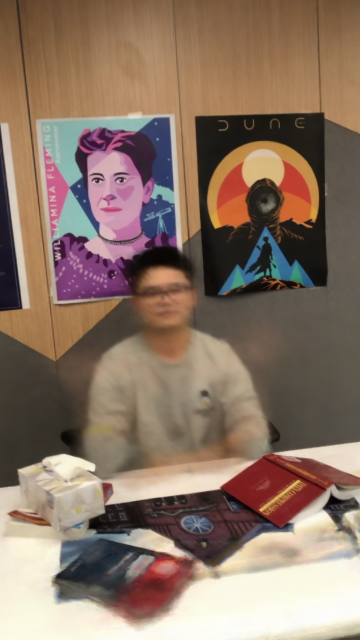} &
     \includegraphics[trim={0 0 0 3cm},clip=true,width=0.15\linewidth]{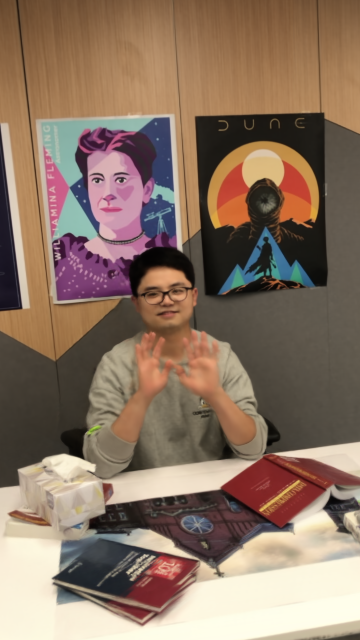}
     \\
   
     & GT & Video-NeRF & Nerfies  & HyperNerf &  NeX & Ours  \\
    \end{tabular}
    \vspace{-0.5mm}
    \caption{{Visual comparisons of different methods on validation novel views. }}
    \label{fig:qualitative}
\end{figure*}

% Our pipeline is implemented in pytorch.  

\subsection{Evaluation}

% We analyze the performance of the proposed approach both quantitatively and qualitatively. We highly recommend the reader to view the supplementary video for the visual results. Our methods are evaluated in the following two aspects:(i) the ability to synthesis the novel views (ii) the ability to generalize to novel poses.

Both quantitative and qualitative analysis is conducted to evaluate our methods thoroughly w.r.t the following two aspects: (i) the ability to synthesize novel views (ii) the ability to generalize to novel poses. We highly recommend the readers to watch our supplementary video for a more comprehensive evaluation.  

\renewcommand{\arraystretch}{1.05}
\setlength{\tabcolsep}{1pt}
\begin{table*} [t]
\scriptsize
\centering 
\begin{tabular}{@{\hspace{0.5mm}}l  @{\hspace{2mm}} c @{\hspace{0.5mm}} c @{\hspace{0.5mm}}  c @{\hspace{0.5mm}}c@{\hspace{2mm}} c @{\hspace{0.5mm}} c @{\hspace{0.5mm}} c @{\hspace{0.5mm}} c@{\hspace{2mm}}c @{\hspace{0.5mm}} c@{\hspace{0.5mm}} c @{\hspace{0.5mm}}c@{\hspace{2mm}} c @{\hspace{0.5mm}} c @{\hspace{0.5mm}} c @{\hspace{0.5mm}}} 
\toprule 
& \multicolumn{3}{c}{\textbf{Sequence 3 }}&& \multicolumn{3}{c}{\textbf{Sequence 4}}&& \multicolumn{3}{c}{\textbf{Sequence 5}} && \multicolumn{3}{c}{\textbf{ }} \\ 
& \multicolumn{3}{c}{\textbf{(755 images) }}&& \multicolumn{3}{c}{\textbf{(755 images)}}&& \multicolumn{3}{c}{\textbf{(5000 images)}} && \multicolumn{3}{c}{\textbf{ MEAN}} \\ 
\cmidrule{2-4} \cmidrule{6-8} \cmidrule{10-12} \cmidrule{14-16}
  &  PSNR & SSIM & LPIPS && PSNR & SSIM& LPIPS&& PSNR&SSIM&LPIPS&& PSNR&SSIM&LPIPS\\ 
\midrule % In-table horizontal line
Ours w/o AD & 18.00 & 0.793 & 0.314 && 23.27 & 0.918 & 0.110 && 20.34 &  0.812 & 0.205 && 20.54 &0.841& 0.210 \\
Ours w/o EC & 26.70 &  0.940 & 0.092&&  26.53 & 0.922 & 0.078 &&  21.82 & 0.822 & 0.187 && 25.02 & 0.895 & 0.119 \\
Ours-s & 26.88 & 0.940 & 0.103 && 26.21 & 0.928 & 0.090 && 21.22  & 0.817 & 0.190 && 24.77  & 0.895  & 0.128 \\
Ours & 26.95 &  \textbf{0.945} & 0.092 && \textbf{26.63} & \textbf{0.932} & \textbf{0.077} && 21.92 & 0.823 & 0.186 && 25.17 & 0.900 & 0.118 \\ 
Ours w/ LCP & \textbf{27.01} &  0.943 & \textbf{0.091} && 26.48 & 0.920 & 0.084 && \textbf{23.12} & \textbf{0.845} & \textbf{0.162} && \textbf{25.54} & \textbf{0.902} & \textbf{0.112}\\ 
% Ours w/ LCP & 27.91 &  0.938 & 0.092 && 24.12 & 0.883 & 0.101 && 26.84 & 0.932 & 0.191&& 27.22 & 0.931 & 0.187 \\ 

% \midrule
% Ours &  28.20 &  0.954 & 0.062 &&  26.72 & 0.957 & 0.082 && 26.95 & 0.945 & 0.102 && 27.29 & 0.952 & 0.082\\

% \cellcolor{yellow!30}
% \cellcolor{blue!30}
% \cellcolor{red!30}
\bottomrule
\end{tabular}
\vspace{0.5mm}
\caption{Quantitative ablation results. The best results are highlighted in \bf{bold}. } 
\label{tab:ablation} 
\end{table*}

\begin{figure}[t]
\centering
\includegraphics[width=\linewidth]{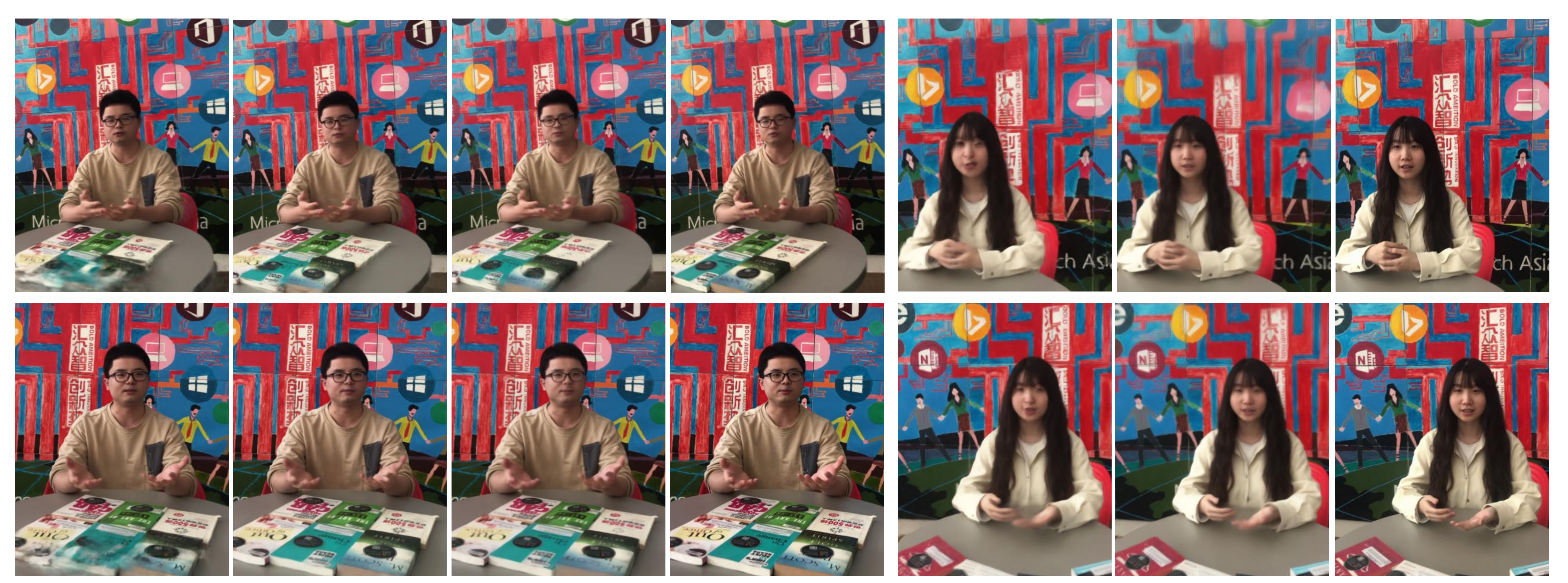}
\put(-296,129){\footnotesize Accurate camera poses}
\put(-126,129){\footnotesize Inaccurate camera poses}
\put(-340,-8){\footnotesize w/o AD}
\put(-281,-8){\footnotesize Ours}
\put(-242,-8){\footnotesize w/ LCP}
\put(-185,-8){\footnotesize GT}
\put(-130,-8){\footnotesize Ours}
\put(-91,-8){\footnotesize w/ LCP}
\put(-35,-8){\footnotesize GT}
\caption{Qualitative results for the ablation study.}
\label{fig:ablation}
% \vspace{-2em}
\end{figure}
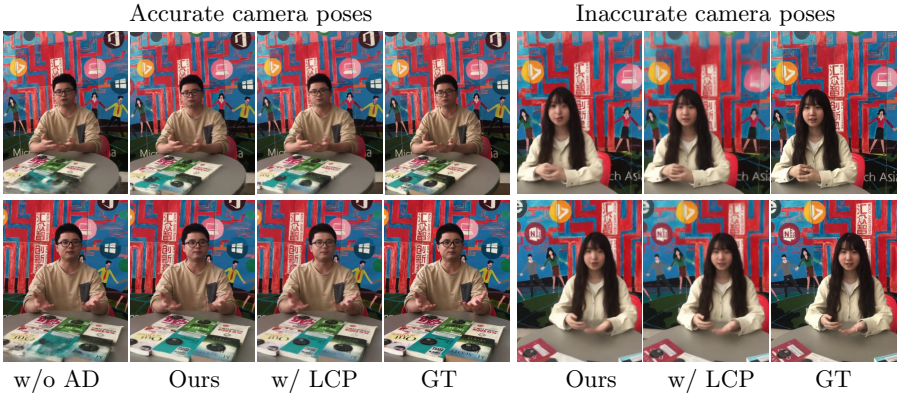

\begin{table} [t]
\footnotesize
\centering 
\begin{tabular}{@{\hspace{1mm}}l @{\hspace{3.5mm}} c @{\hspace{3.5mm}} c @{\hspace{3.5mm}} @{\hspace{3.5mm}} c @{\hspace{3.5mm}} c @{\hspace{3.5mm}} c  @{\hspace{3.5mm}} c  } 
\toprule 

Method  &    NeX & VideoNerf & Nerfies & HyperNerf & Ours & Ours-s     \\ 
\midrule % In-table horizontal line
Inference Time (s) & 3.232  &  12.315   &  18.661 & 19.676 & 0.049 & \textbf{0.031}   \\
FPS & 0.309  &   0.081  &  0.053 & 0.051 & 20.41 & \textbf{32.26}   \\
\bottomrule
\end{tabular}
\vspace{1em}
\caption{Inference speed comparison. The performance is measured on images with $360\times640$ resolution using a single Tesla V100 GPU. 
% For fair comparison, report inference time of  NeX includes both the MPI generation and MPI rendering since dynamic scenes requires to generate different MPIs.
} 
\label{tab:time-size} 
\end{table}

\noindent\textbf{Novel view synthesis} 
% We compare with one state-of-the-art static MPI method NeX-MPI and three dynamic nerf methods Video-NeRF (based on time code), Nerfies (based on deformation) and HyperNerf (based on a combination of learnable time code and deformation). 
We compare our approach with four other scene representation methods. Among them, NeX \cite{wizadwongsa2021nex} is the state-of-the-art MPI-based method. It is proposed to represent only static scenes. The other three are NeRF-variants, which comprise different extensions to handle dynamic scenes. 
In specific, Video-NeRF \cite{li2021neural} conditions NeRF functions on an extra time-variant latent code; Nerfies \cite{park2021nerfies} introduces dynamic deformations into NeRF; HyperNeRF \cite{park2021hypernerf}, on the other hand, combines both the latent code and the deformation into NeRF representation and achieves better performance. Moreover, we also introduce weighted sampling to emphasize the foreground training. Some dataset-specific settings are adopted for the baselines to ensure a fair comparison. We also make sure that images with the same timestamp share the same learned latent code and images from the same camera share the same appearance code. During training, we raise the sampling probability of foreground rays to be $10 \times$ of the sampling probability corresponding to background. We find that employing these settings generally improves the results of baseline methods.

\begin{figure}[t]
\centering
\includegraphics[width=\linewidth]{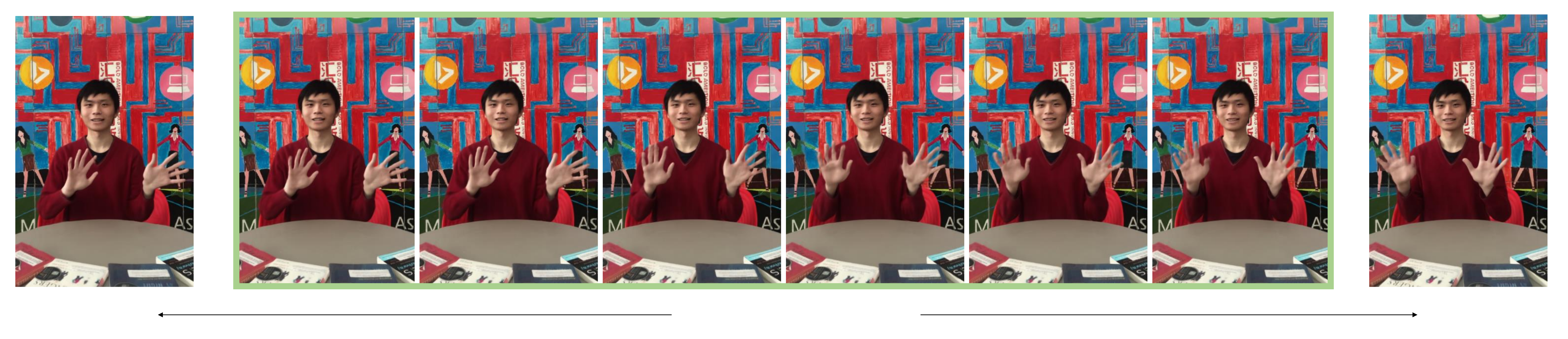}
\put(-194,2){\scriptsize Interpolation}
\put(-338,2){\scriptsize Pose 1}
\put(-32,2){\scriptsize Pose 2}
\put(-210,-10){(a) Interpolation}
\\
%\vspace{-0.5em}
\includegraphics[width=\linewidth]{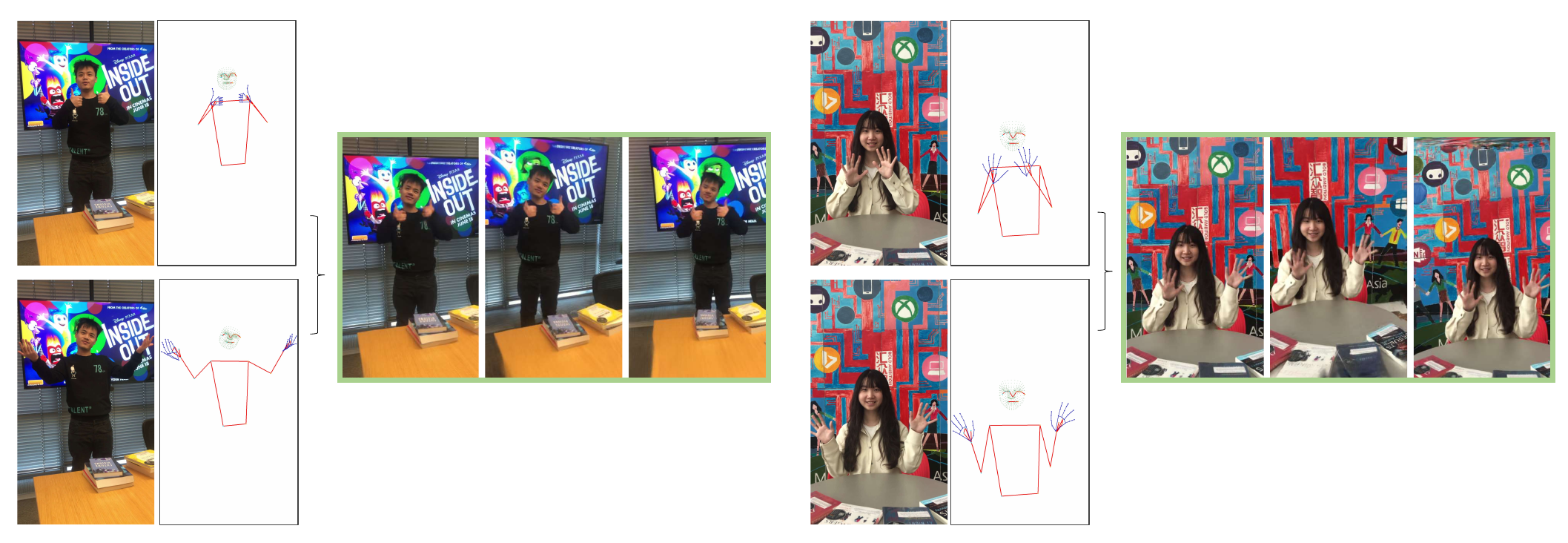}
\put(-220,134){}
\put(-322,119){\scriptsize Pose 1}
\put(-146,119){\scriptsize Pose 1}
\put(-322,-3){\scriptsize Pose 2}
\put(-146,-3){\scriptsize Pose 2}
\put(-264,27){\scriptsize Novel view generation}
\put(-250,18){\scriptsize on new pose}
\put(-90,27){\scriptsize Novel view generation}
\put(-71,18){\scriptsize on new pose}
\put(-232,-16){(b) Novel pose combination}
\\
\vspace{-1em}
\includegraphics[width=0.85\linewidth]{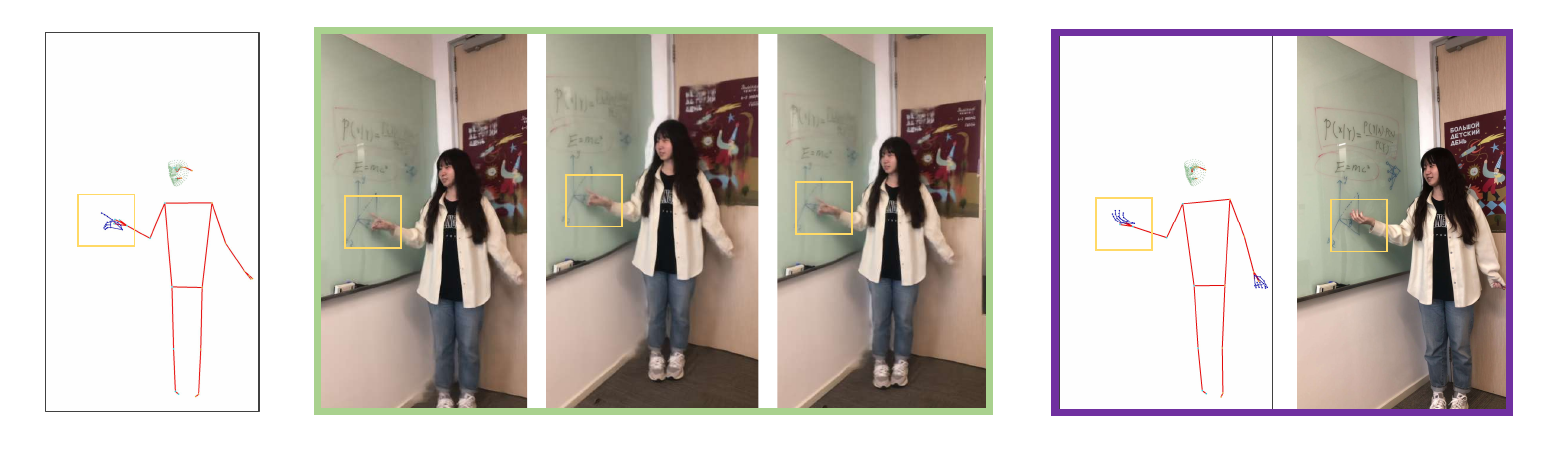}
\put(-215,95){}
\put(-288,1){\scriptsize Unseen pose}
\put(-210,1){\scriptsize Novel view generation}
\put(-103,1){\scriptsize Nearest neighbor in training poses}
\put(-190,-12){(c) Extrapolation}
\caption{Generalization study. Our approach performs well on motion interpolation and novel pose combination, and allows pose extrapolation to some extent. }
\label{fig:control}
\vspace{-1em}
\end{figure}

We illustrate visual comparisons in Fig. \ref{fig:qualitative}. 
Without a dynamic modeling capability, NeX fails to capture motions and generates only blurry bodies and faces. As for Video-NeRF, Nerfies, and Hyper-NeRF, they hardly capture the complex motion of hands since their conditioning input is latent vectors, which are too compressed to represent a detailed pose of a character. In comparison, our method well handles the challenging motions with delicate details.

We also conduct a quantitative comparison using three metrics: LPIPS, MS-SSIM, and PSNR.
We report scores on three testing sequences in Table \ref{tab:metric}.
Our method achieves better SSIM and LPIPS on all sequences but relatively inferior PSNR on Sequences 2 and 3.
We agree with \cite{park2021nerfies} that PSNR is very sensitive to small misalignments between prediction and ground truth 
hence it might not well reflect real perceptual quality under these cases.
% From Fig.~\ref{fig:qualitative}, NeX-MPI is designed for static scenes and thus generate blurry body and face. Video-NeRF, Nerfies and Hyper-NeRF fails to capture the complex motion of hands and generate blurry hands while our model can generate clear details. 
% In addition,  our inference takes only a single forward pass which leads to hundreds times of acceleration than all the baselines as in Table~\ref{tab:time-size} . 
Moreover, we report the inference time of all these methods in Table \ref{tab:time-size}. 
NeRF-base methods rely on a densely sampling procedure, both time and memory-consuming. Instead, our results is inferred with a single forward pass of a CNN, which leads to hundreds of times of acceleration over all the other baseline methods.

% \noindent\textbf{Novel poses} We test our method on three types of novel poses: (i) Interpolation: our model can interpolate a smooth transition between two poses guided by the interpolated pose input as in .~\ref{fig:control}. (ii)Novel combination: benefited from the translational equivalency of the convolution operation, our method can easily generate a never-seen pose by the novel pose combination. As in Fig.~\ref{fig:control}, our method combines two poses and generate the tilted-head man with thump-up gestures and the girl with both hands in the same side. These poses are never observed in the training dataset. (iii) Small extrapolation. our framework can also handle small extrapolation as the trained model is essentially a generative model for a single person.  More results on generating novel poses are attached in the supplement. 
\begin{figure*}[!t]
    \centering 
    \small
    \begin{tabular}{@{}l@{}c@{\hspace{0.5mm}}c@{\hspace{1.5mm}}c@{\hspace{0.5mm}}c@{\hspace{0.5mm}}c@{\hspace{0.5mm}}}
    % \rotatebox{90}{{}}&
     \includegraphics[trim={0 0 0 9cm},clip=true,width=0.15\linewidth]{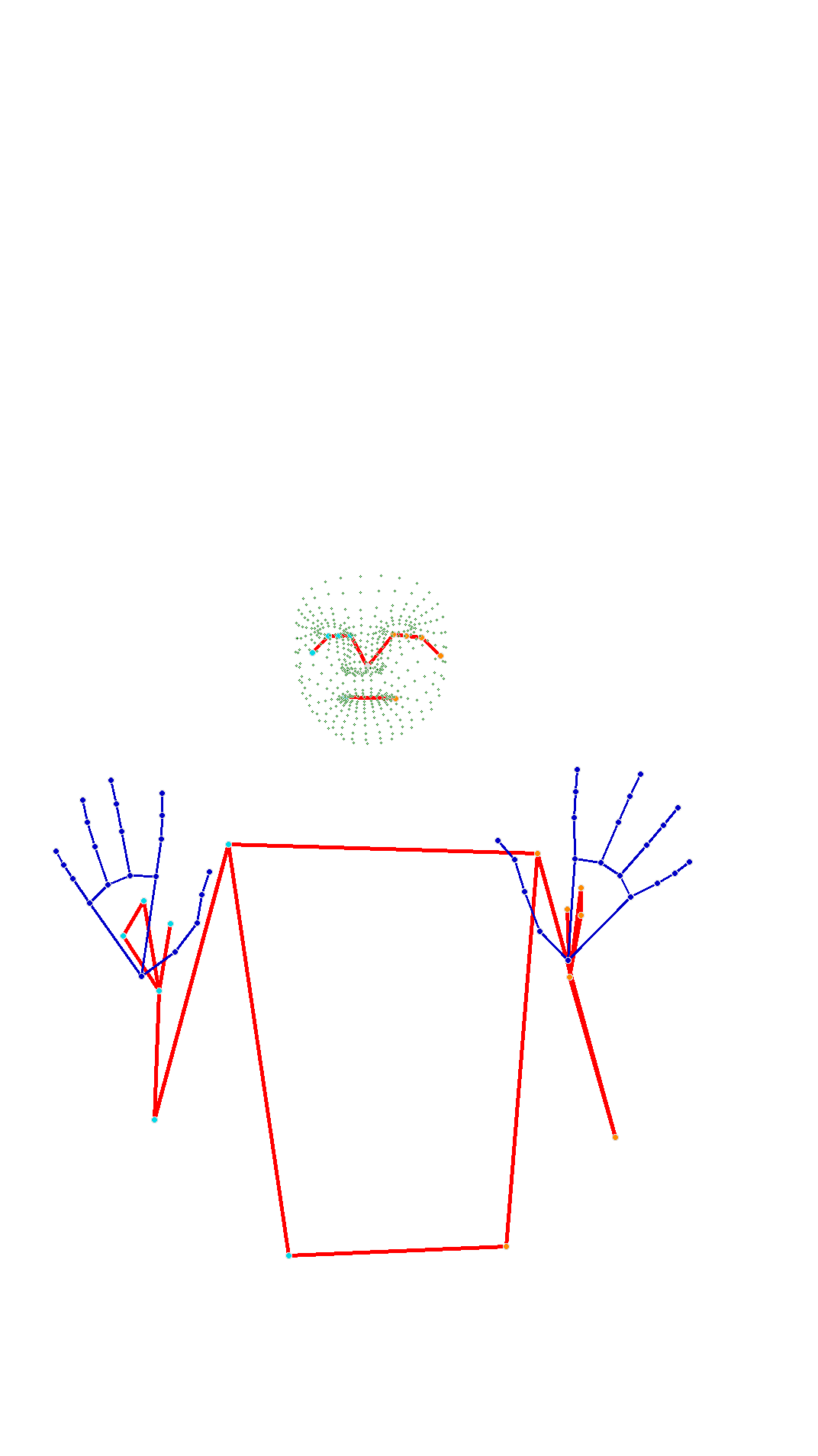} & 
     \includegraphics[trim={0 0 0 9cm},clip=true,width=0.15\linewidth]{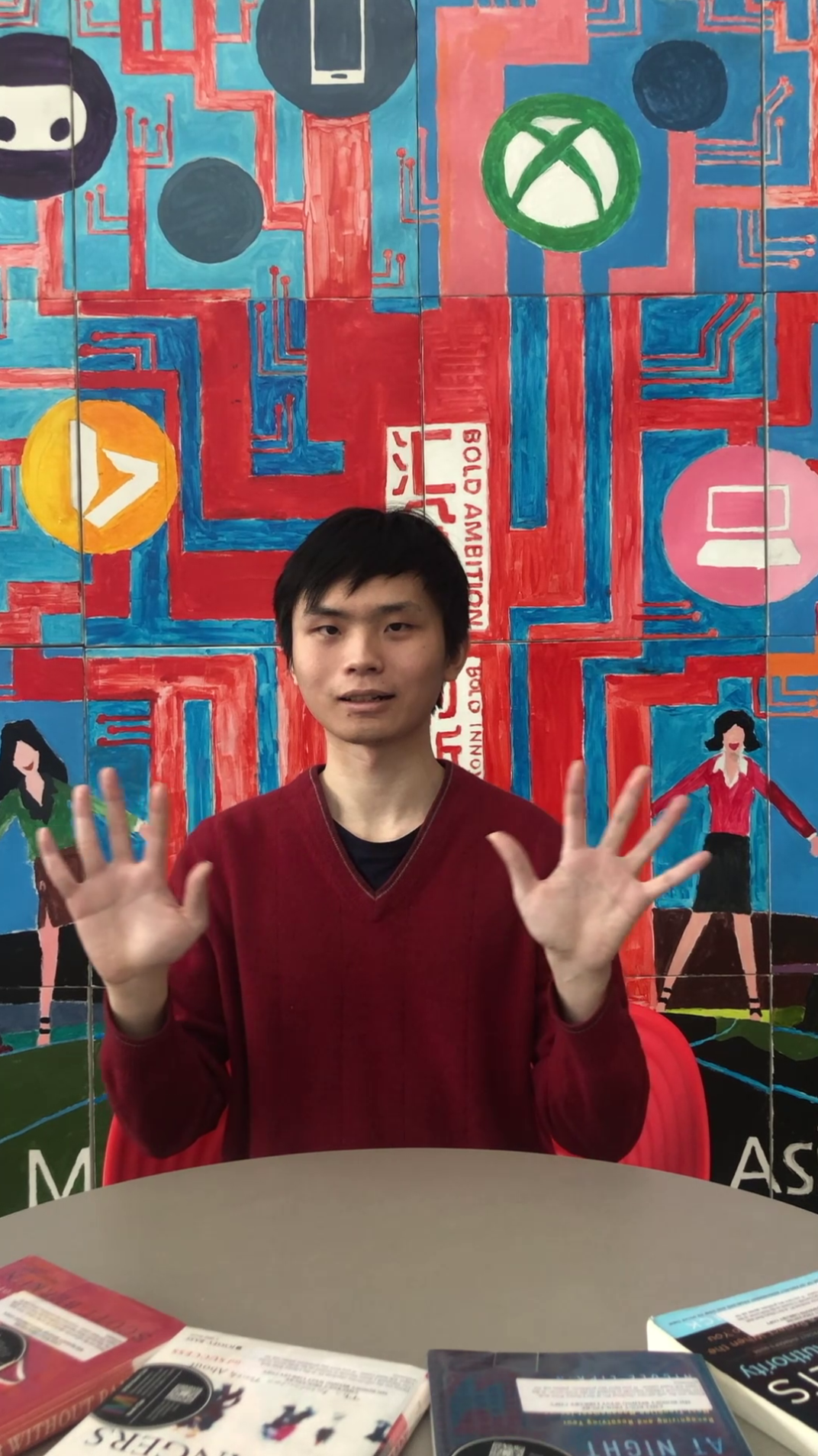} &
     \includegraphics[trim={0 0 0 3cm},clip=true,cfbox=bbox_color 9pt -9pt,width=0.15\linewidth]{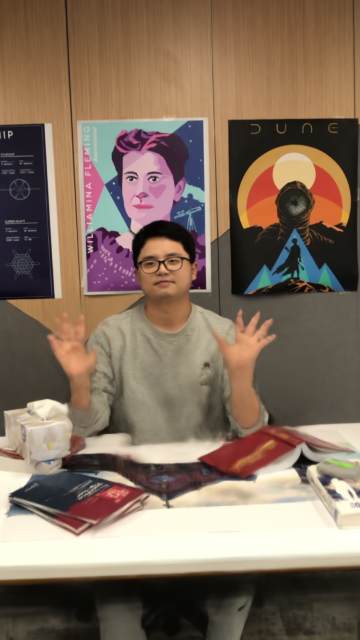} & 
     \includegraphics[trim={0 0 0 9cm},clip=true,width=0.15\linewidth]{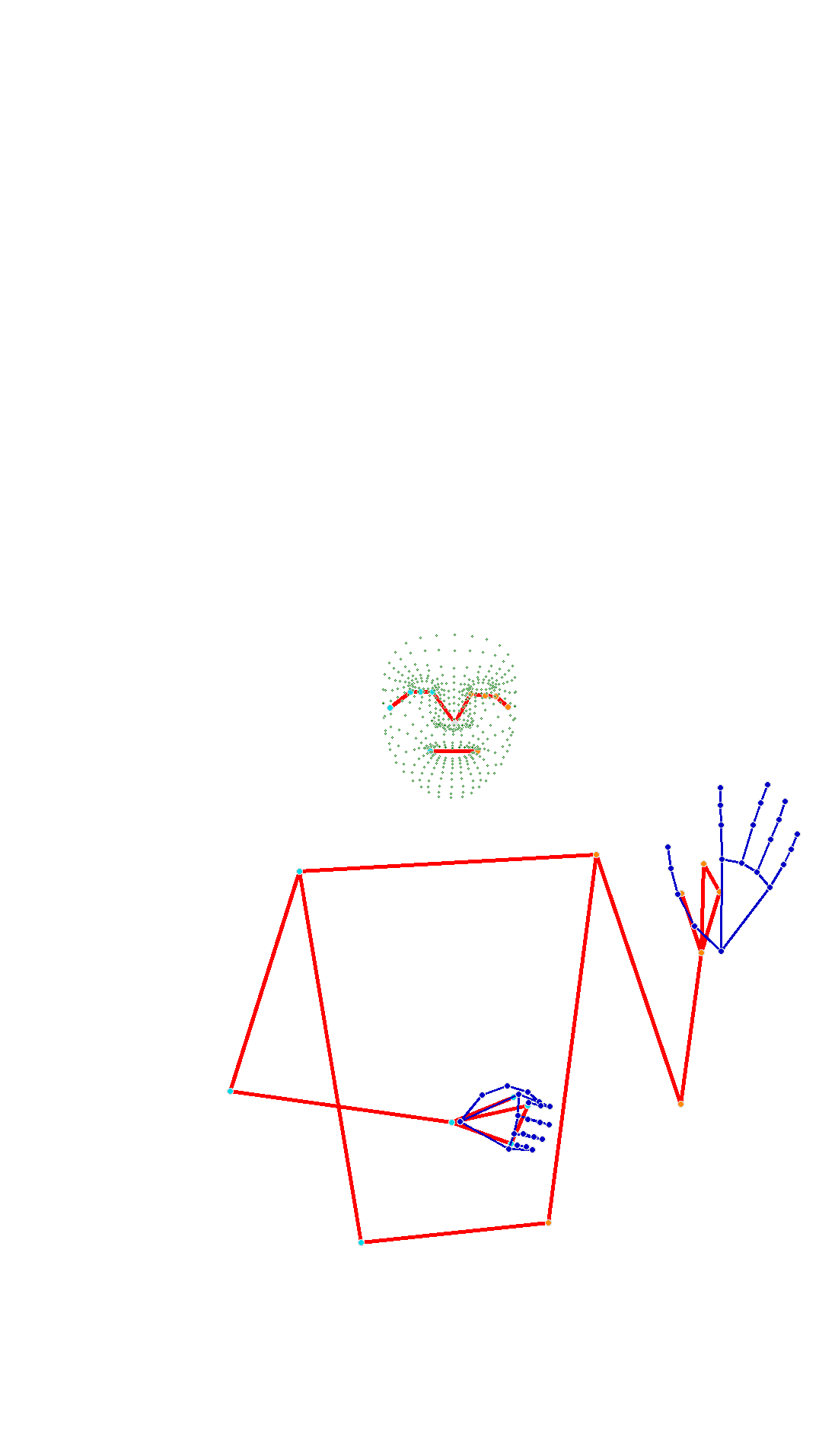}&
     \includegraphics[trim={0 0 0 9cm},clip=true,width=0.15\linewidth]{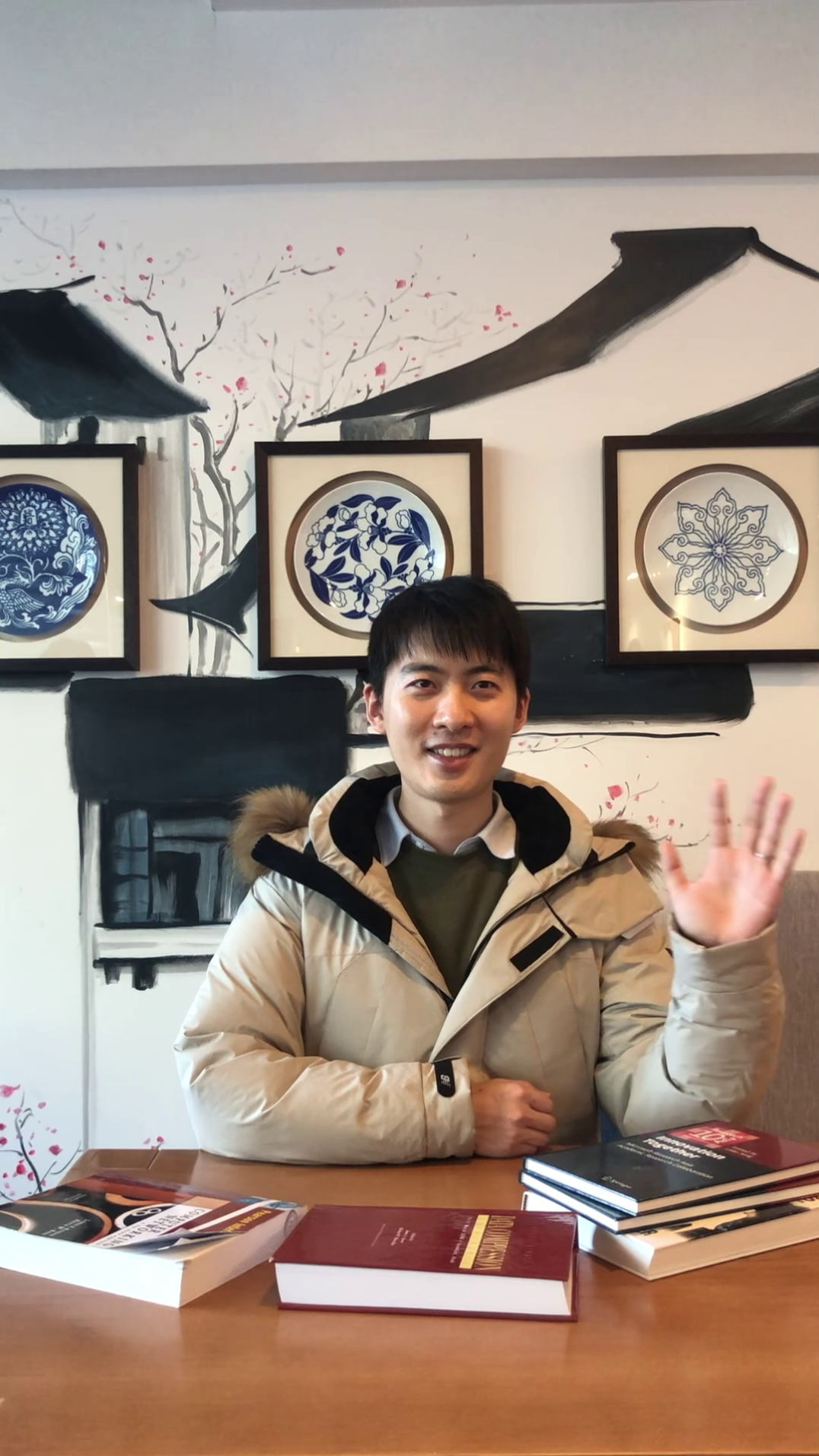}&
     \includegraphics[trim={0 1.5cm 0 1.5cm},clip=true,cfbox=bbox_color 9pt -9pt,width=0.15\linewidth]{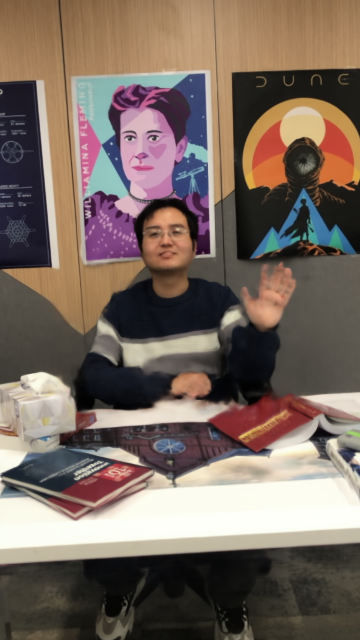}

     \\
     \includegraphics[trim={0 0 0 9cm},clip=true,width=0.15\linewidth]{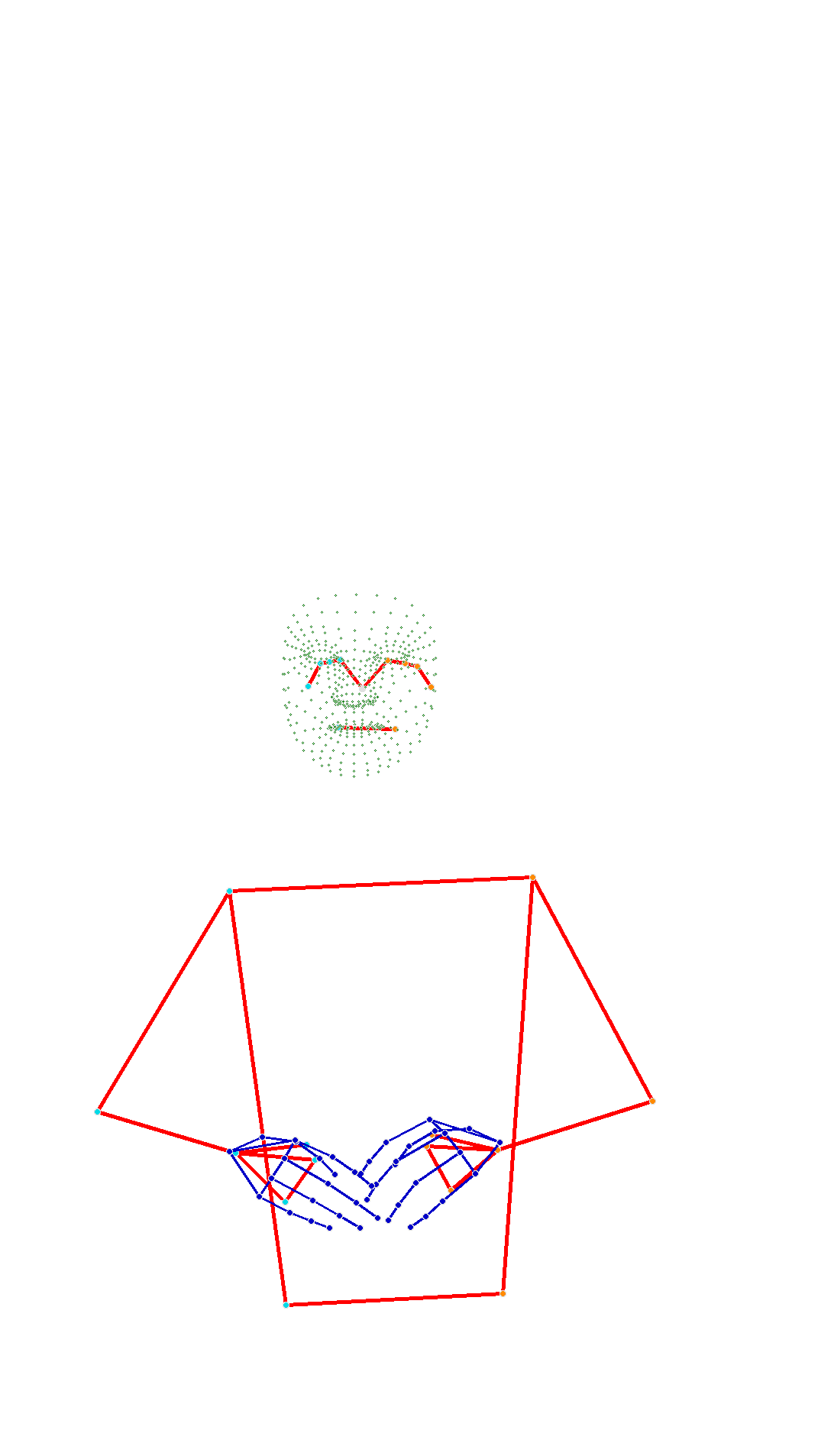}&
     \includegraphics[trim={0 0 0 9cm},clip=true,width=0.15\linewidth]{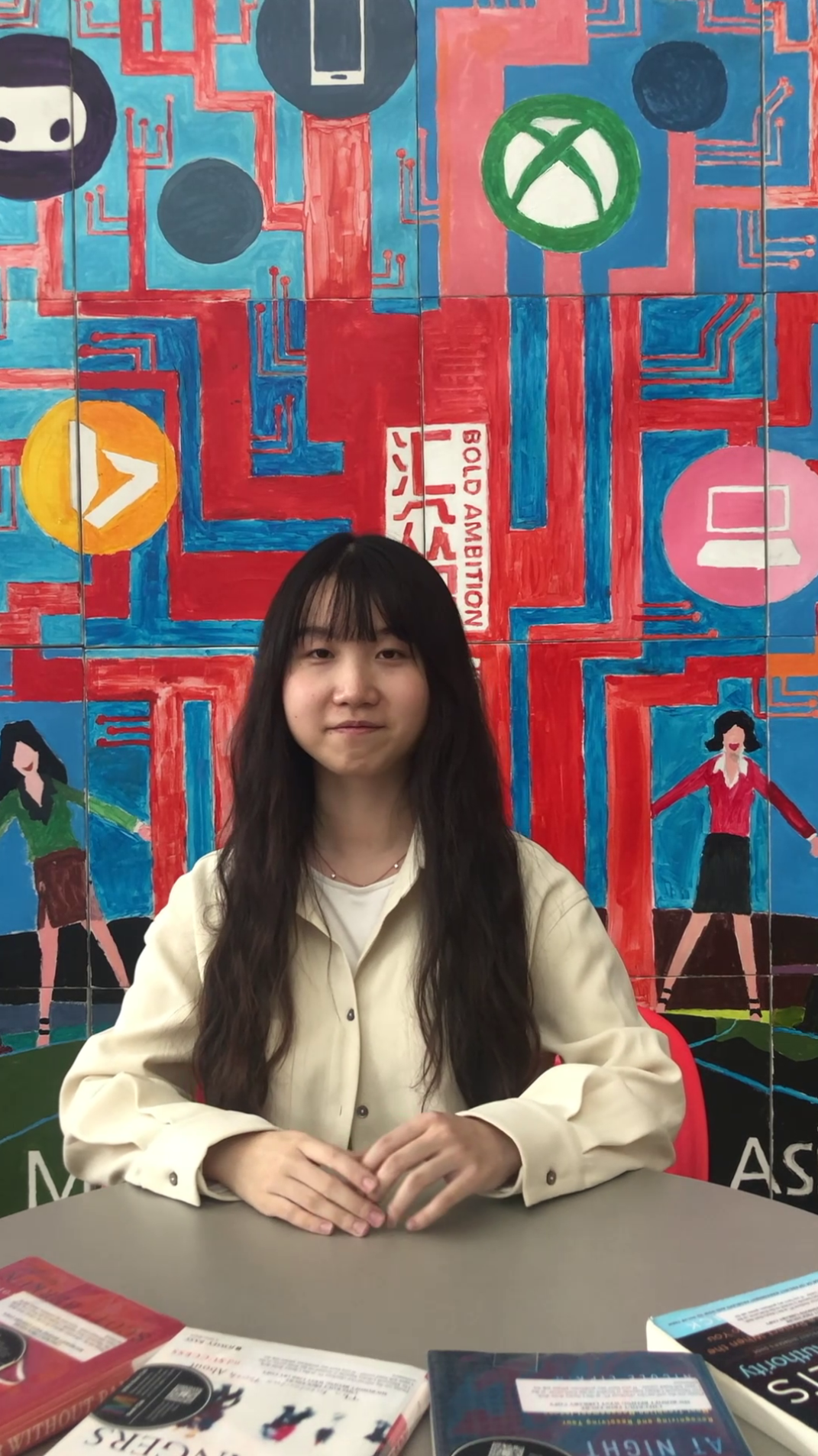}&
     \includegraphics[trim={0 0 0 3cm},clip=true,cfbox=bbox_color 9pt -9pt,width=0.15\linewidth]{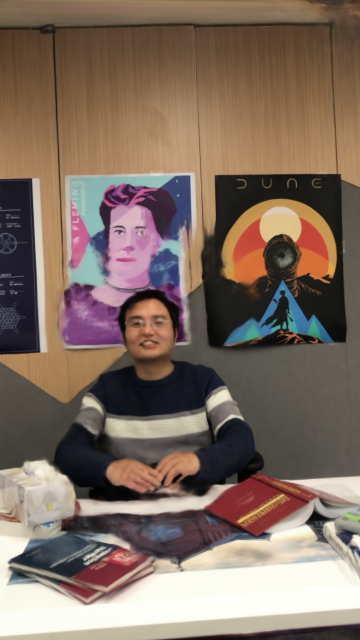}&
     \includegraphics[trim={0 0 0 9cm},clip=true,width=0.15\linewidth]{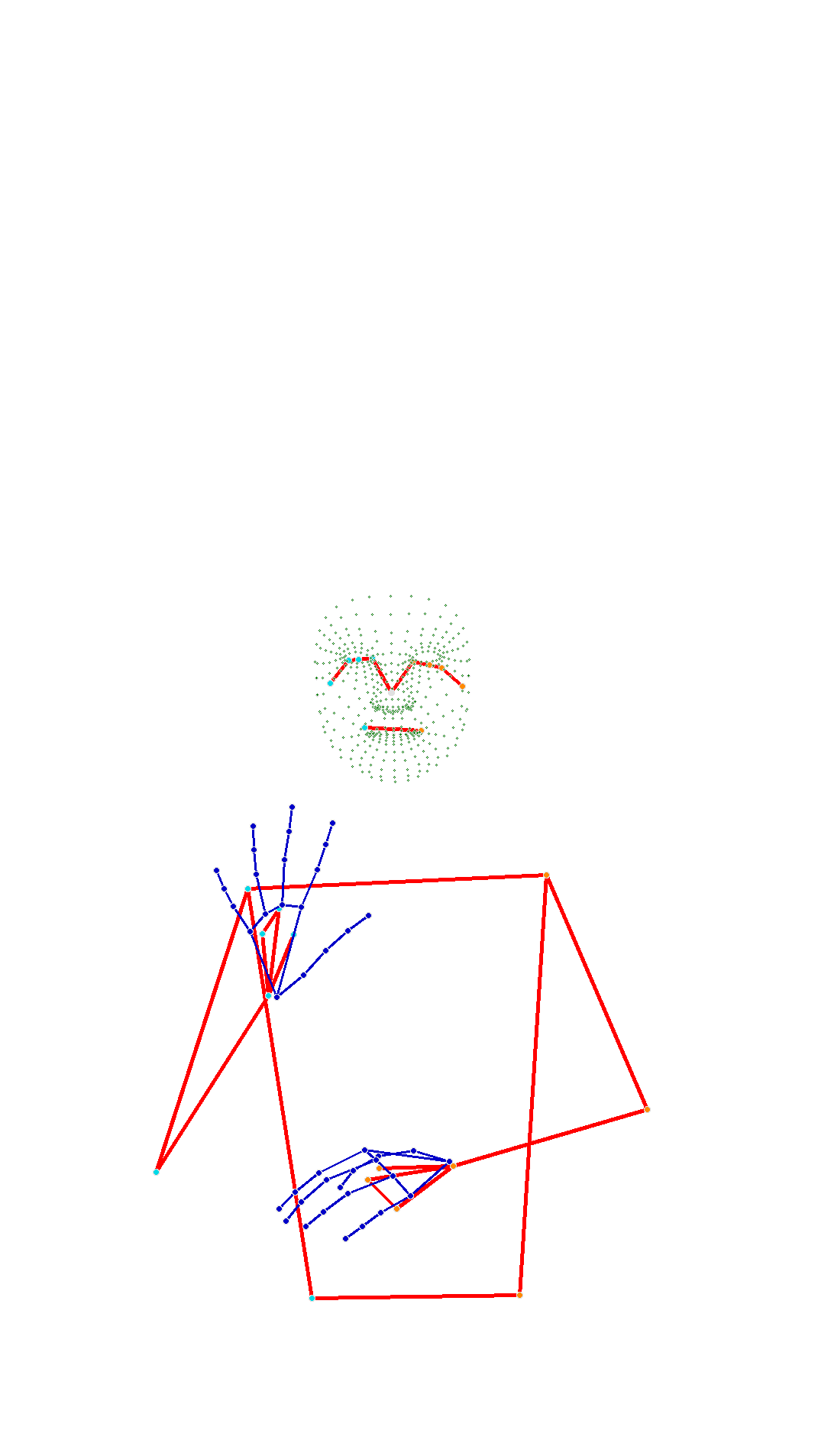}&
     \includegraphics[trim={0 0 0 9cm},clip=true,width=0.15\linewidth]{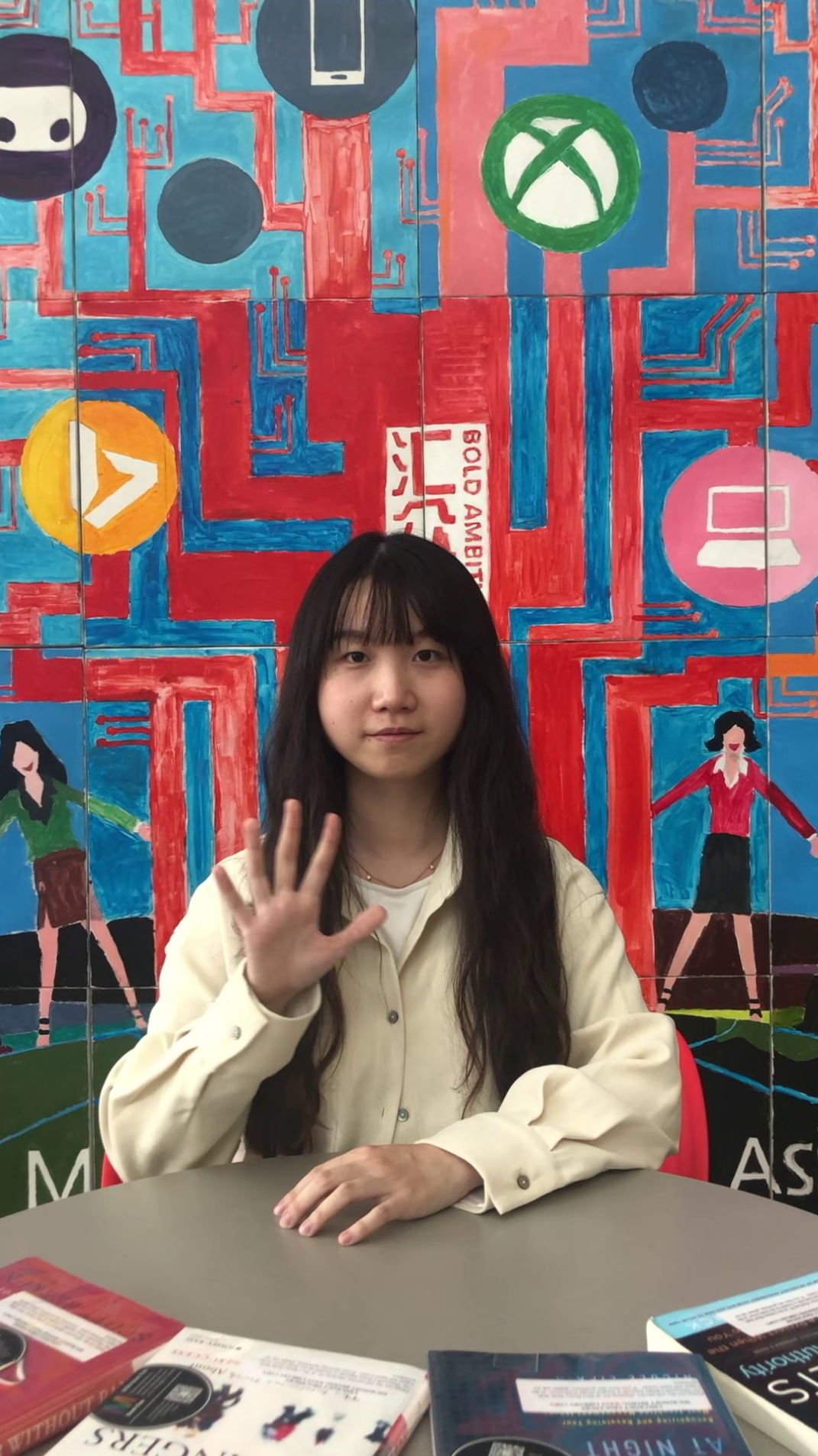}&
     \includegraphics[trim={0 0 0 3cm},clip=true,cfbox=bbox_color 9pt -9pt,width=0.15\linewidth]{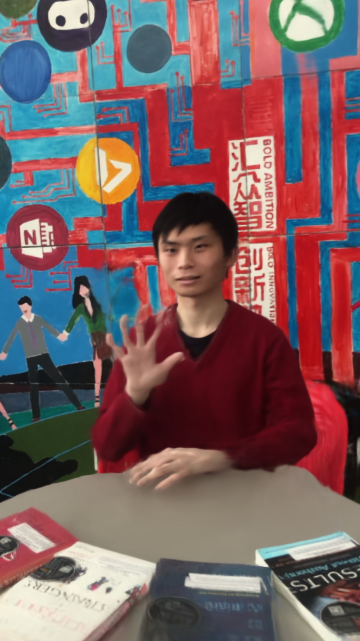}

    \end{tabular}\\
    \vspace{-0.2em}
    \caption{{The generated character (rightmost results) can be driven by the subject with photorealistic quality.}}
    \vspace{1em}
    \label{fig:synchronized}
\end{figure*}

\noindent\textbf{Novel poses} As shown in Fig. \ref{fig:control}, we introduce three settings to illustrate the capability of our method in handling novel poses. 
\emph{Interpolation}: unseen poses are smoothly interpolated between two given poses. As shown by Fig. \ref{fig:control} (a), our method can generate a character that smoothly acts following the interpolated poses.
\emph{Novel combination}: unseen poses are generated by combining different parts of given poses. 
% benefited from the translational equivalency of the convolution operation, our method can easily generate a never-seen pose by the novel pose combination. 
As shown by Fig.~\ref{fig:control} (b), the 3D character generated from our method faithfully combines the two poses, \eg the tilted-head man with thumb-up gestures, the girl with both hands waving to the same side. These poses are never observed in the training dataset. 
\emph{Small extrapolation}: As shown by Fig. \ref{fig:control} (c), our framework can also handle small extrapolation as the trained network is essentially a generative model for a single person.  
Please refer to our supplemental material for more results on novel poses.

% \noindent\textbf{Ablation study} We conduct ablation study to evaluate the effect of the proposed designs. As in Fig.~\ref{fig:ablation}, without the adaptive plane depth, the learning process is sensitive to the initial plane depth assignment and fails to generate the details of the book.  In the cases where the initial camera registration is inaccuate, the learnable pose adjusts the initial camera poses and lead to the obvious improvement in generating clearer faces.  Besides, the exposure code helps to adjust the global luminance level which results in the slight improvement in quantitative metrics. We also test the model using small u-net with less channels and the degradation is acceptable. 

\noindent\textbf{Ablation study} We conduct an ablation study to evaluate the effectiveness of each proposed design. We report scores in Table \ref{tab:ablation}, with results visualized in Fig. \ref{fig:ablation}. AD stands for the ``adaptive depth'', EC means to use the ``exposure coefficients'', while LCP stands for the ``learnable camera poses''. We have the following findings: (i) Without the adaptive plane depth, the learning process is sensitive to the initial plane depth assignment and fails to represent the details of some objects, \eg the book. (ii) In cases where the initial camera registration is inaccurate, the learnable pose adjusts the initial camera poses and leads to obvious improvement in generating clearer faces. (iii) Besides, the exposure code helps adjust the global luminance level, which results in slight improvements in quantitative metrics. (iv) We also test our method with smaller U-Nets (Ours-s in Table \ref{tab:ablation}) with fewer channels and only observe minor quality degradation.

\noindent\textbf{Applications}
% A straightforward application of our framework is to render a synchronized image with same pose for different people as shown in Fig.~\ref{fig:synchronized}. Given the trained model of the target, we transfer the motion from the source and generate the target with a similar pose. Note that our model can also render novel views for the target with the new pose in real-time.  
% A straightforward application of our framework is to create a synchronized video for different people under various scenes as shown in Fig.~\ref{fig:synchronized}. Given all the trained models, we transfer the motion from the same source video and apply to multiple persons. Note that our model can also synchronize the camera path and new views for all subjects. Please see the supplementary videos for more details.   
As a straightforward application, our framework is able to synchronize different 3D characters with the same pose, as shown by Fig. \ref{fig:synchronized}.
With the learned representation of a 3D character, we can extract a 2D pose motion from the driving character and transfer it to the character.
The 3D character can not only be controllable by any given 2D poses but also be rendered to novel views in real time.

\renewcommand{\arraystretch}{.8}
\begin{figure*}[!t]
    \centering 
    \small
    \vspace{-1em}
    \begin{tabular}{@{}c@{\hspace{0.mm}}c@{\hspace{.4mm}}c@{\hspace{0.mm}}c@{\hspace{.4mm}}c@{\hspace{0.mm}}c@{}}
     \includegraphics[width=0.16\linewidth]{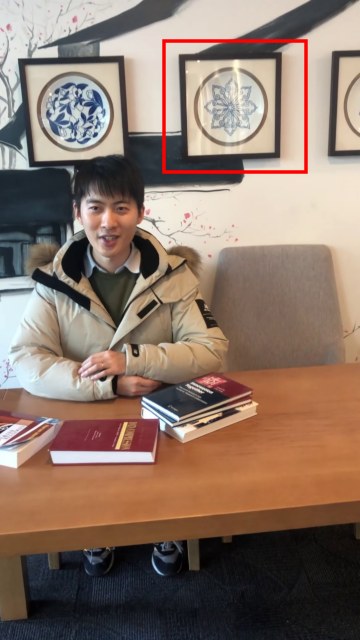}       &   
     \includegraphics[width=0.16\linewidth]{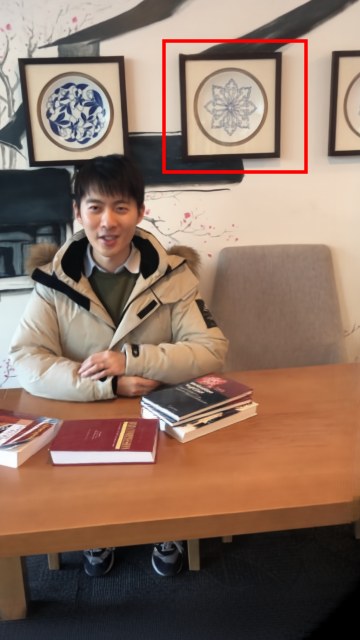}       &        
     \includegraphics[width=0.16\linewidth]{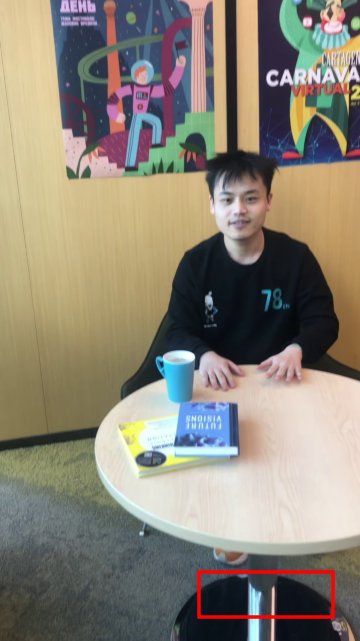}        &   
     \includegraphics[width=0.16\linewidth]{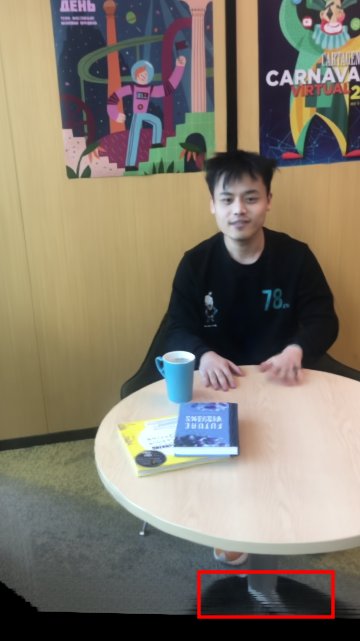} &   
     \includegraphics[width=0.16\linewidth]{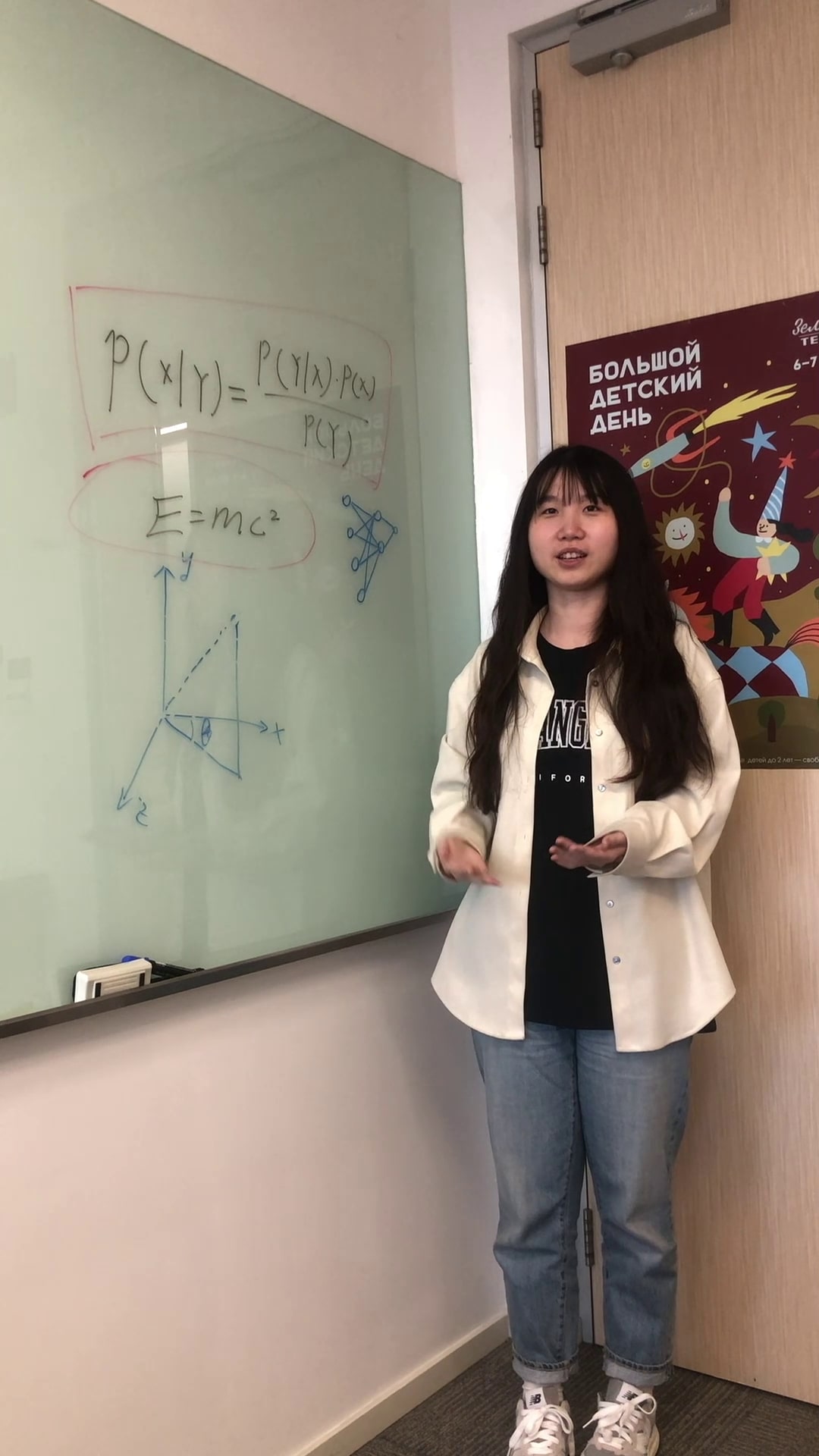} &
     \includegraphics[width=0.16\linewidth]{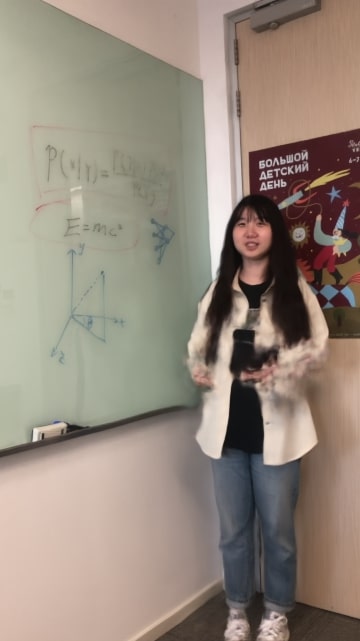}\\
    GT & Ours & GT & Ours & GT & Ours \\
    \end{tabular}
    \vspace{-0.5mm}
   \caption{{Failure cases. Our model fails to generate the reflection effects and may present undesired layered textures. The quality also degrades for large extrapolated motions.}}
    \label{fig:failure}
\end{figure*}

\section{Limitations and Conclusion}
% We study rendering a real-scene 3D character using a pose-guided MPI. Extensive experiments show that our approach real-time in rendering high-quality novel views for different human poses. However, there are certain limitation as shown in \ref{fig:failure}. Our model cannot generate view-dependent effects and suffers from the typical MPI artifacts such as the obvious layers near the boundary. Although our model is able to generalize new pose fairly well from the training data, the quality of extrapolation still degrades (e.g. details of hands are missing). Future work could focus on improving the quality of training data or integrate the view-dependent component in an efficient way.

This work studies a controllable representation of a real-scene 3D character based on a pose-guided MPI. 
Extensive experiments show that our approach is able to represent a 3D character in real scenes with unprecedented quality, which can be not only rendered into novel views in real-time speed but also be controllably guided by different human poses. 
Still, our method suffers from some artifacts as shown in Fig.~\ref{fig:failure}.
Our MPI representation only outputs diffuse colors, and the rendered images thus cannot handle view-dependent effects such as the reflections.
Our method also suffers from typical MPI artifacts, such as the layered section exposed near the boundary.
Even though our framework can generalize to some new unseen poses fairly well, the quality of synthesis would still degrade, especially when the new pose deviates too far away from the training data. 
The proposed method is promising to serve as a practical solution for character rendering and we expect further works to solve the remaining issues.

% We explore the possibility of learning a personalized real-scene 3D controllable human avatar and approach this problem using conditional MPI. Extensive experiments show that our approach is fast in inferencing new pose and real-time in rendering high-quality novel views. Howerver,  as our data capture process relies on camera registration, the inaccurate camera pose limits the quality of the training data.  Although our model is albel to generalize new pose fairly well from the training data, the quality of extrapolation still degrades (e.g. details of hands are missing). Future work could focus on improving the quality of training data in two ways: achieving more accurate camera poses for extremely long video sequences and finding more effective training poses.  

\clearpage
\bibliographystyle{splncs04}
\bibliography{egbib}

\clearpage

\appendix 
\section*{\LARGE Appendix}
\section{Implementation Details}
\subsection{Network structure}
Our network structure adopts the encoder-decoder structure following U-Net~\cite{ronneberger2015u}, which has been proved very effective for conditional image generation. The encoder consists of a series of residual blocks, gradually reducing the feature spatial size and doubling the number of channels. From the latent feature, the decoder recovers the feature spatial size with skip connection from the features of the encoder. The number of features of the first block is $64$ and we adopt $5$ blocks for the encoder or decoder respectively. The final layer of the decoder is a $1\times1$ convolutional layer that outputs the multiplane images (MPIs). 
We adopt a texture sharing strategy following Nex~\cite{wizadwongsa2021nex} to achieve a more compact MPI, where every $4$ plane shares the same RGB textures. Specifically, the network outputs $240$ channels, where the first $192$ channels represent the alpha channels and the remaining $48$ channels are the RGB textures. 

Also, we introduce a small U-Net structure where we make the following modifications: 1) Reduce the number of feature channels to $1/3$ of the baseline. 2) Remove one residual block for encoder and decoder respectively. As verified in the quantitative study, this smaller network only introduces minor quality degradation but enables real-time rendering.

\subsection{Data preprocess}
After we capture the multi-camera video sequences, we perform the following data processing for training

\subsubsection{Synchronization} Our pose-guided MPI synthesis requires  temporally synchronized driving frame and multi-views. We use audio to synchronize the videos from multiple cameras. We utilize a commercial software, Adobe Premiere, to synchronize the videos according to their audio tracks, which offers synchronization accuracy within a few tens milliseconds to five milliseconds. Because there inevitably exists misalignment, we ask the character to move a little bit slower than usual to reduce the misalignment between frames. Instead of processing the whole video sequence, we divide the videos into overlapping clips and synchronize the video clips, which we find leads to more accurate alignment. 

\subsubsection{Video clip and frame extraction} We capture the character for $1\sim3$ minutes and encourage the character to perform diverse poses during the video capture. We subsample the video frames by $3\sim 10$ fps depending on the clip length so that we obtain roughly $1000$ frames (or $\sim 200$ different human poses) for training. It is noteworthy that our network has the potential to fit more frames and we expect this will lead to improved character modeling. However, leveraging more training frames from the moving capture rig is computationally prohibitive in the SfM stage and we intend to address this in our future work.

\subsubsection{Camera pose estimation from SfM} We use COLMAP~\cite{schonberger2016structure} to estimate the camera poses. For all the sequences, we first parse the foreground person using~\cite{li2020self} along with mask dilation to exclude the moving regions during the feature matching. For video sequences with less than $1000$ frames, we use the exhaustive matcher with the guided feature match and run the standard mapper for the scene construction. When dealing with more than $1000$ frames, we use vocabulary tree matching for feature matching. Then we adopt a hierarchical mapper followed by a few iterations of triangulation and bundle adjustment to reconstruct the scene. Since we use smartphone cameras of the same type, we enforce shared camera intrinsics during the COLMAP computation. After the sparse reconstruction, we apply the undistortion operation on the original image based on the estimated distortion parameter. These undistorted images are utilized as the ground truth.

\subsubsection{Keypoints Extraction} We use MediaPipe~\cite{lugaresi2019mediapipe} to extract the holistic human keypoints for the driving frames. Specifically, we obtain $468$, $33$ and $42$ landmarks for face, body and hands respectively. After the detection, we draw the extracted keypoints with a pre-defined color scheme. As MediaPipe does not support frames with multiple persons, for such cases we crop the input image, detect the keypoints for each person individually and finally stitch the results altogether. 

% \subsection{SfM settings}
\subsection{Motion transfer}
\begin{figure*}[h]
    \centering
    \includegraphics[width=1.0\linewidth]{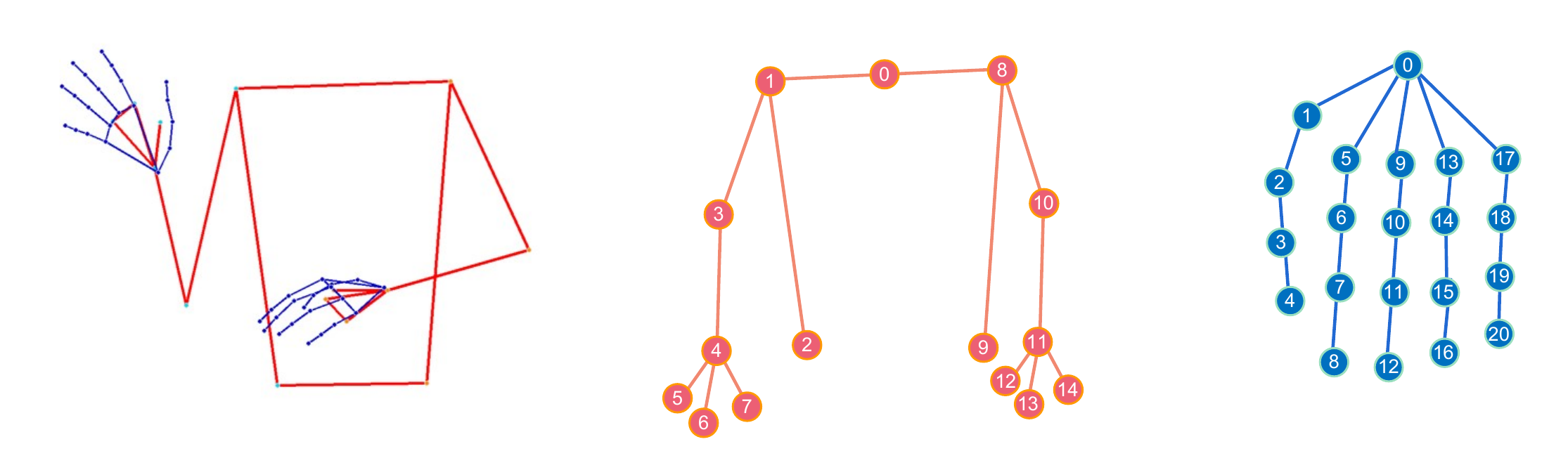}
    \put(-275,-2){(a)}
    \put(-158,-2){(b)}
    \put(-45,-2){(c)}
    %\vspace{-0.6cm}
    \caption{The illustration of the body tree structure and hand tree structure. (a) Captured body and hand pose. (b) The body tree structure where node 0 is the tree root. (c) The hand tree structure where node 0 is the tree root. }
    \label{fig:pose_tree}
\end{figure*}

When trying to transfer the body pose of the driving character to the source character, we need to keep the direction of driving limbs and the limb length of the source. To achieve this goal, we treat the body landmarks as a tree structure, as shown in Fig.~\ref{fig:pose_tree}(b). The tree root is the midpoint of the left shoulder and right shoulder. When the location of the parent node is known, we can calculate its children nodes. We use  $t_b^c$ to denote the children node of the  driving body and $t_b^p$ for the parent node of the driving body. The transferred children node is 
\begin{equation}
\frac{t_b^c - t_b^p}{l_b^t} * l_b^s,
\end{equation}
where $l_b^t$ is the length of the driving body limb and $l_b^t$ is the length of the corresponding source limb. And the limb length is determined as the maximum limb length of all the driving/source frames. The gesture transfer is similar to body pose transfer, except that the tree root is changed to the wrist. Fig.~\ref{fig:pose_tree}(c) depicts the tree structure for fingers.

\section{Generalization Ability of NeRF-based Approaches}
We also explore the generalization ability of NeRF-based approaches. To achieve this, we modify the state-of-the-art method Nerfies~\cite{park2021nerfies} and HyperNerf~\cite{park2021hypernerf} with the keypoints as input. Specifically, for Nerfies, instead of learning a latent code for the deformation of each pose, we directly use the body keypoints as the input to learn the deformation. For HyperNeRF, we modify it to a hybrid way, where we expect the input body keypoints to learn the large deformation --- leg movement or head pose --- relative to the canonical template  while the hyper code modulates the canonical space and accounts for small deformation such as expression changes. 

\begin{figure*}[!t]
    \centering 
    \small
    \begin{tabular}{@{}c@{\hspace{0.2mm}}c@{\hspace{.2mm}}c@{\hspace{0.2mm}}c@{\hspace{.2mm}}c@{\hspace{0.2mm}}c@{}}
     \includegraphics[width=0.16\linewidth]{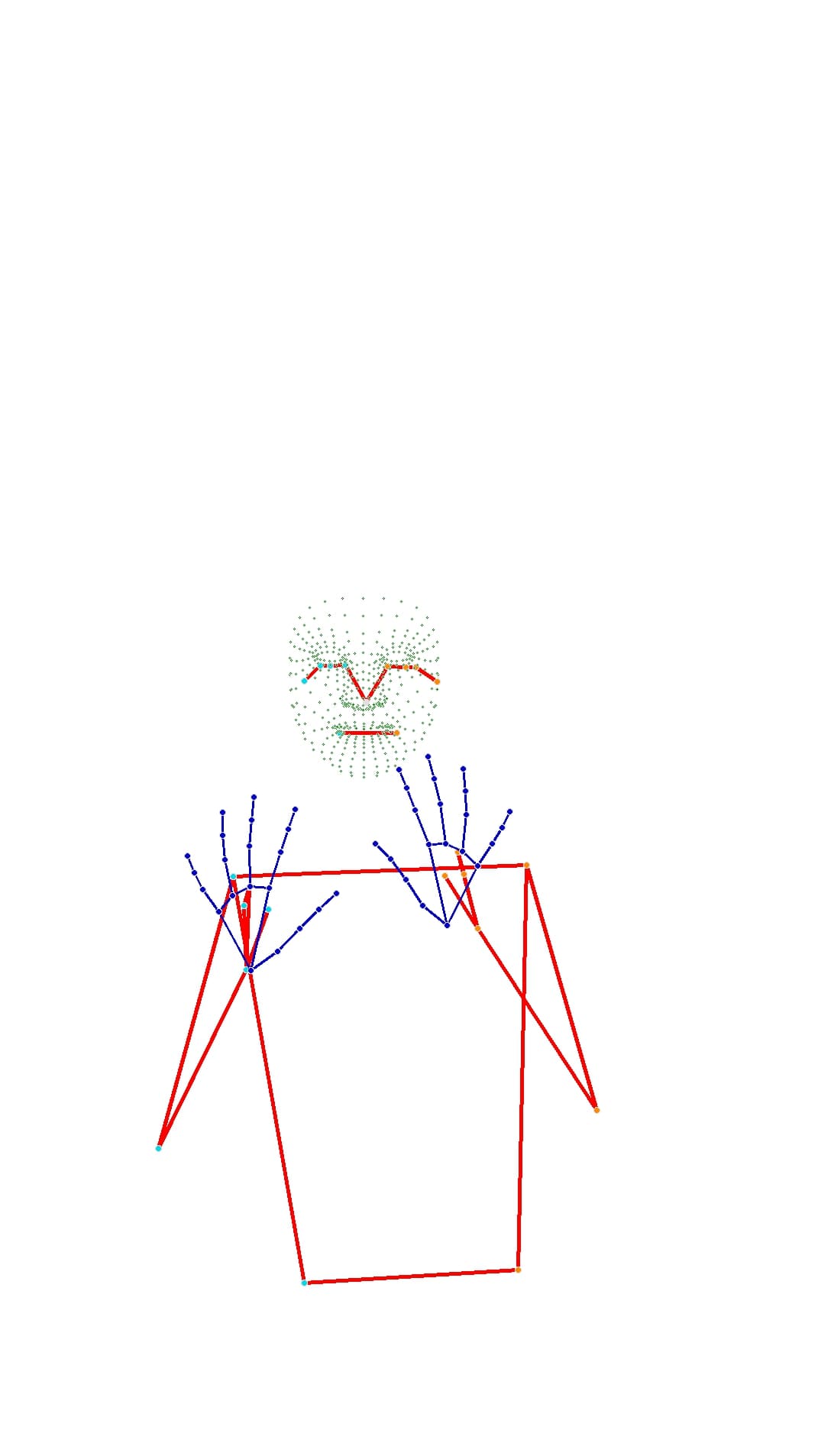}       &   
     \includegraphics[width=0.16\linewidth]{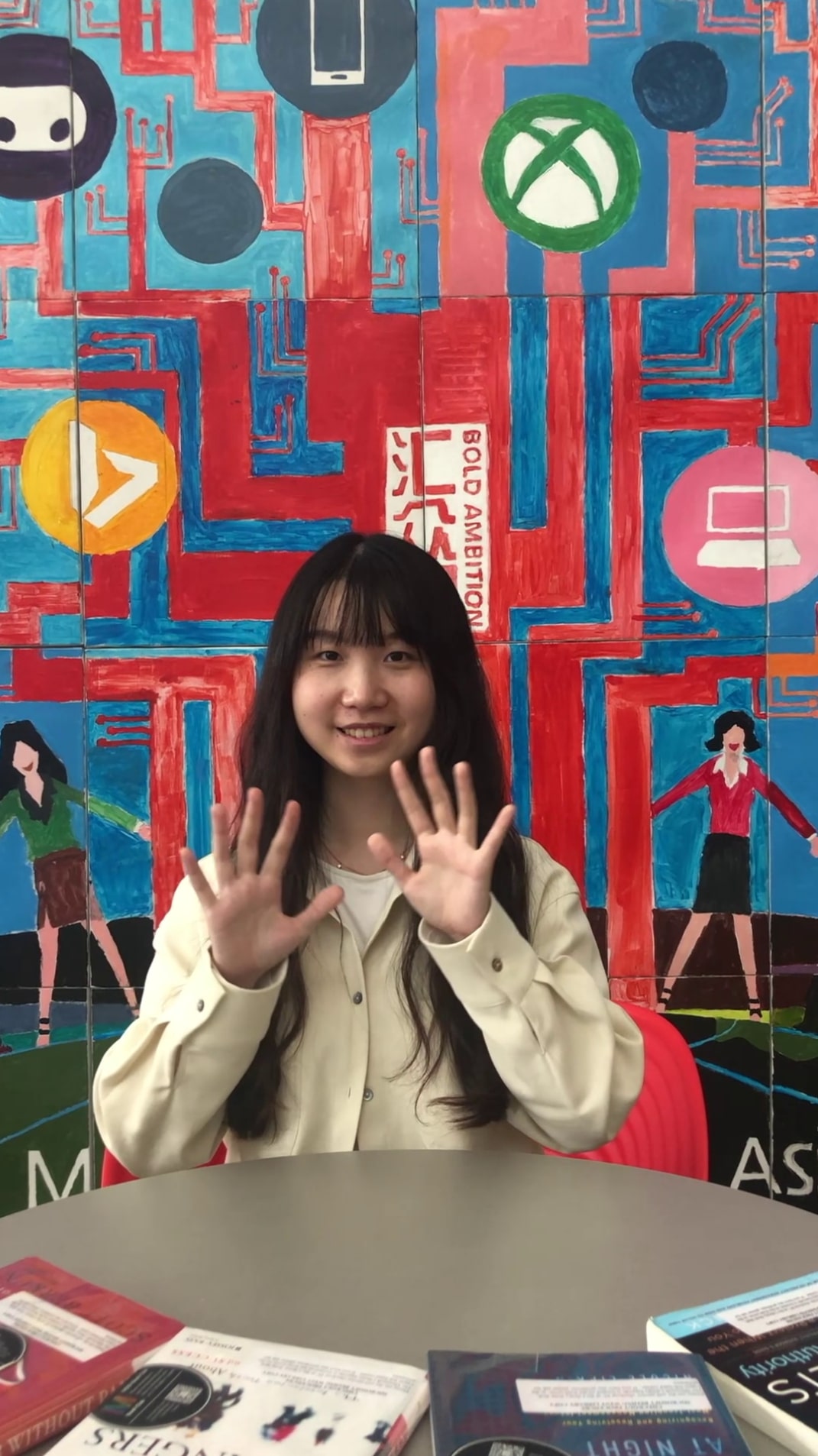}       &        
     \includegraphics[width=0.16\linewidth]{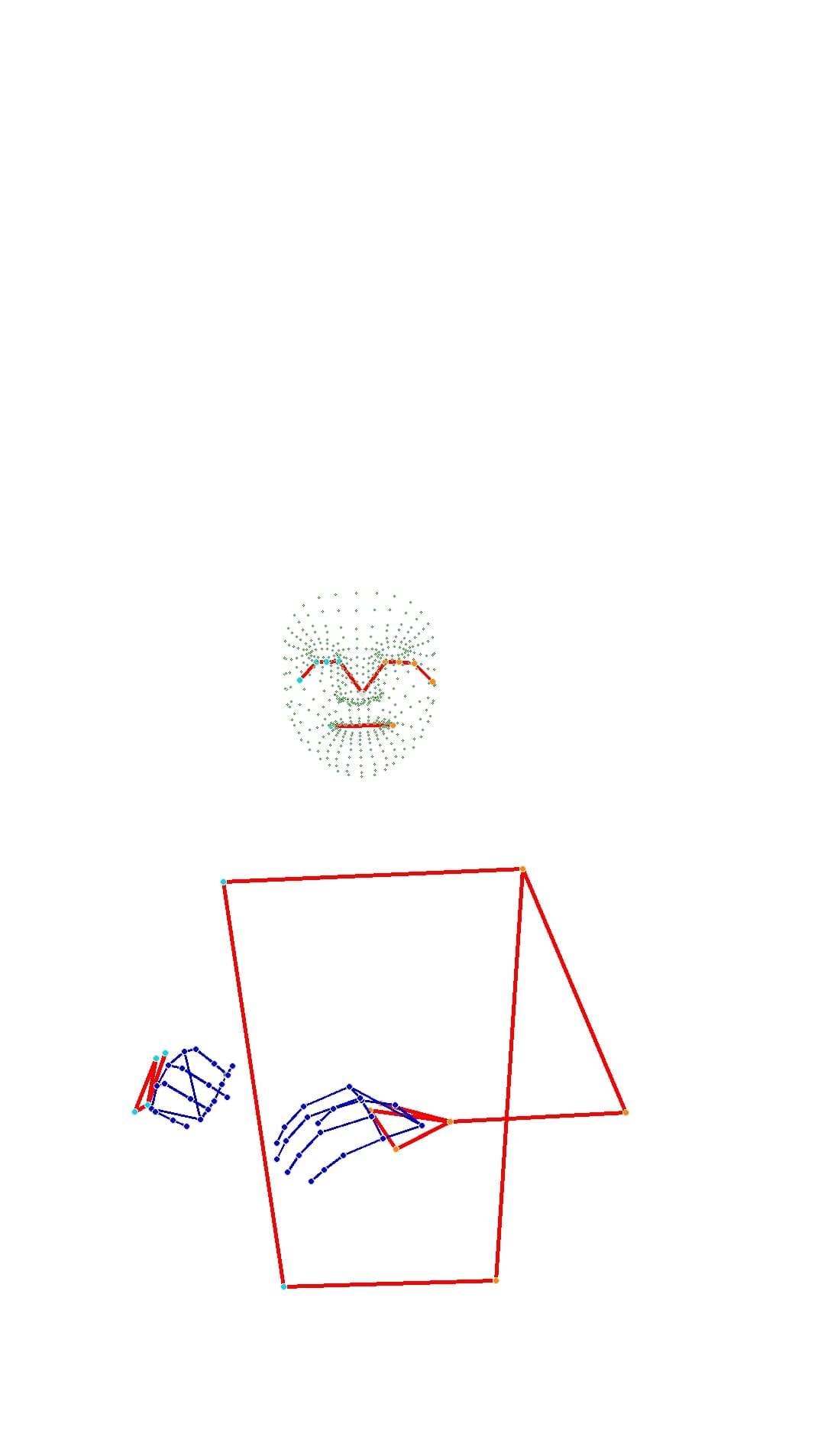}        &   
     \includegraphics[width=0.16\linewidth]{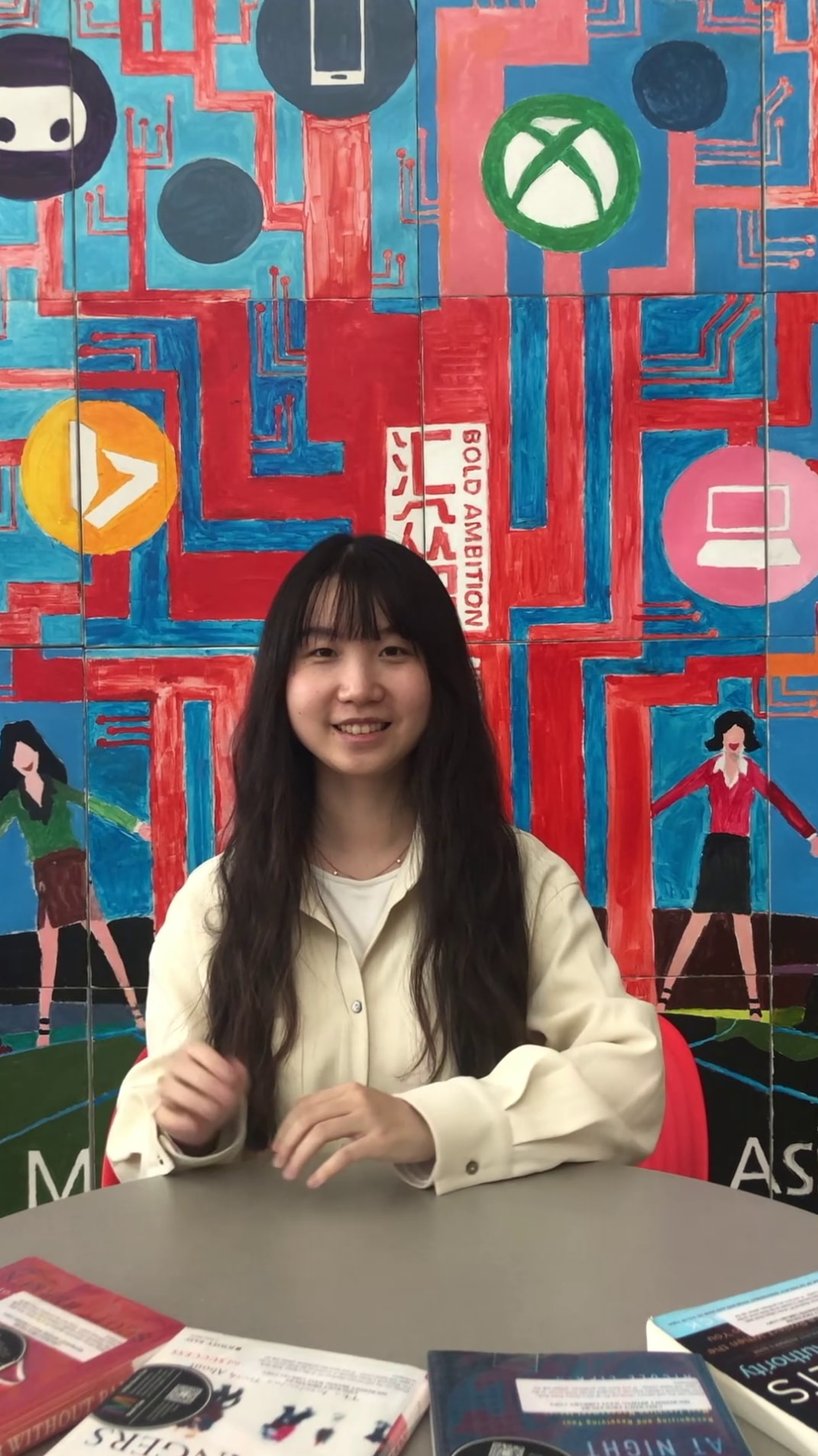} &   
     \includegraphics[width=0.16\linewidth]{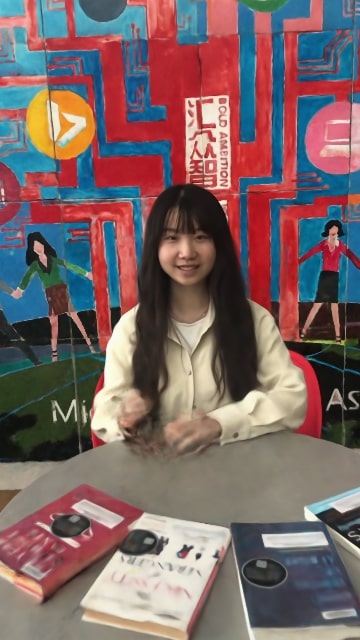} &
     \includegraphics[width=0.16\linewidth]{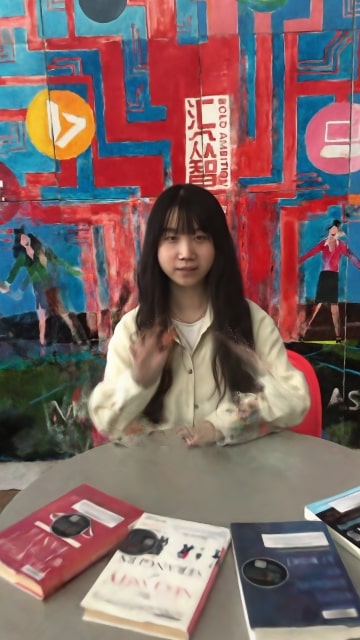}\\
     \includegraphics[width=0.16\linewidth]{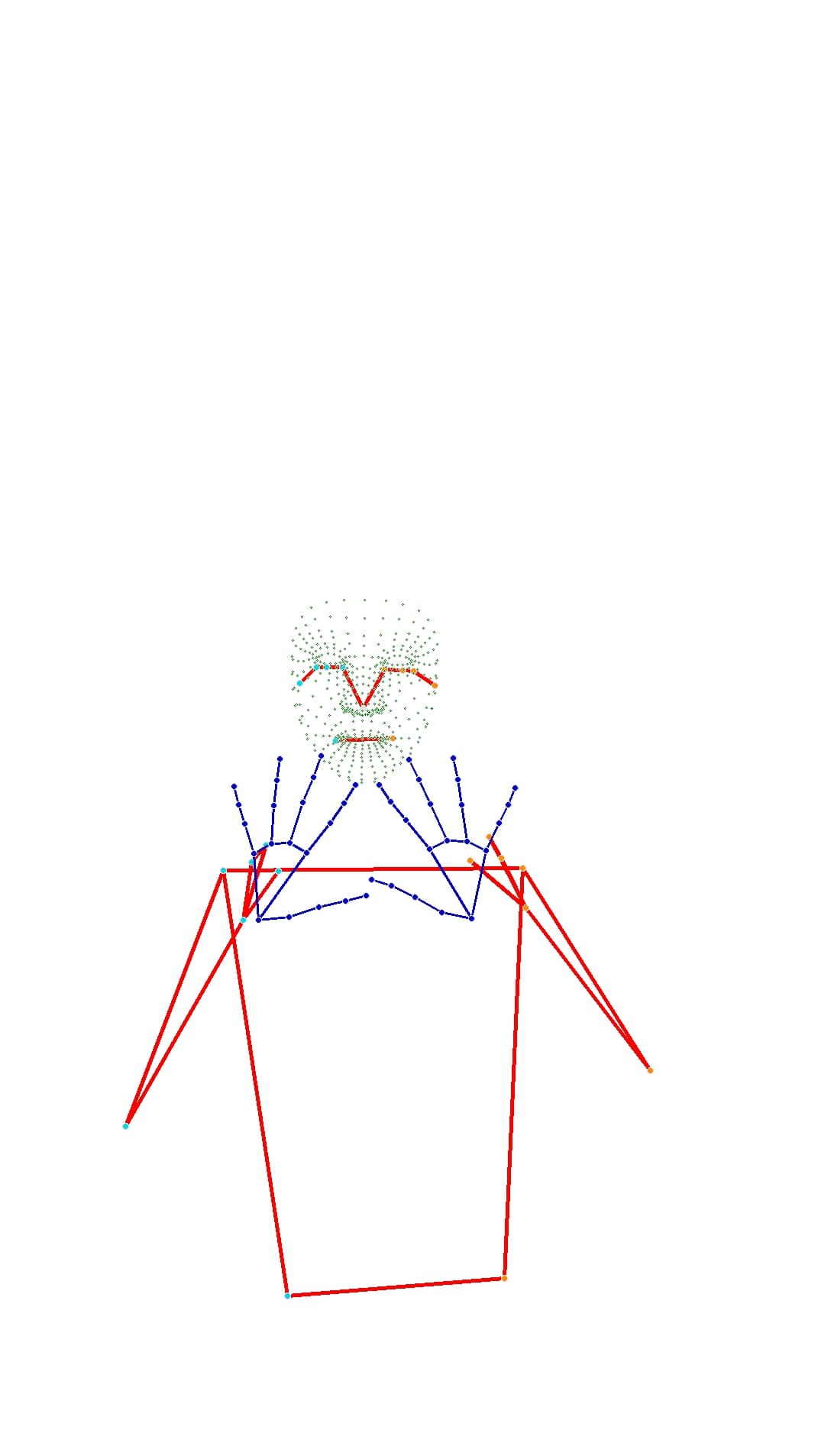}       &   
     \includegraphics[width=0.16\linewidth]{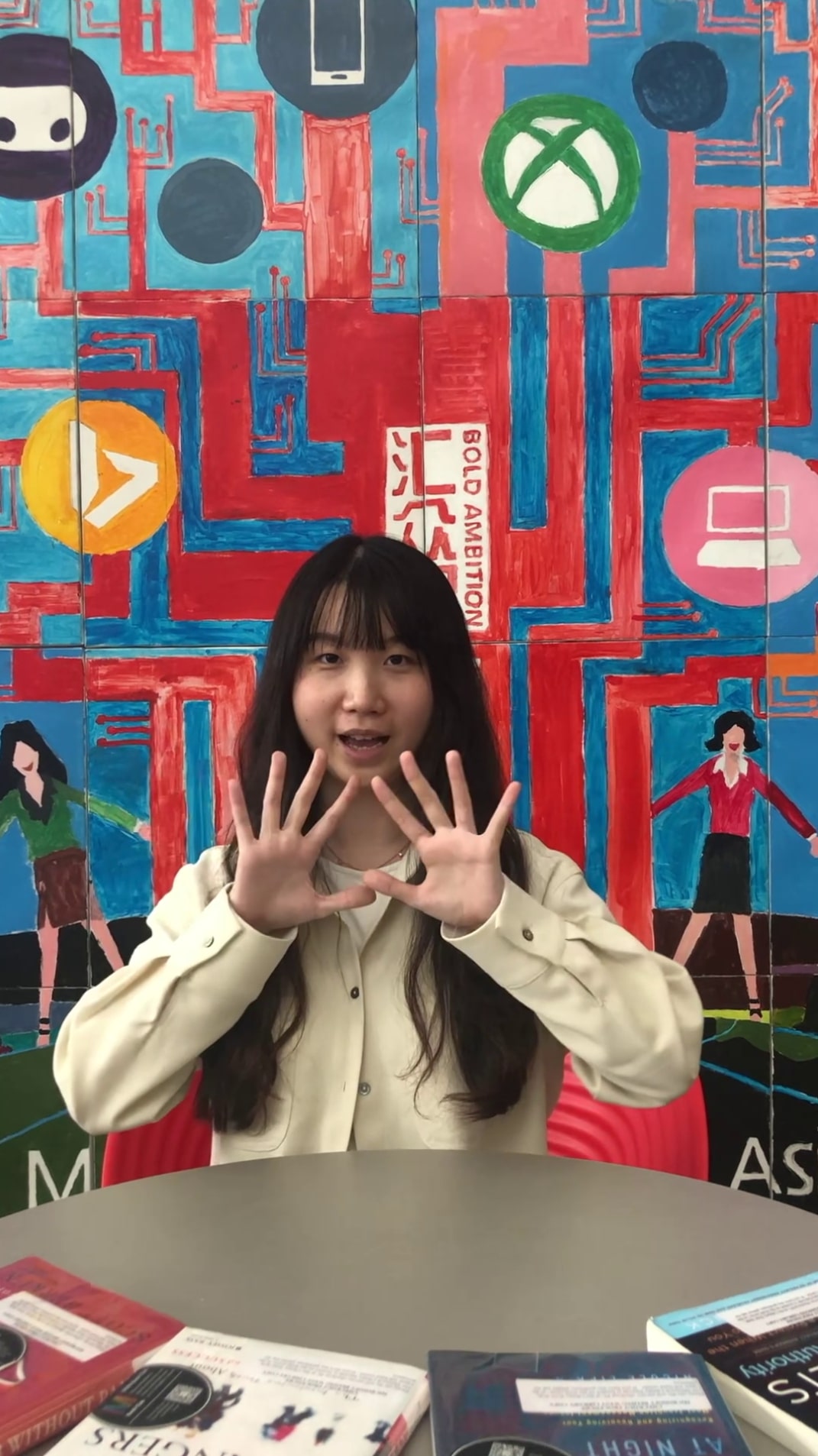}       &        
     \includegraphics[width=0.16\linewidth]{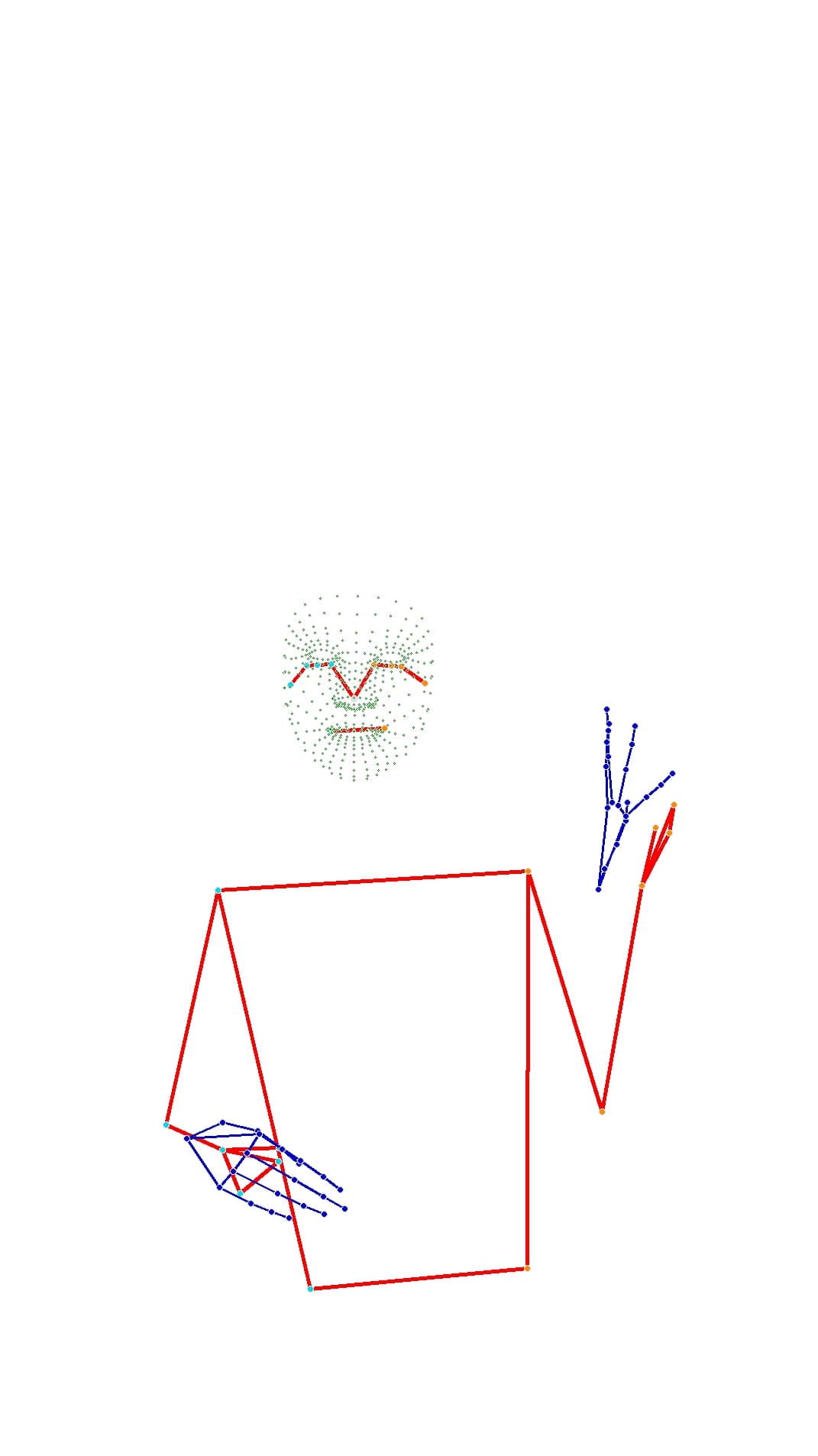}        &   
     \includegraphics[width=0.16\linewidth]{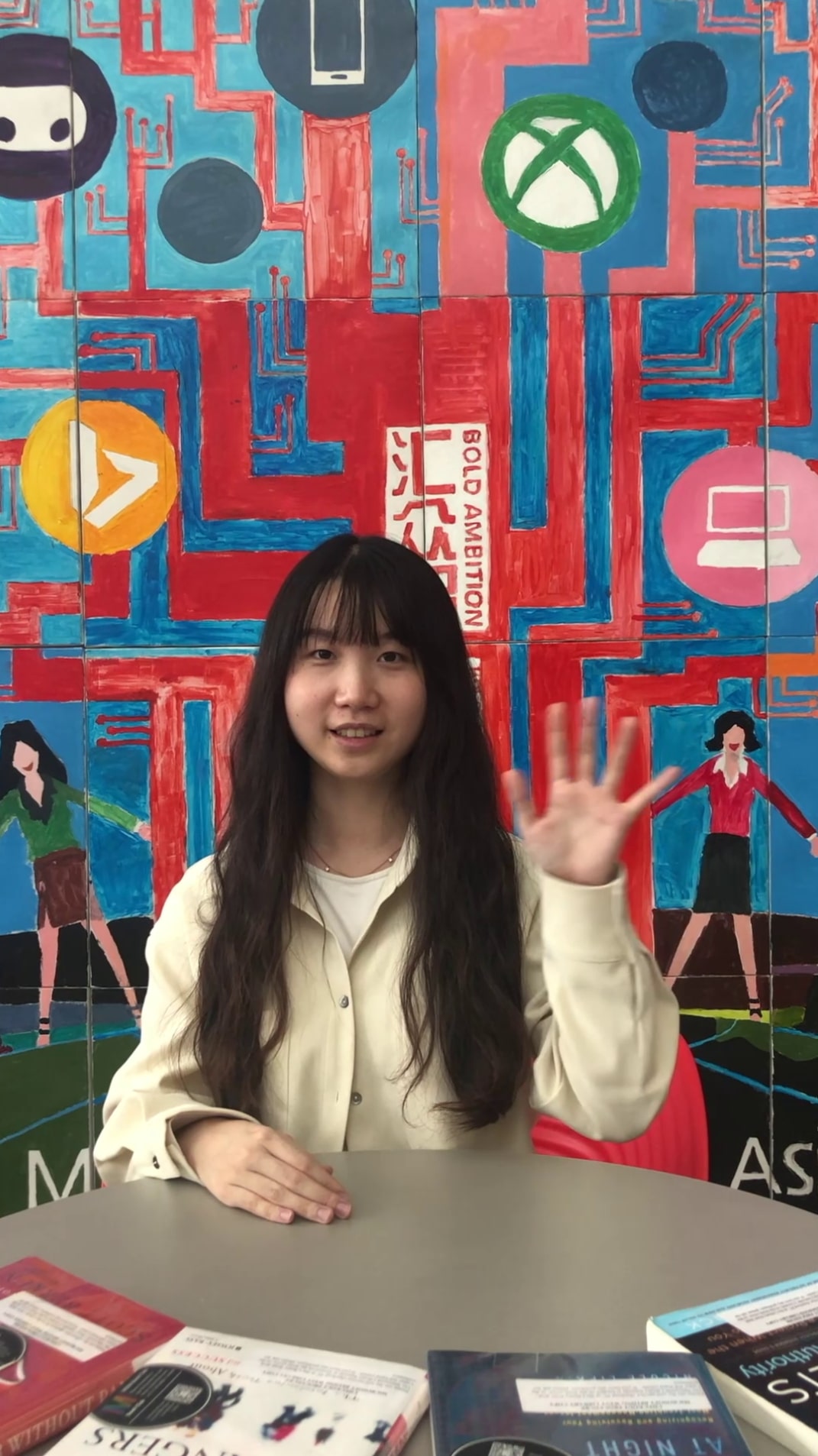} &   
     \includegraphics[width=0.16\linewidth]{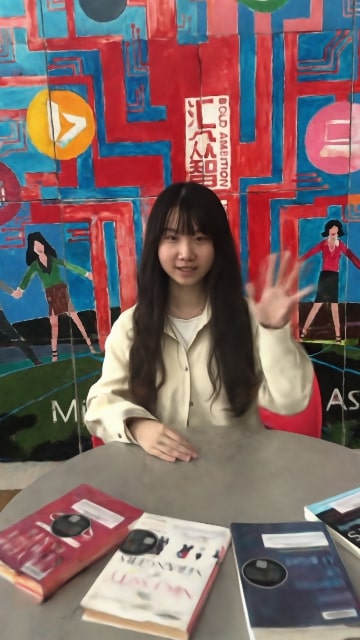} &
     \includegraphics[width=0.16\linewidth]{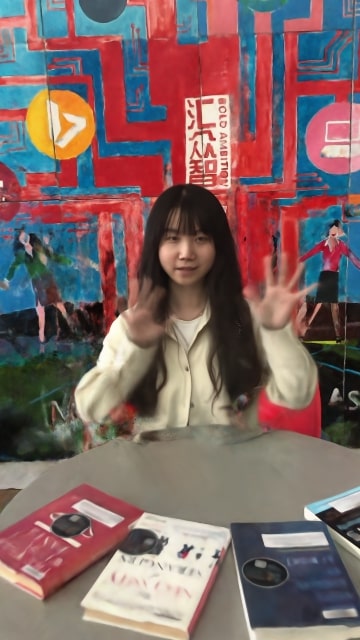}\\
          \includegraphics[width=0.16\linewidth]{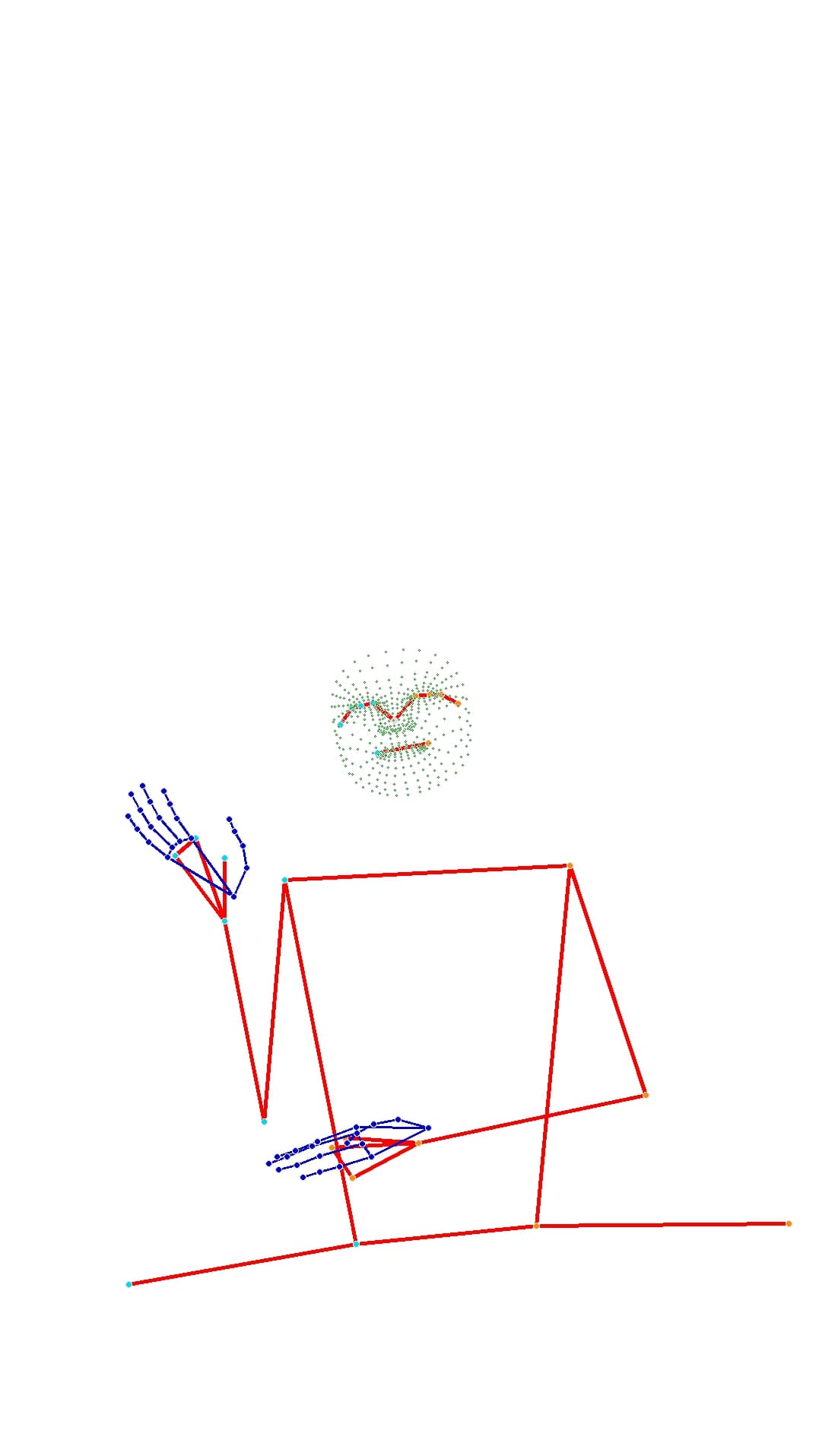}       &   
     \includegraphics[width=0.16\linewidth]{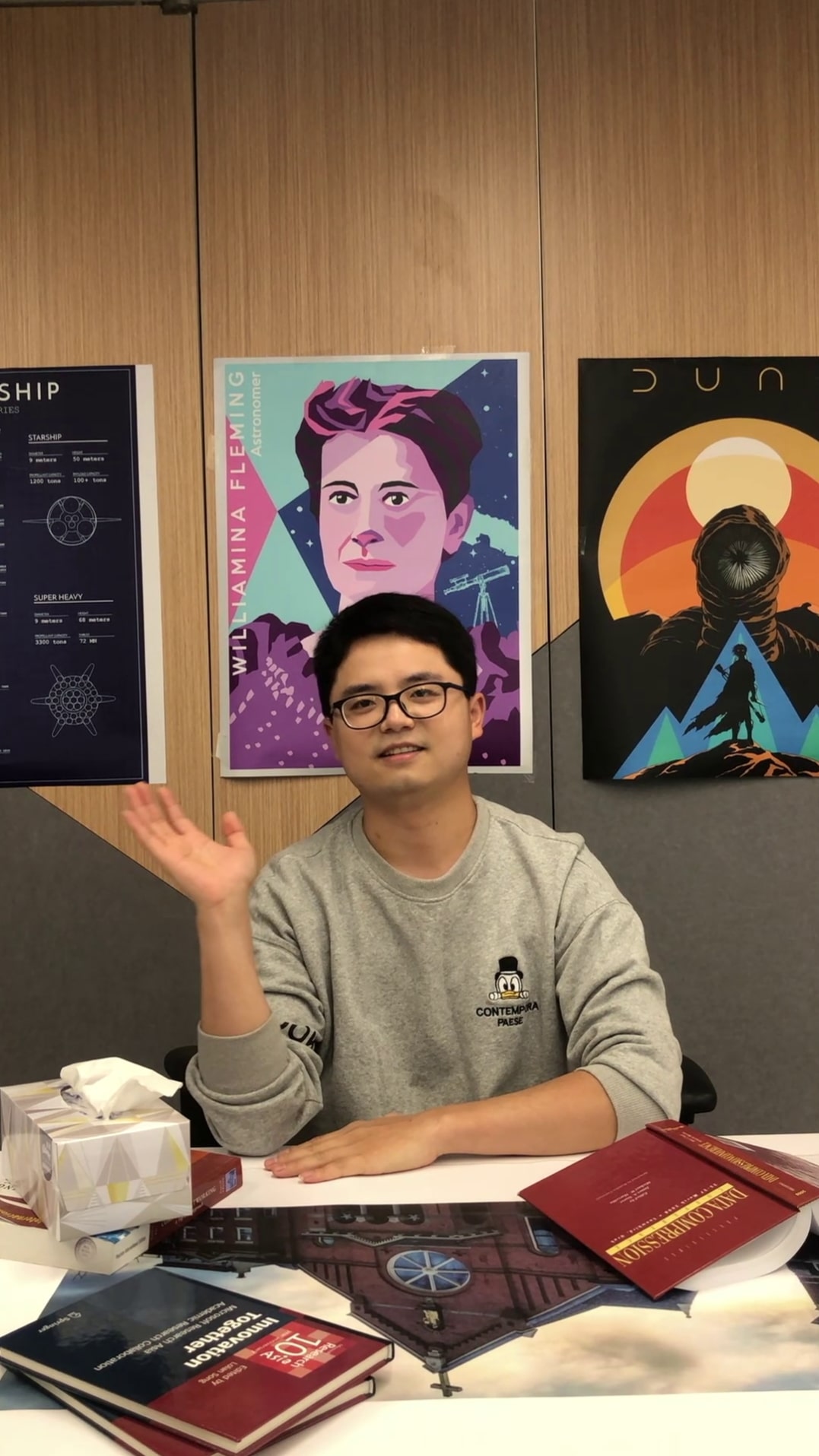}       &        
     \includegraphics[width=0.16\linewidth]{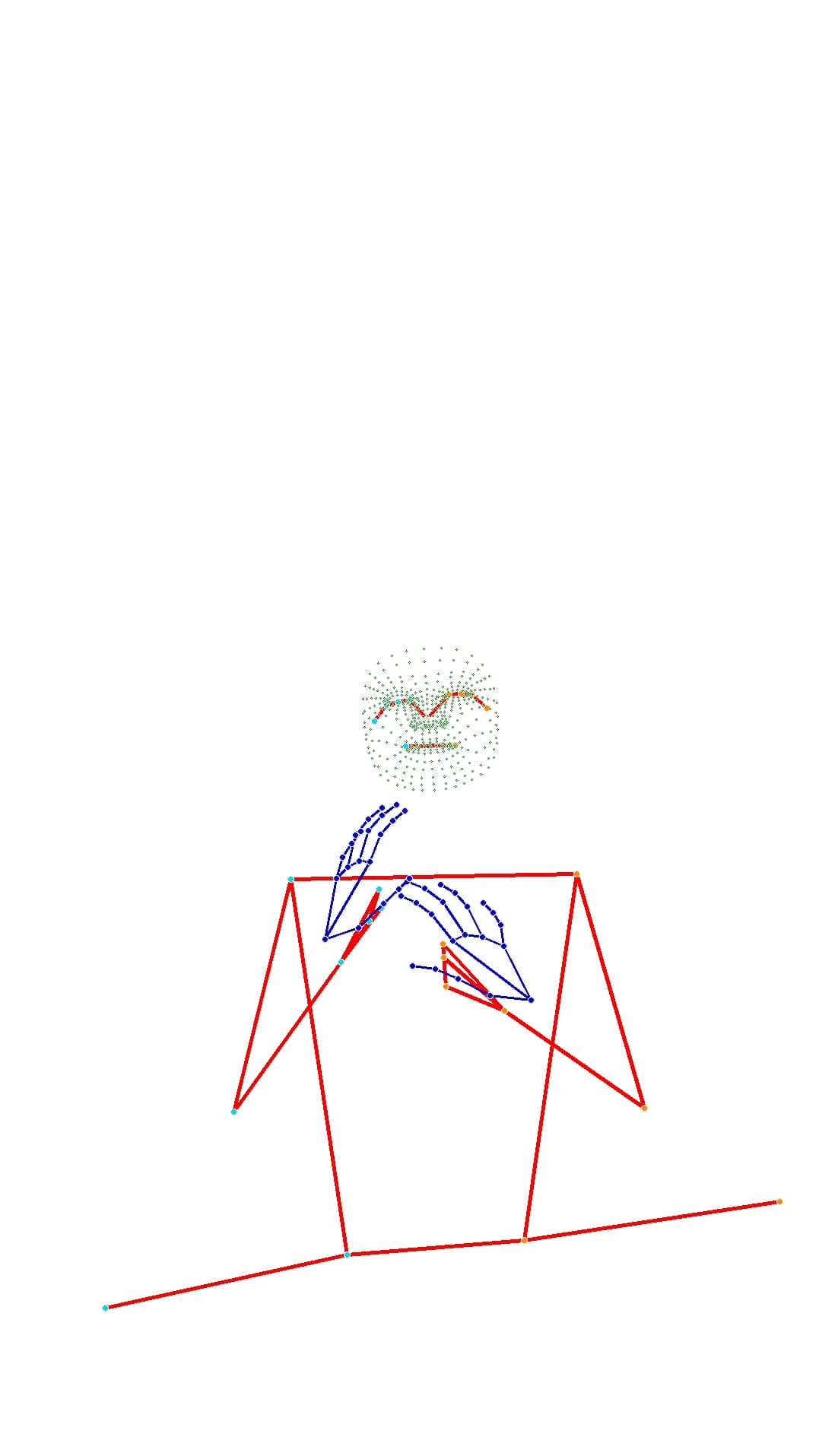}        &   
     \includegraphics[width=0.16\linewidth]{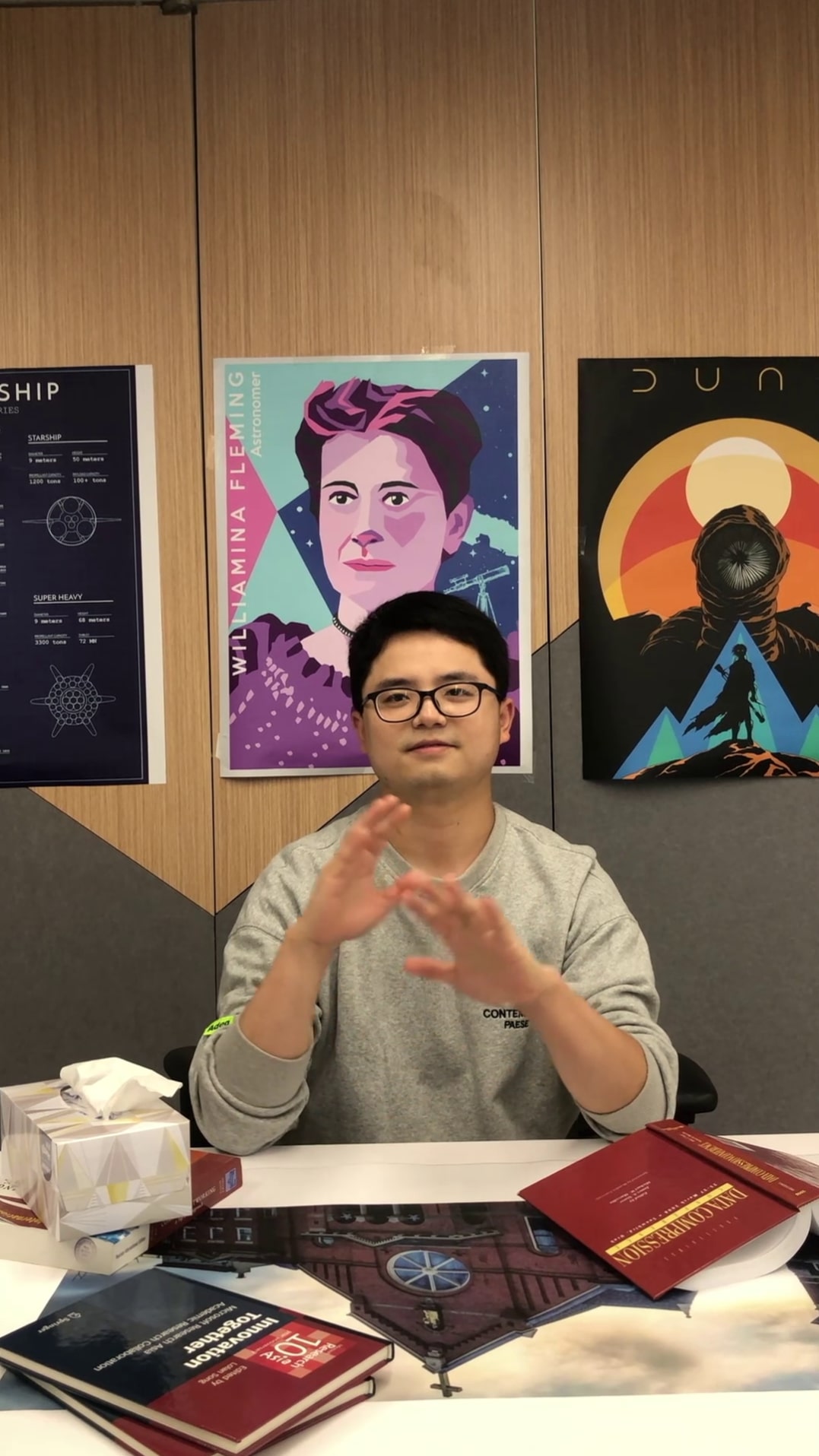} &   
     \includegraphics[width=0.16\linewidth]{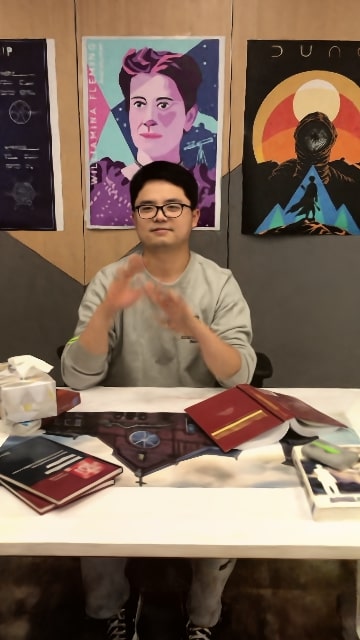} &
     \includegraphics[width=0.16\linewidth]{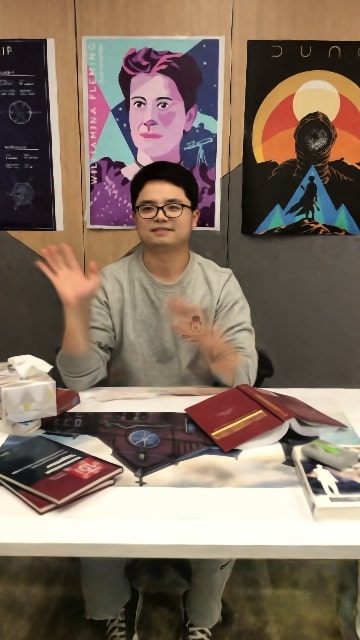}\\
      \includegraphics[width=0.16\linewidth]{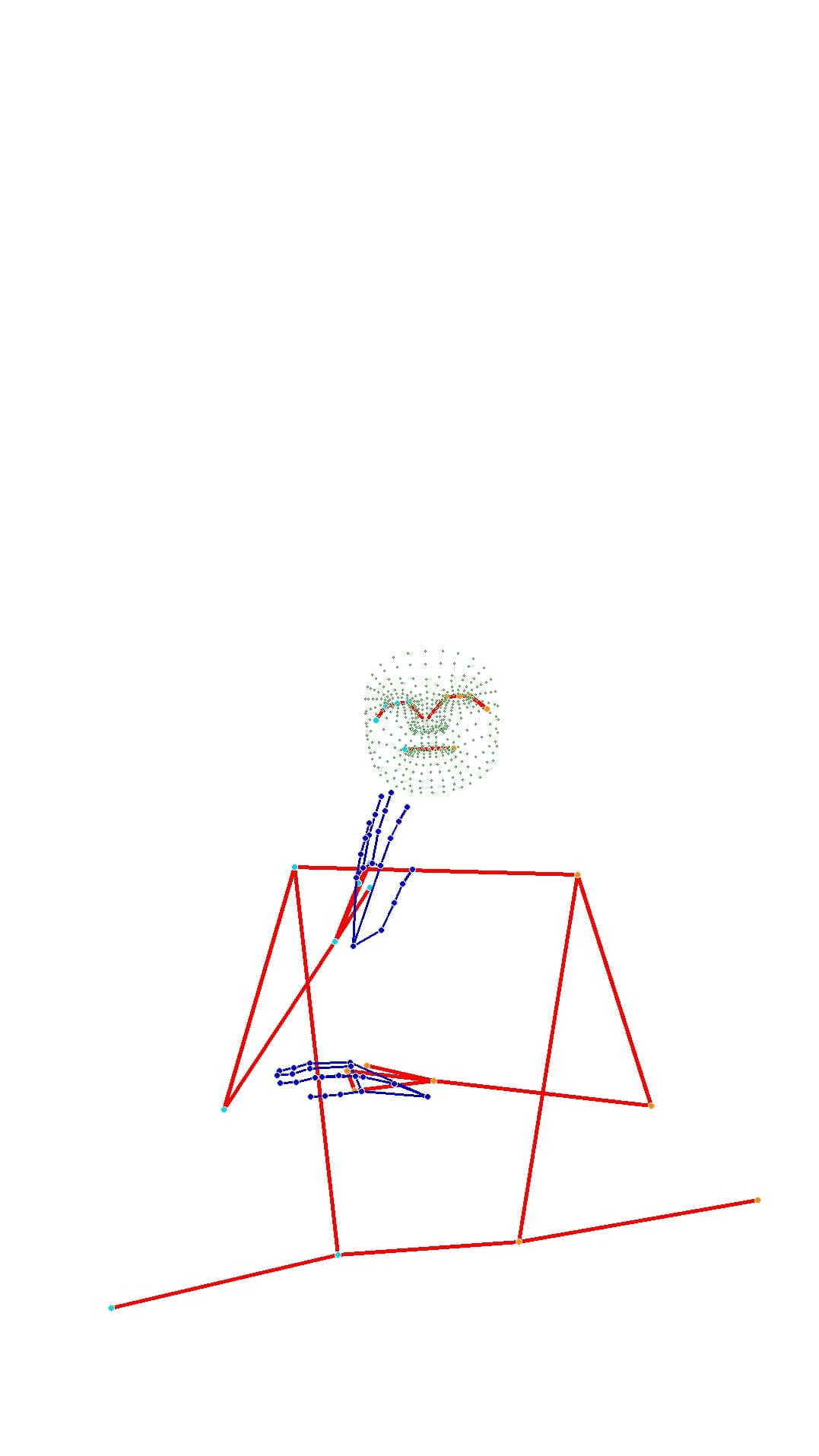}       &   
     \includegraphics[width=0.16\linewidth]{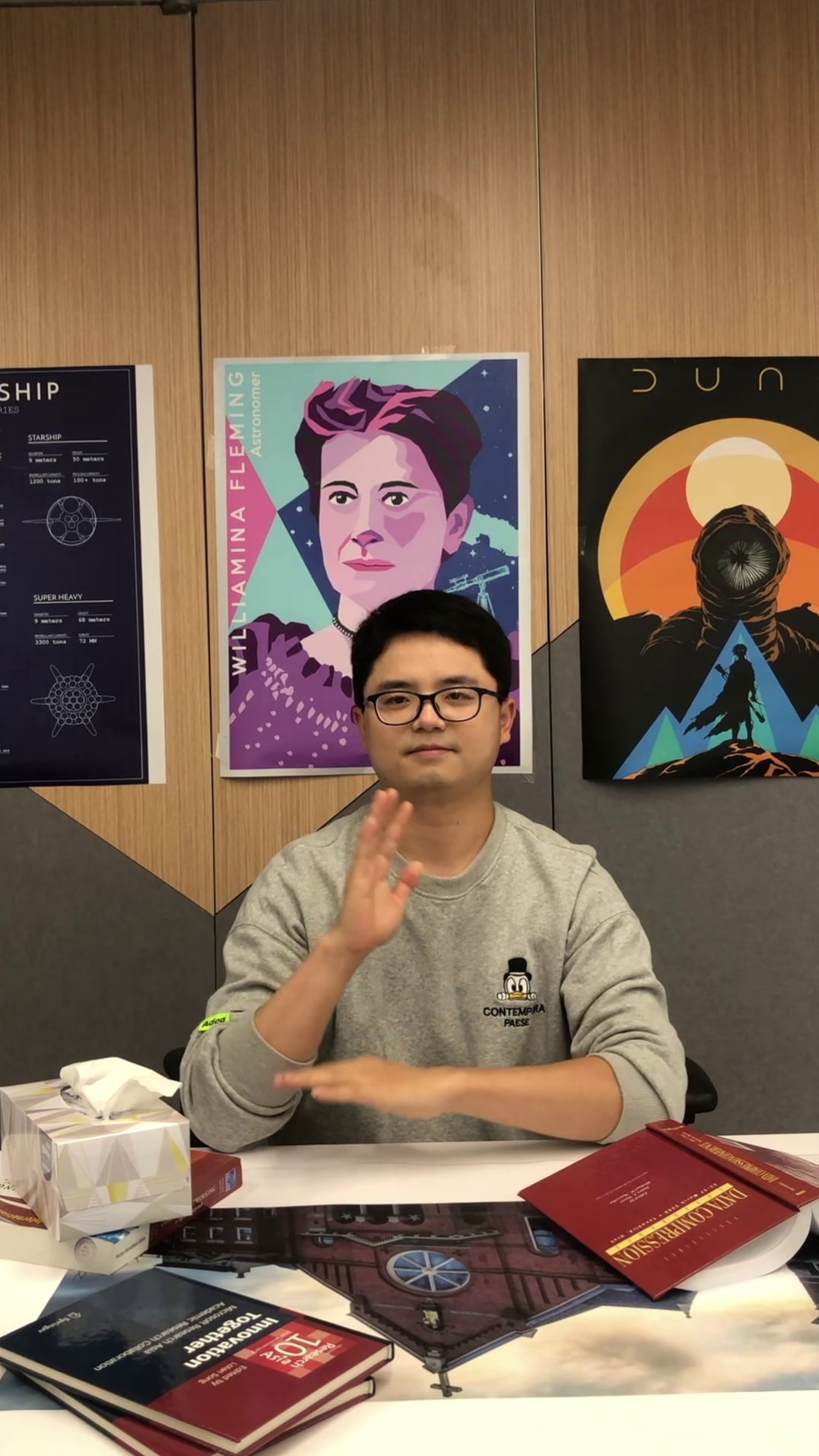}       &        
     \includegraphics[width=0.16\linewidth]{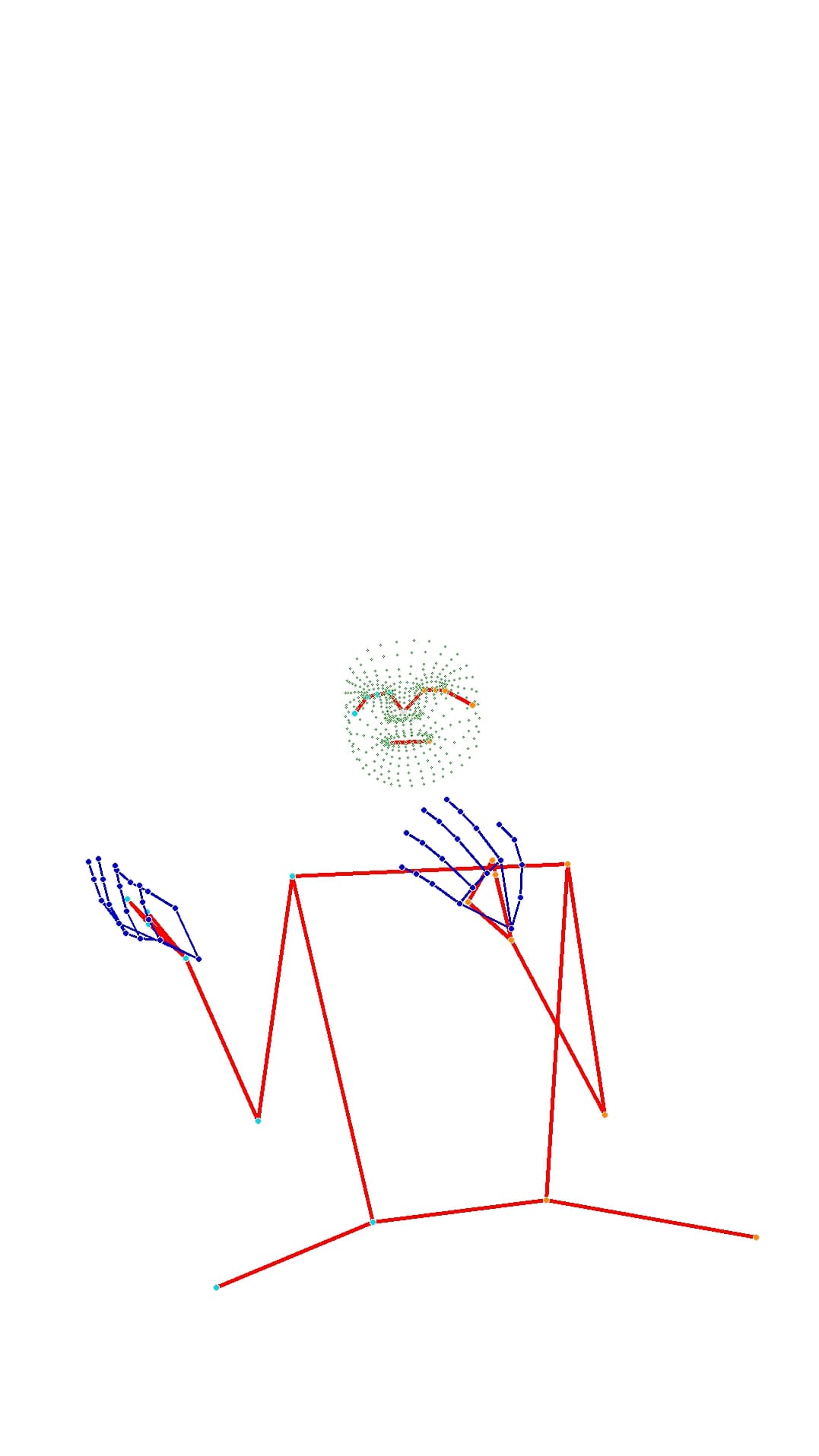}        &   
     \includegraphics[width=0.16\linewidth]{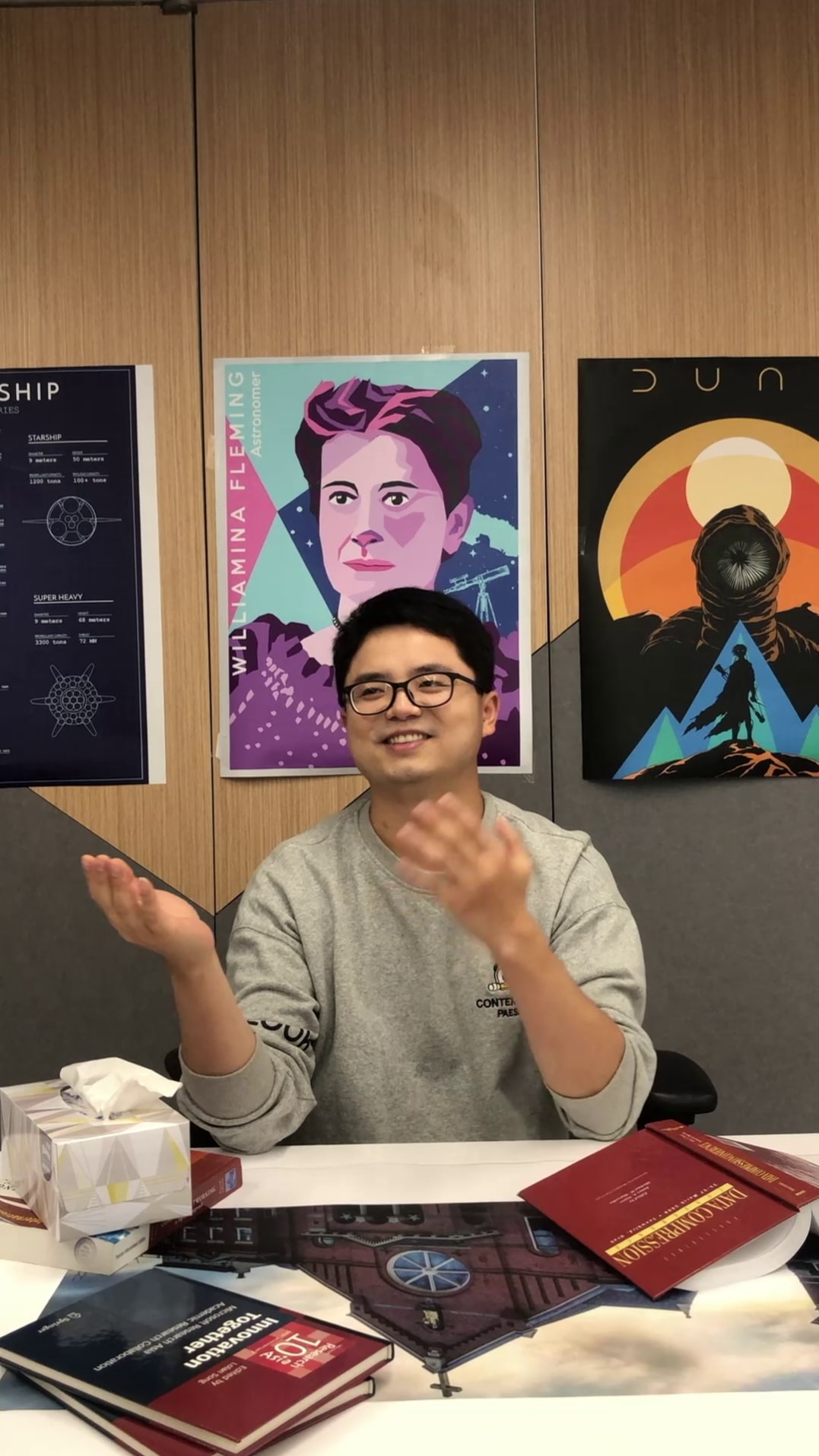} &   
     \includegraphics[width=0.16\linewidth]{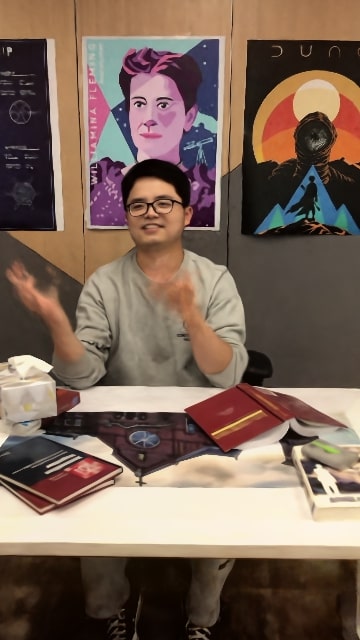} &
     \includegraphics[width=0.16\linewidth]{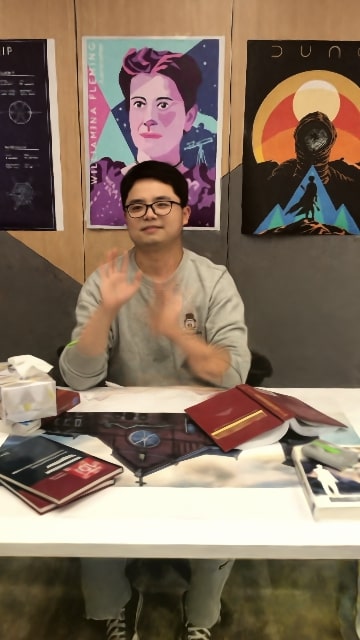}\\

    Pose1 & GT1 & Pose2 & GT2 & HyperNeRF & Nerfies \\
    \end{tabular}
    \vspace{-0.5mm}
   \caption{{Novel pose combination using NeRF-based approaches. Due to the implicit modeling of deformation, these methods are good at memorizing the scene but are not friendly to explicit control, thus failing to generalize to unseen poses that are slightly different from the training samples.}}
    \label{fig:pose_combination}
\end{figure*}

We show the results for novel pose combination in Fig.~\ref{fig:pose_combination}. In this example, we want to combine the right hand of the first input with the body and the face with the second input. We can see that HyperNeRF does not fully utilize the information of the body keypoints and fails in the combination. Nerfies is able to generate the coarse new pose but suffers from large distortion and obvious artifacts in the resulting image. The generated pose is also not accurately aligned with the desired combination. As the NeRF-based method is based on the implicit field, we show that the generalization ability is not as good as the proposed method which allows explicit control and shows better generalization ability. 

% \subsection{Learning adaptive planes}
\section{More Results}
In the next, we provide additional results including the depth visualization and generalization ability experiments. We highly recommend the readers refer to the accompanied videos for more visual results.
\subsubsection{Depth visualization} 
\begin{figure*}[!t]
    \centering 
    \small
    \begin{tabular}{@{}c@{\hspace{0.mm}}c@{\hspace{.4mm}}c@{\hspace{0.mm}}c@{\hspace{.4mm}}c@{\hspace{0.mm}}c@{}}
     \includegraphics[width=0.16\linewidth]{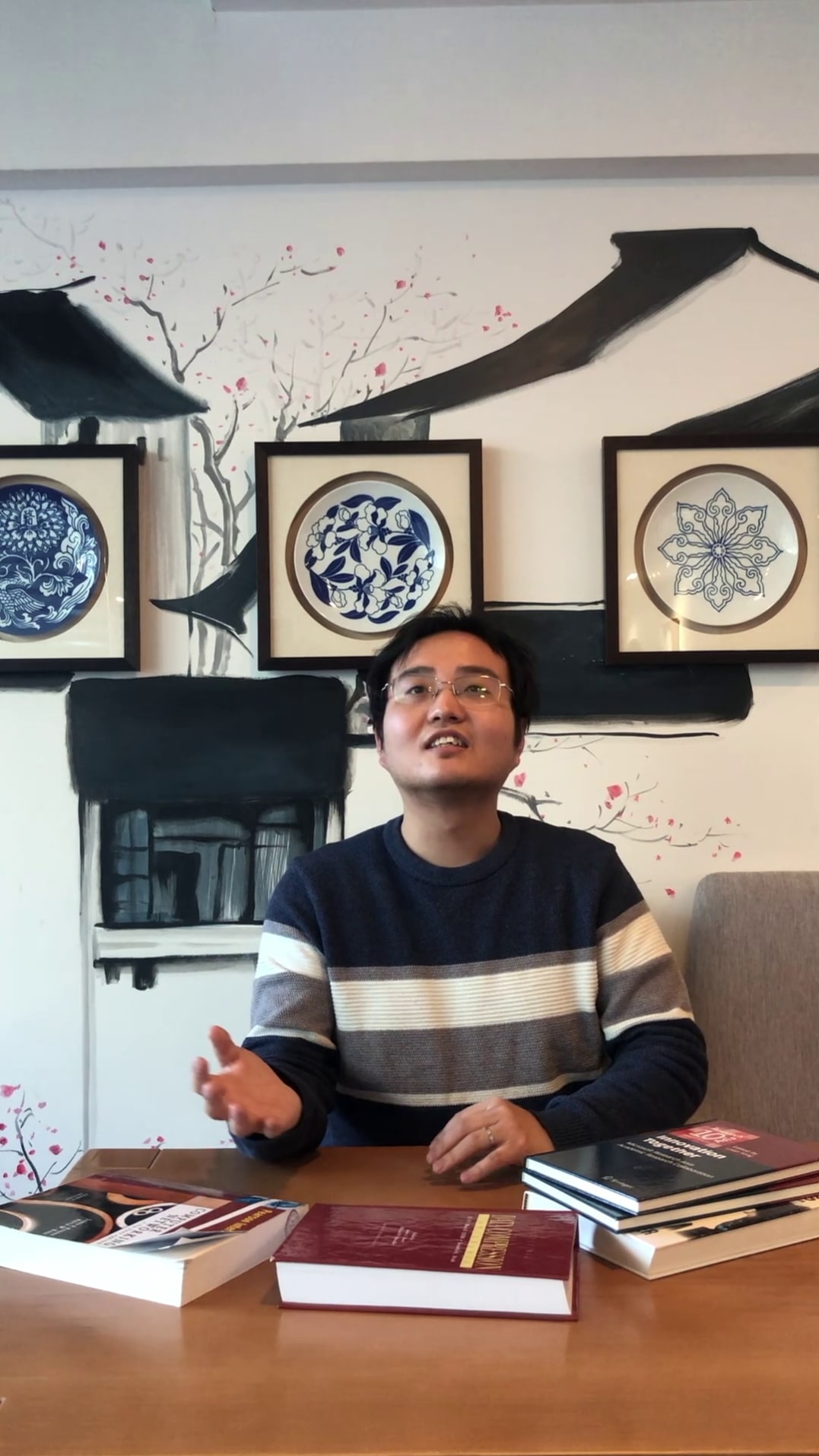}       &   
     \includegraphics[width=0.16\linewidth]{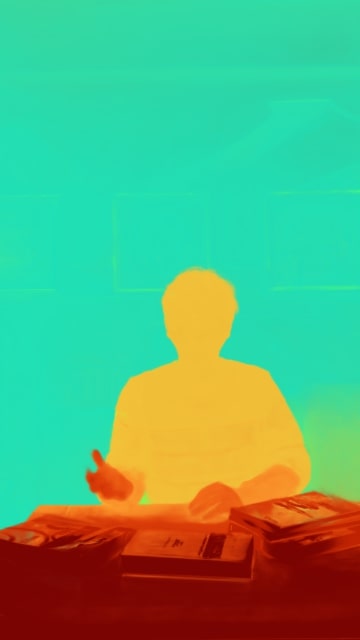}       &        
     \includegraphics[width=0.16\linewidth]{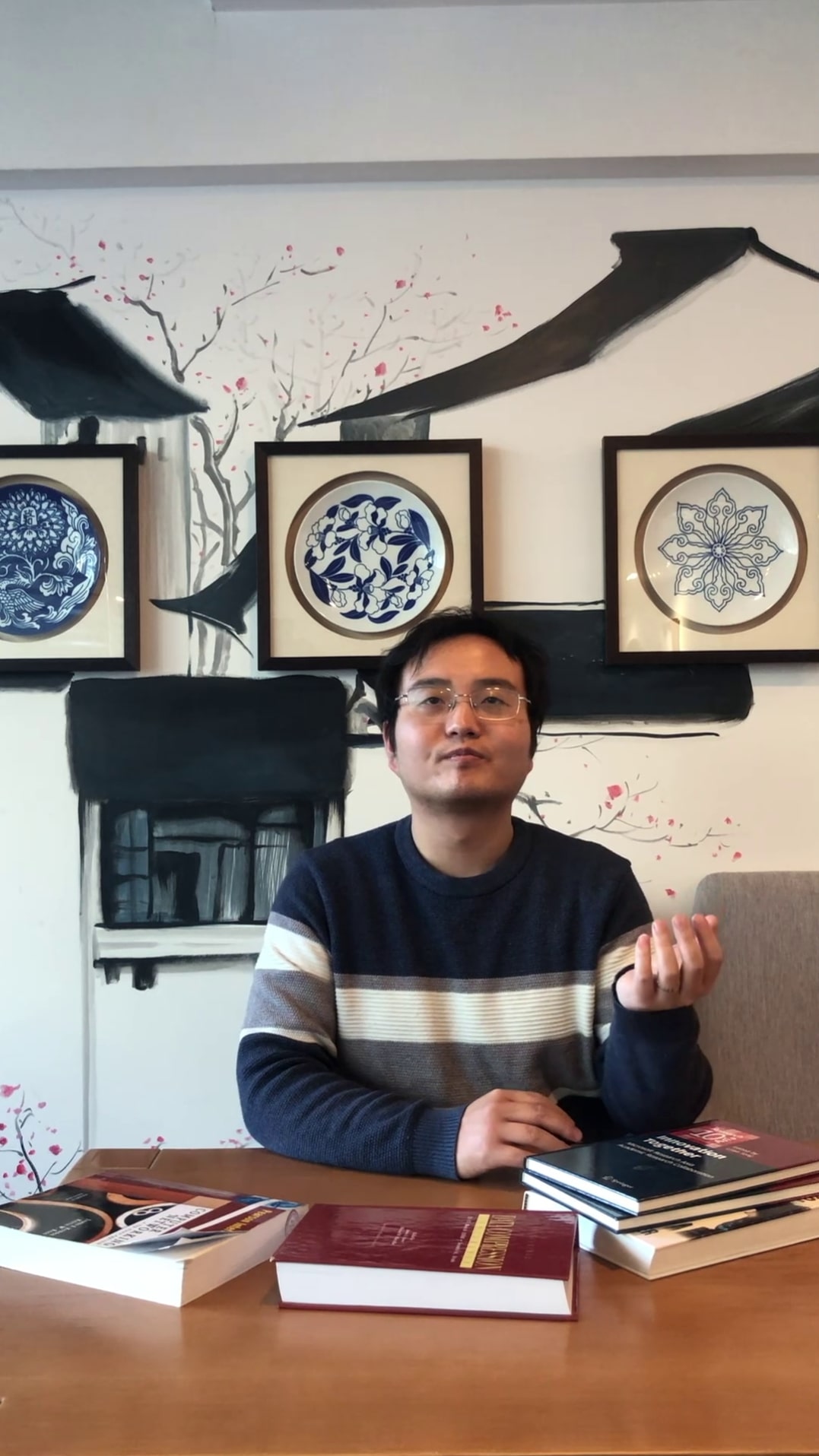}        &   
     \includegraphics[width=0.16\linewidth]{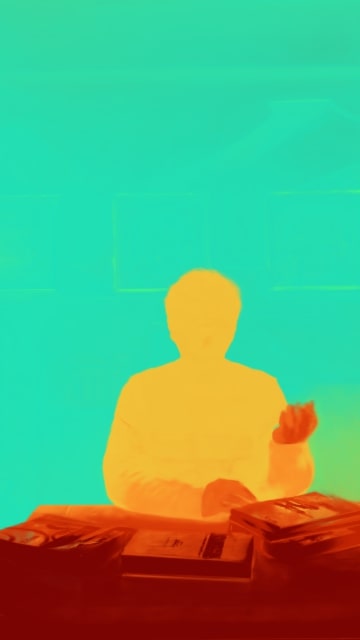} &   
     \includegraphics[width=0.16\linewidth]{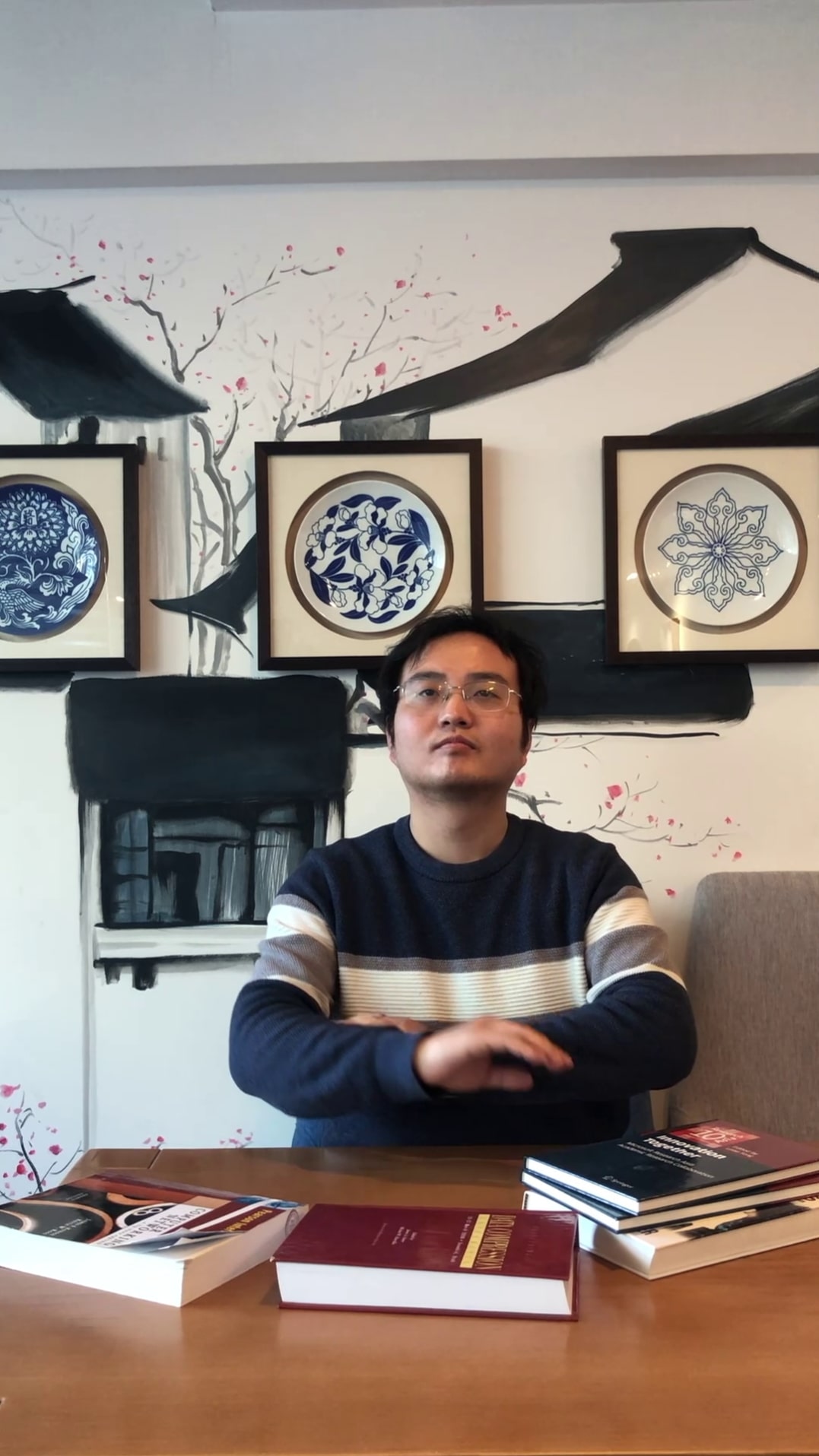} &
     \includegraphics[width=0.16\linewidth]{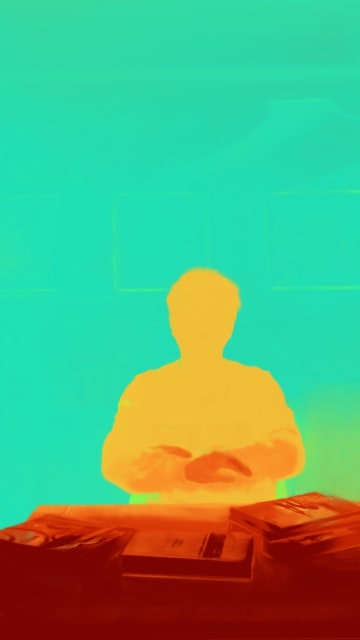}\\
     \includegraphics[width=0.16\linewidth]{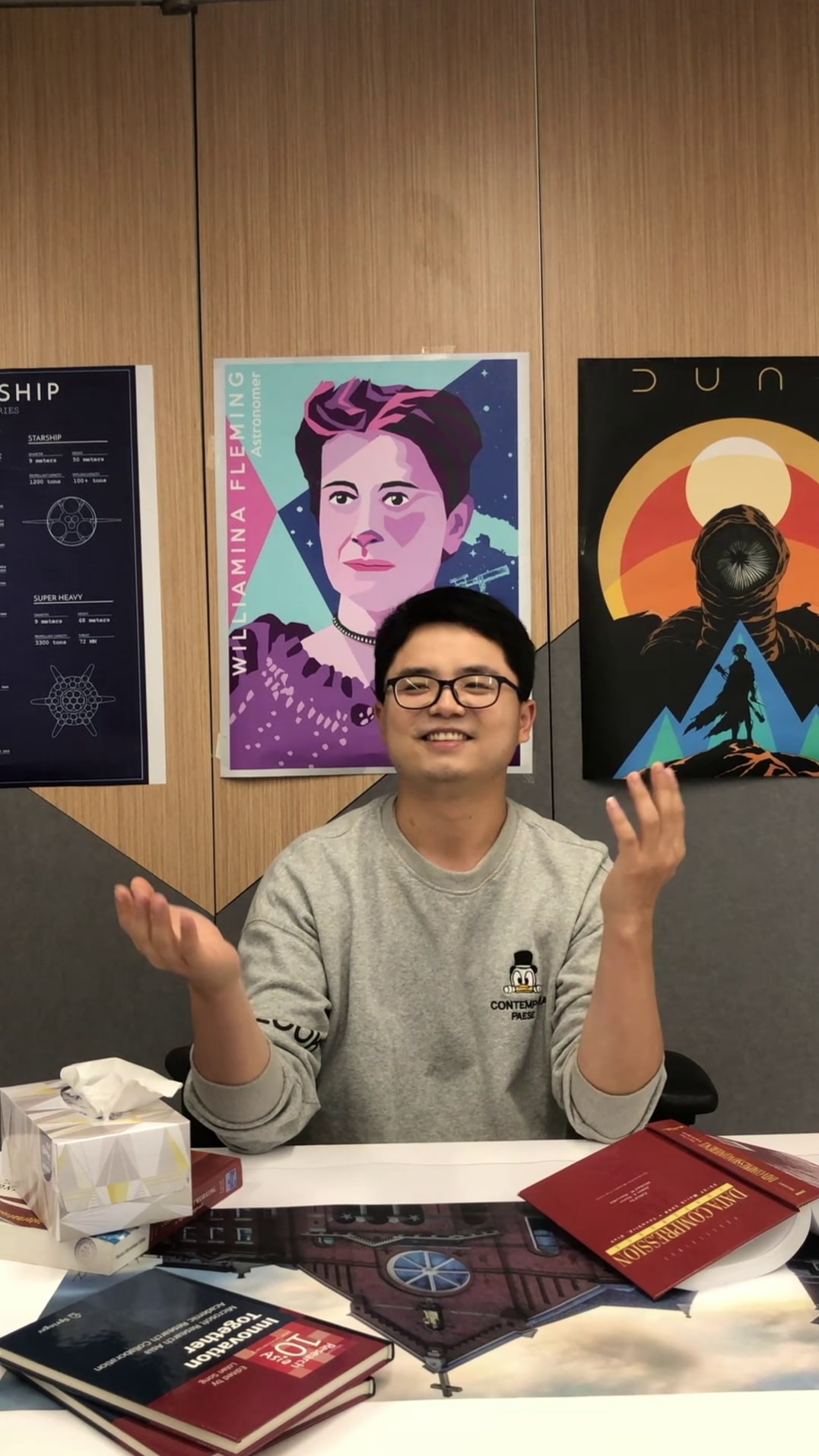}       &   
     \includegraphics[width=0.16\linewidth]{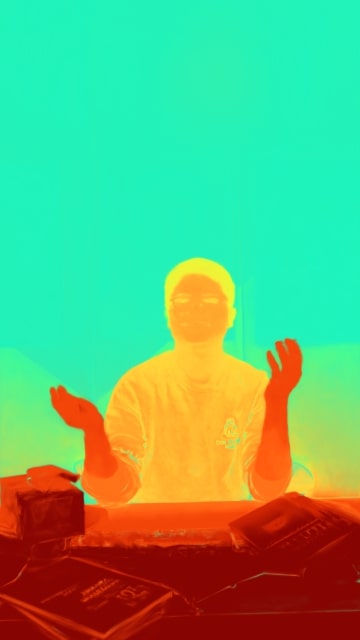}       &        
     \includegraphics[width=0.16\linewidth]{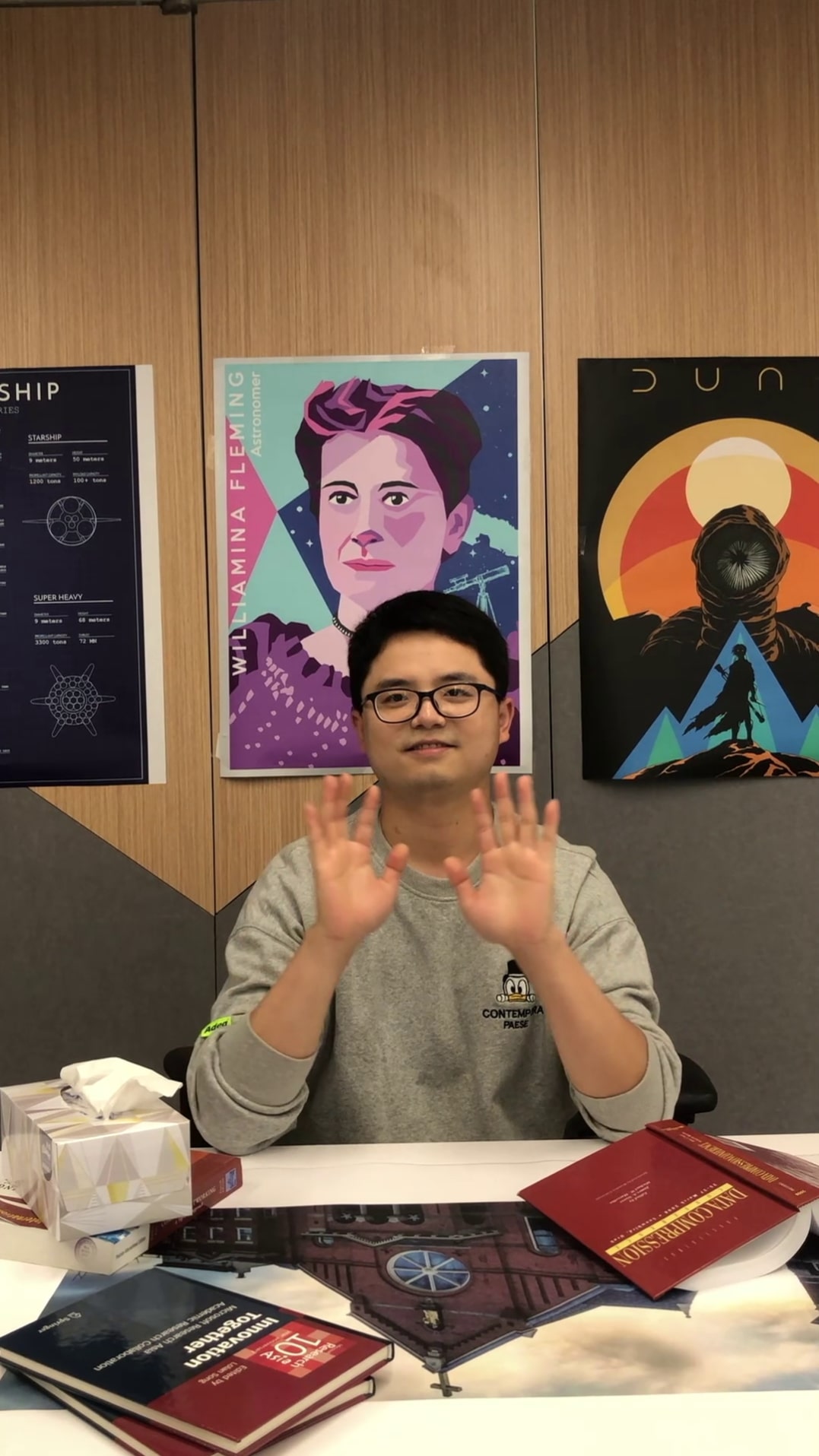}        &   
     \includegraphics[width=0.16\linewidth]{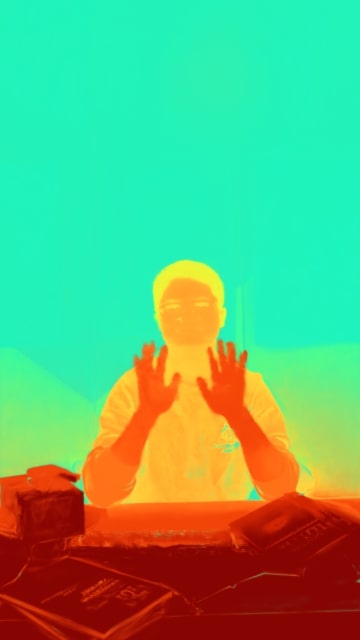} &   
     \includegraphics[width=0.16\linewidth]{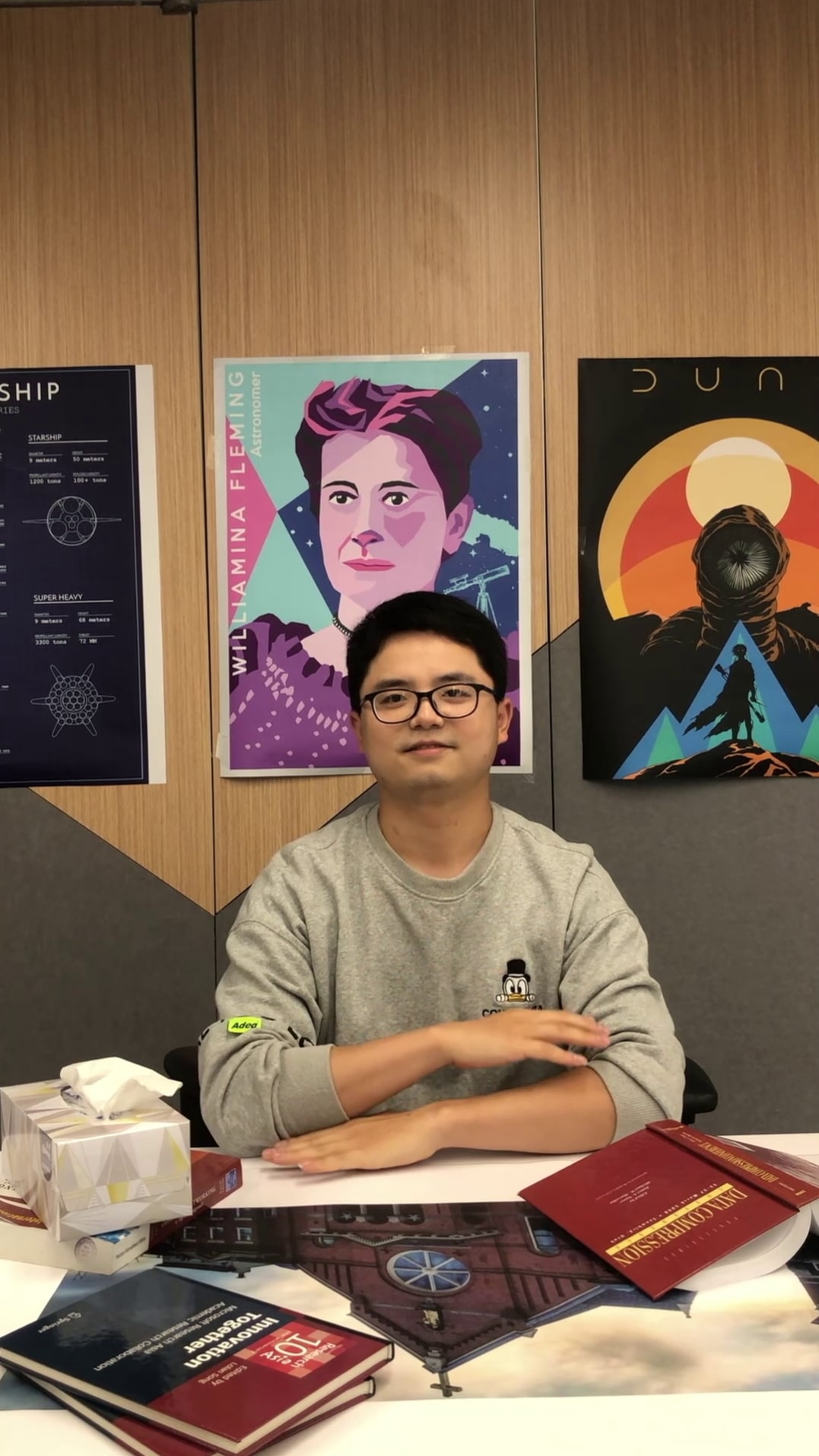} &
     \includegraphics[width=0.16\linewidth]{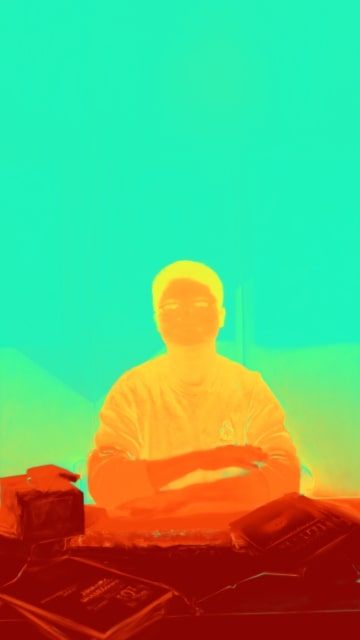}\\
         \includegraphics[width=0.16\linewidth]{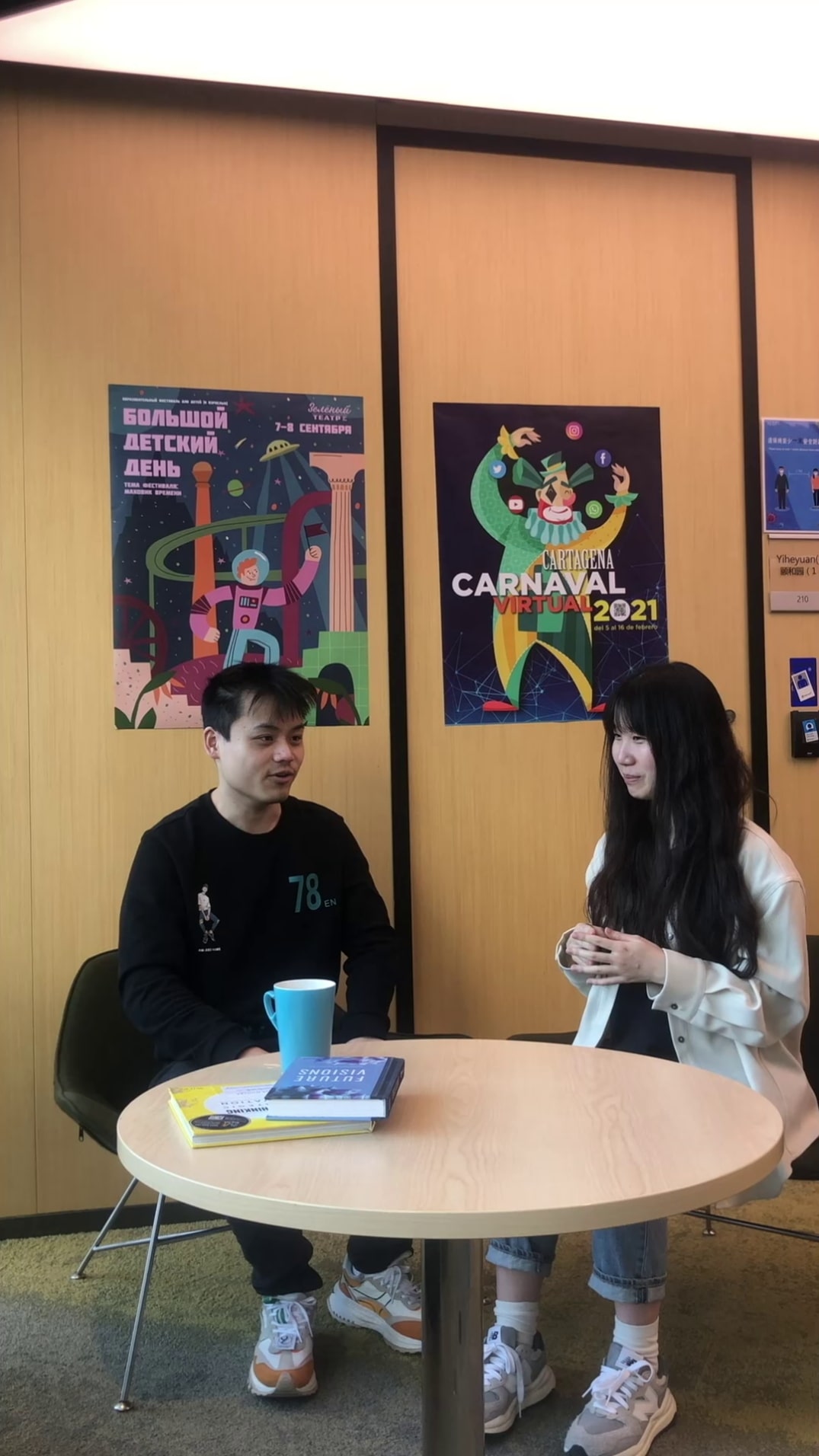}       &   
     \includegraphics[width=0.16\linewidth]{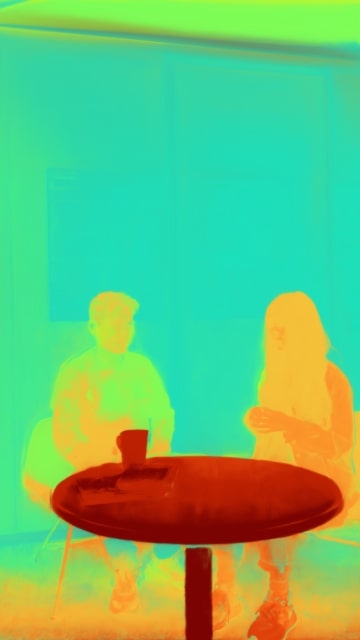}       &        
     \includegraphics[width=0.16\linewidth]{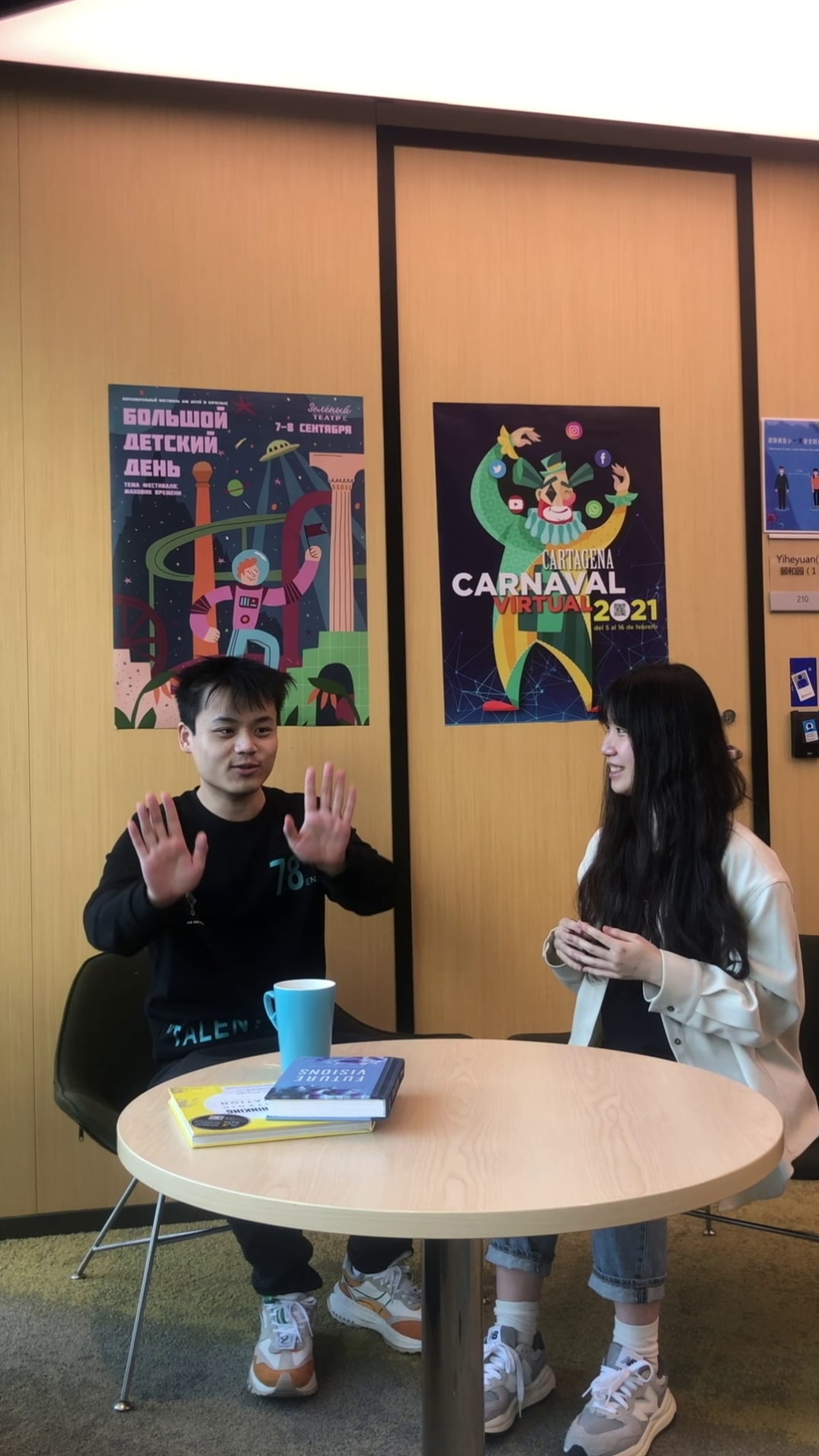}        &   
     \includegraphics[width=0.16\linewidth]{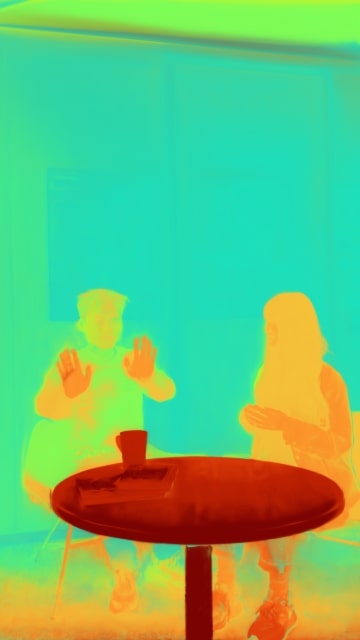} &   
     \includegraphics[width=0.16\linewidth]{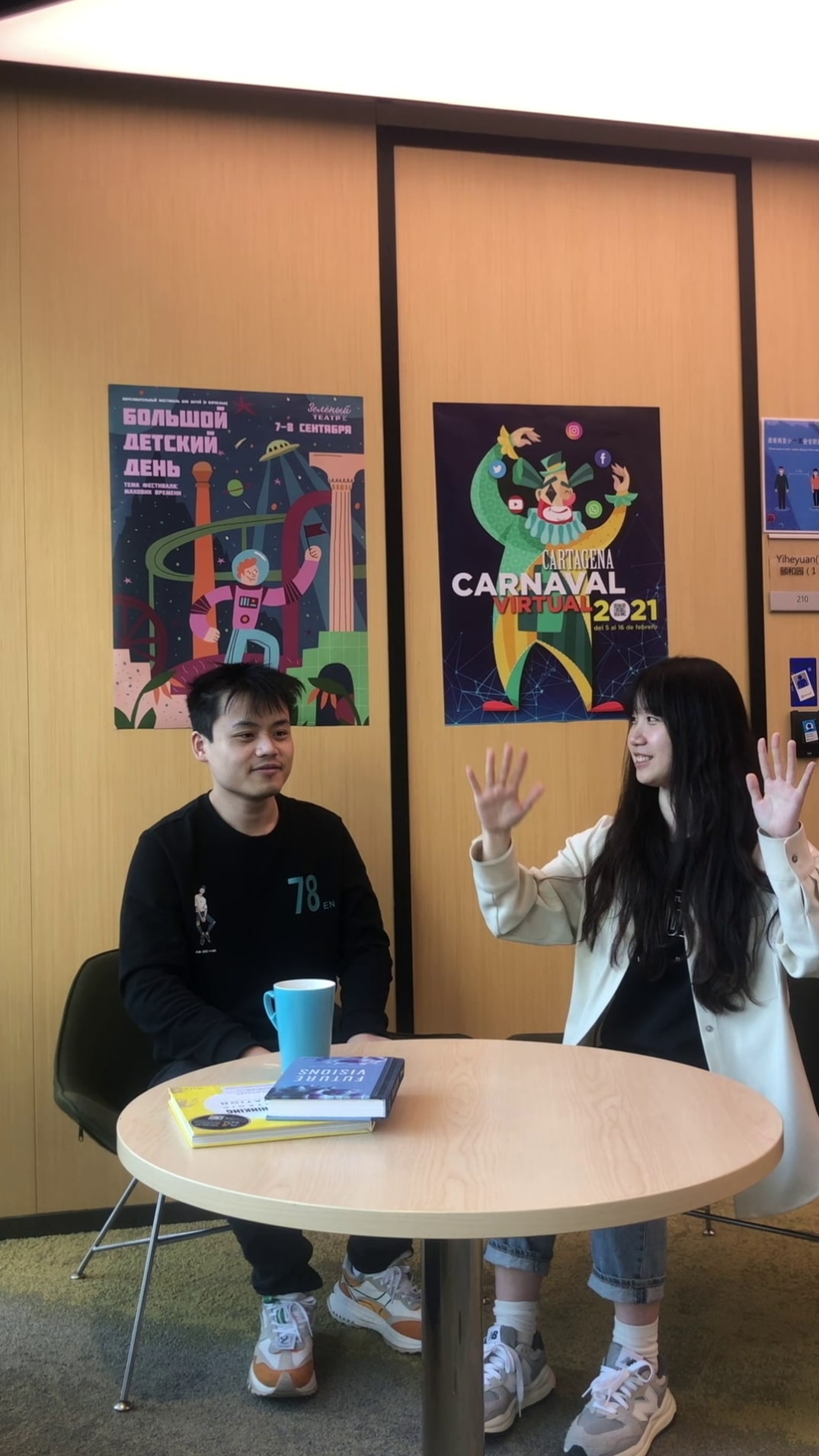} &
     \includegraphics[width=0.16\linewidth]{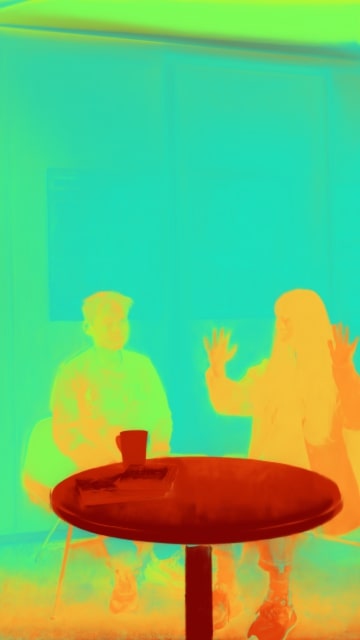}\\
              \includegraphics[width=0.16\linewidth]{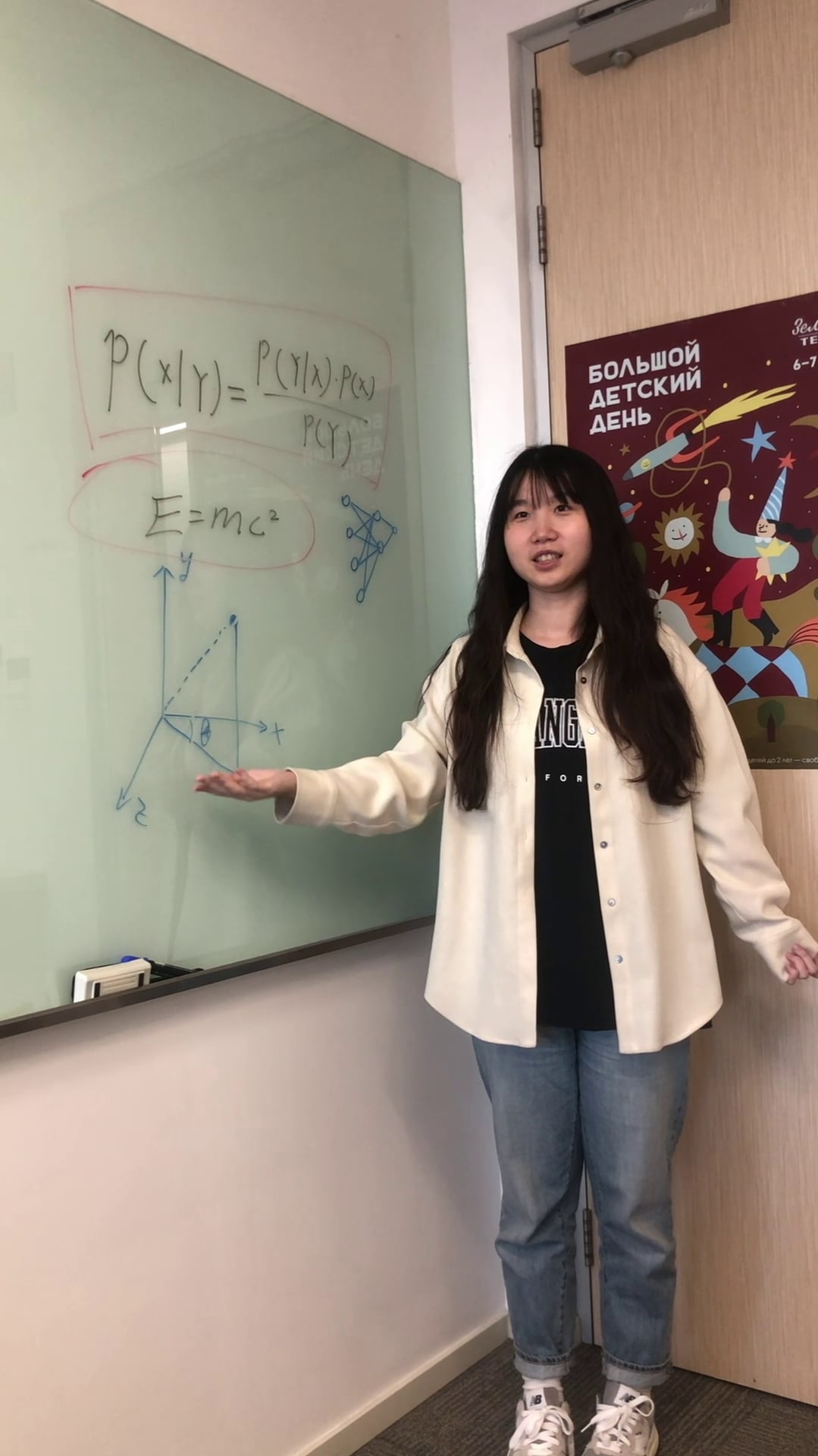}       &   
     \includegraphics[width=0.16\linewidth]{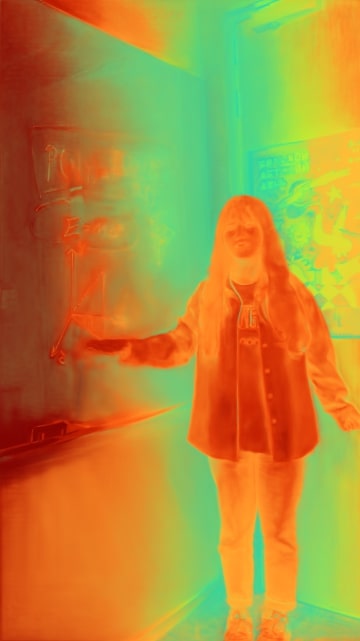}       &        
     \includegraphics[width=0.16\linewidth]{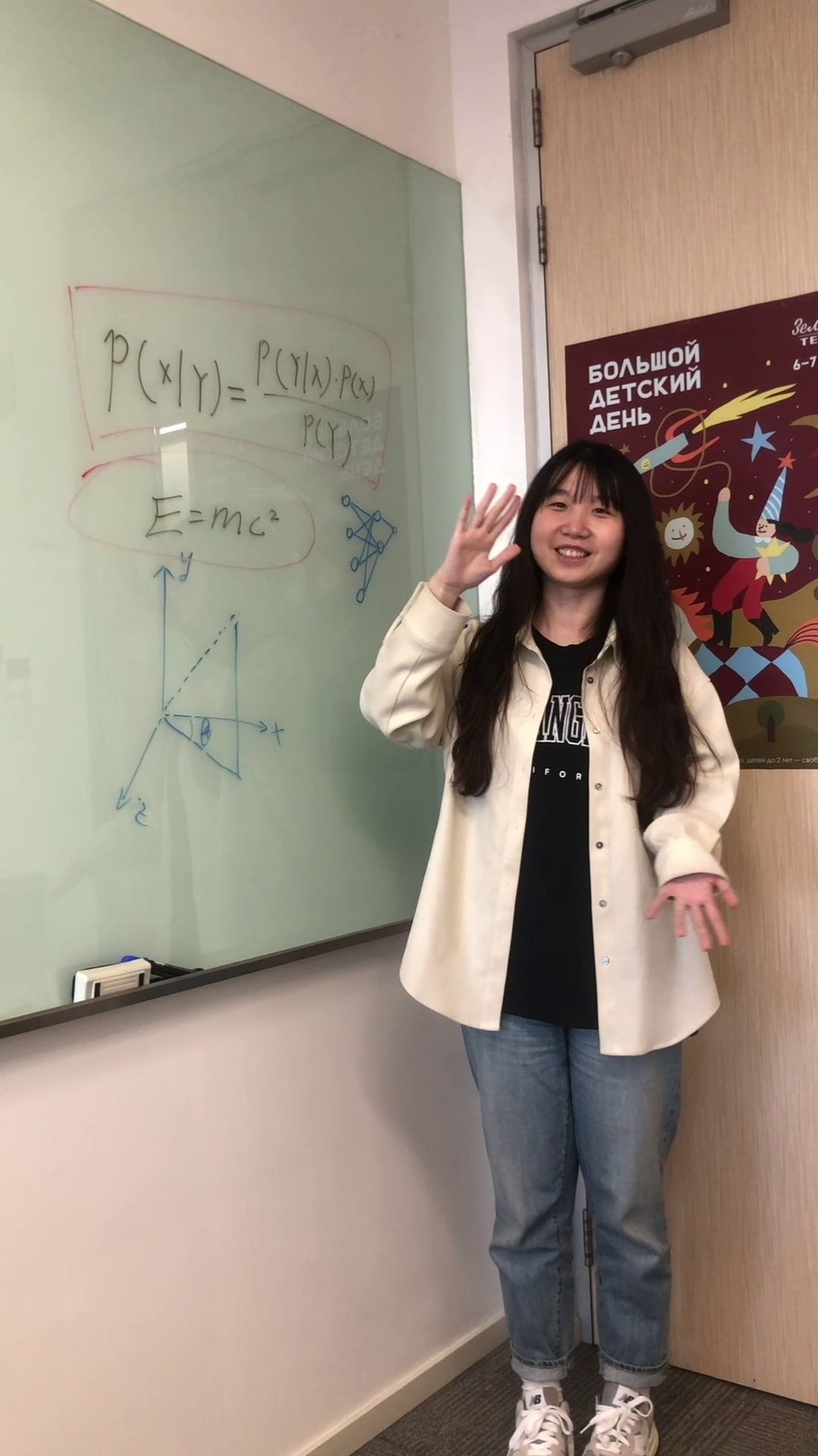}        &   
     \includegraphics[width=0.16\linewidth]{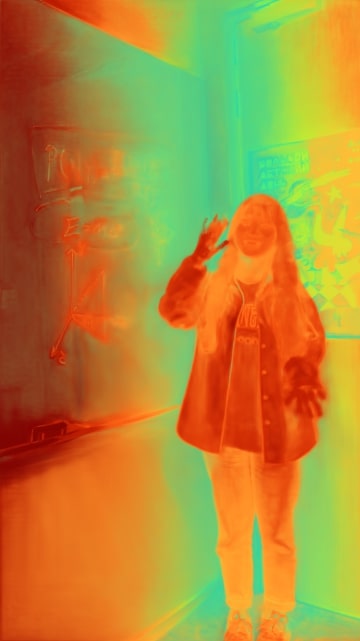} &   
     \includegraphics[width=0.16\linewidth]{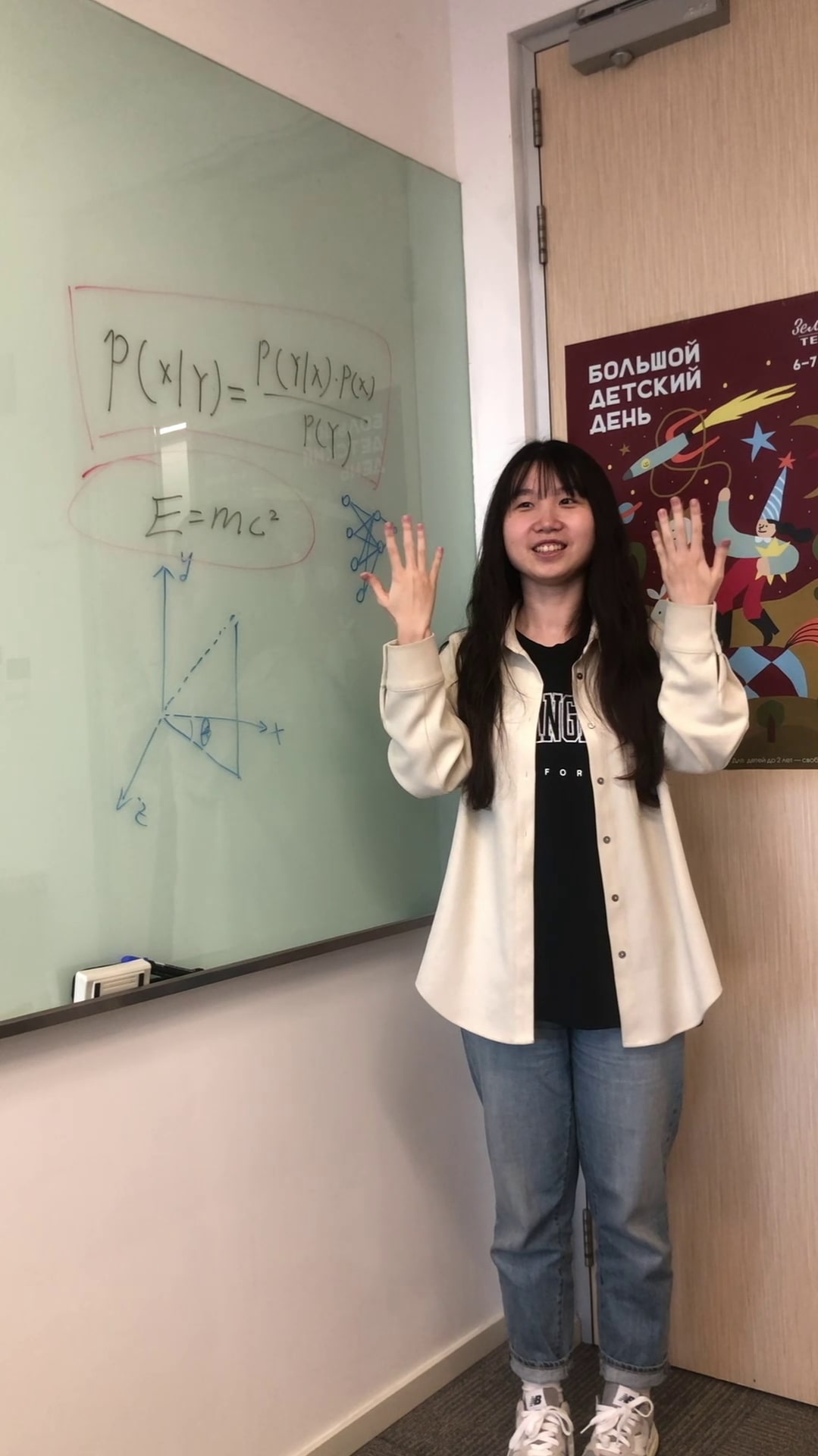} &
     \includegraphics[width=0.16\linewidth]{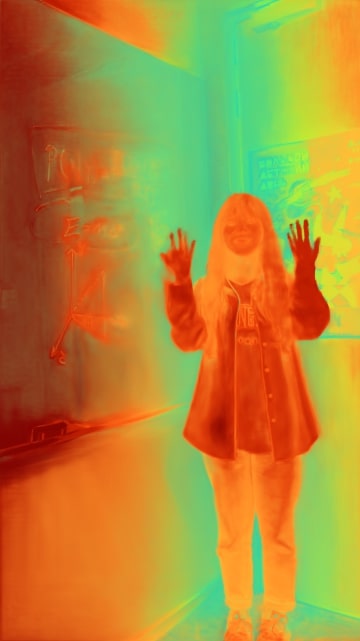}\\
    Input & Depth & Input & Depth & Input & Depth \\
    \end{tabular}
    \vspace{-0.5mm}
   \caption{{Depth visualization. Our method can generate a smooth, boundary crisp depth map, showing that the learned MPI does well model the 3D geometry of the scene.}}
    \label{fig:depth}
\end{figure*}

By compositing the depth instead of the color texture, we can generate the depth map for each frame as visualized in Fig.~\ref{fig:depth}.

\subsubsection{Generalization ability}
We also include more examples on pose interpolation and extrapolation as in Fig.~\ref{fig:control}.

\begin{figure}[t]
\centering
\includegraphics[width=\linewidth]{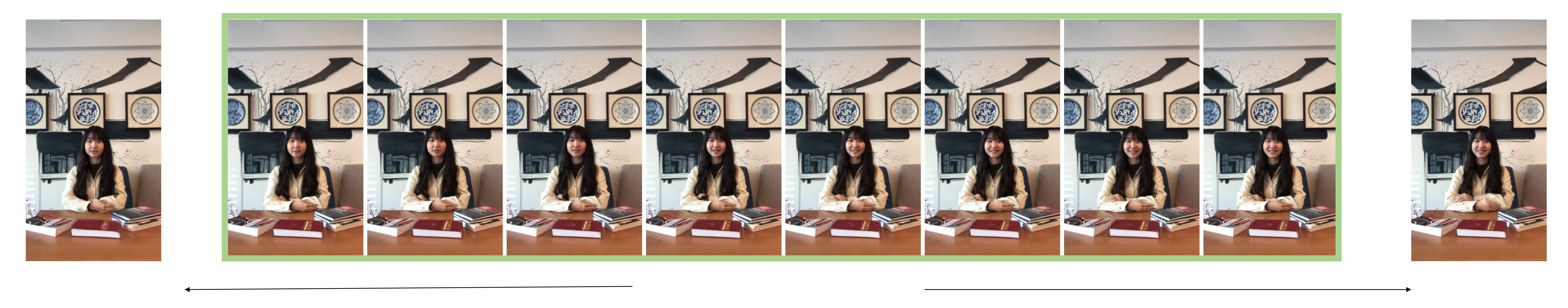}
\put(-194,2){\scriptsize Interpolation}
\put(-338,2){\scriptsize Pose 1}
\put(-32,2){\scriptsize Pose 2}
\put(-210,-10){(a) Interpolation}
\\
\includegraphics[width=0.85\linewidth]{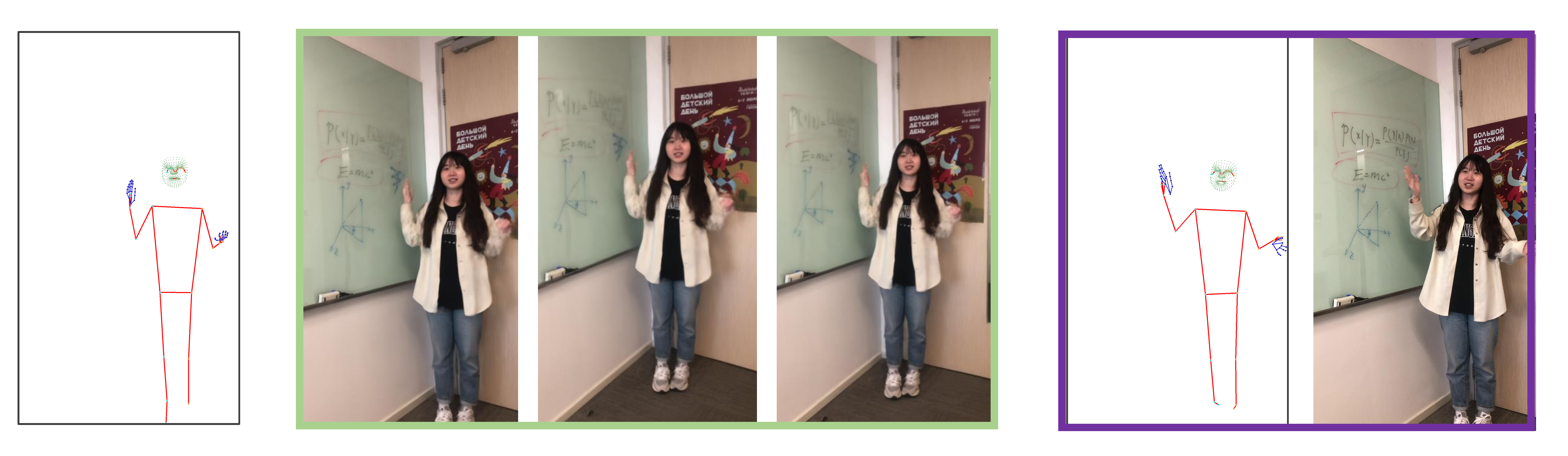}
\put(-215,95){}
\put(-288,-2){\scriptsize Unseen pose}
\put(-210,-2){\scriptsize Novel view generation}
\put(-103,-2){\scriptsize Nearest neighbor in training poses}
\put(-190,-16){(b) Extrapolation}
\caption{Generalization study. Comparing to implicit approaches, our method can generalize to unseen poses due to the generative ability of CNNs. Here we show examples on motion interpolation and extrapolation. }
\label{fig:control}
\end{figure} 

\subsubsection{More results of motion transfer} In Fig.~\ref{fig:synchronized} we showcase additional results on motion transfer. Our method is able to animate the character following the driving subject while yielding photo-realistic quality. 

\begin{figure*}[!t]
    \centering 
    \small
    \begin{tabular}{@{}l@{}c@{\hspace{0.5mm}}c@{\hspace{1.5mm}}c@{\hspace{0.5mm}}c@{\hspace{0.5mm}}c@{\hspace{0.5mm}}}
    % \rotatebox{90}{{}}&
    %  \includegraphics[trim={0 0 0 9cm},clip=true,width=0.15\linewidth]{figures/transfer2/000092_video_0_pose.png} & 
    %  \includegraphics[trim={0 0 0 9cm},clip=true,width=0.15\linewidth]{figures/transfer2/000092_video_0_drv.png} &
    %  \includegraphics[trim={0 0 0 3cm},clip=true,cfbox=bbox_color 9pt -9pt,width=0.15\linewidth]{figures/transfer2/frame_0091.png} & 
    
    %  \includegraphics[trim={0 0 0 9cm},clip=true,width=0.15\linewidth]{figures/transfer2/scene3_bo_000093_video_0_pose.png}&
    %  \includegraphics[trim={0 0 0 9cm},clip=true,width=0.15\linewidth]{figures/transfer2/scene3_bo_000093_video_0_drv.png}&
    %  \includegraphics[trim={0 1.5cm 0 1.5cm},clip=true,cfbox=bbox_color 9pt -9pt,width=0.15\linewidth]{figures/transfer2/scene3_bo_000093_video_0_out.png}
     
     \includegraphics[trim={0 0 0 9cm},clip=true,width=0.15\linewidth]{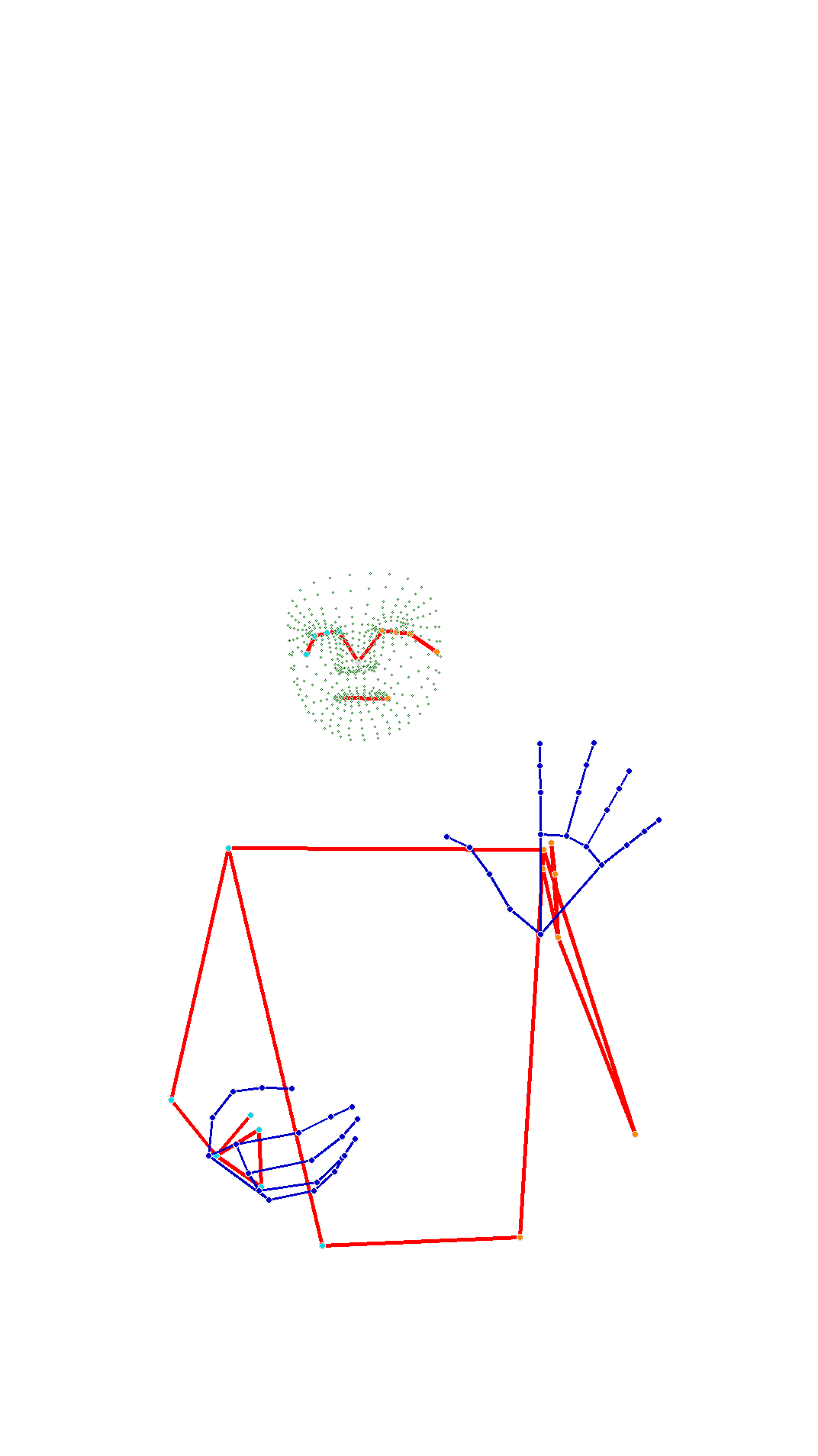}&
     \includegraphics[trim={0 0 0 9cm},clip=true,width=0.15\linewidth]{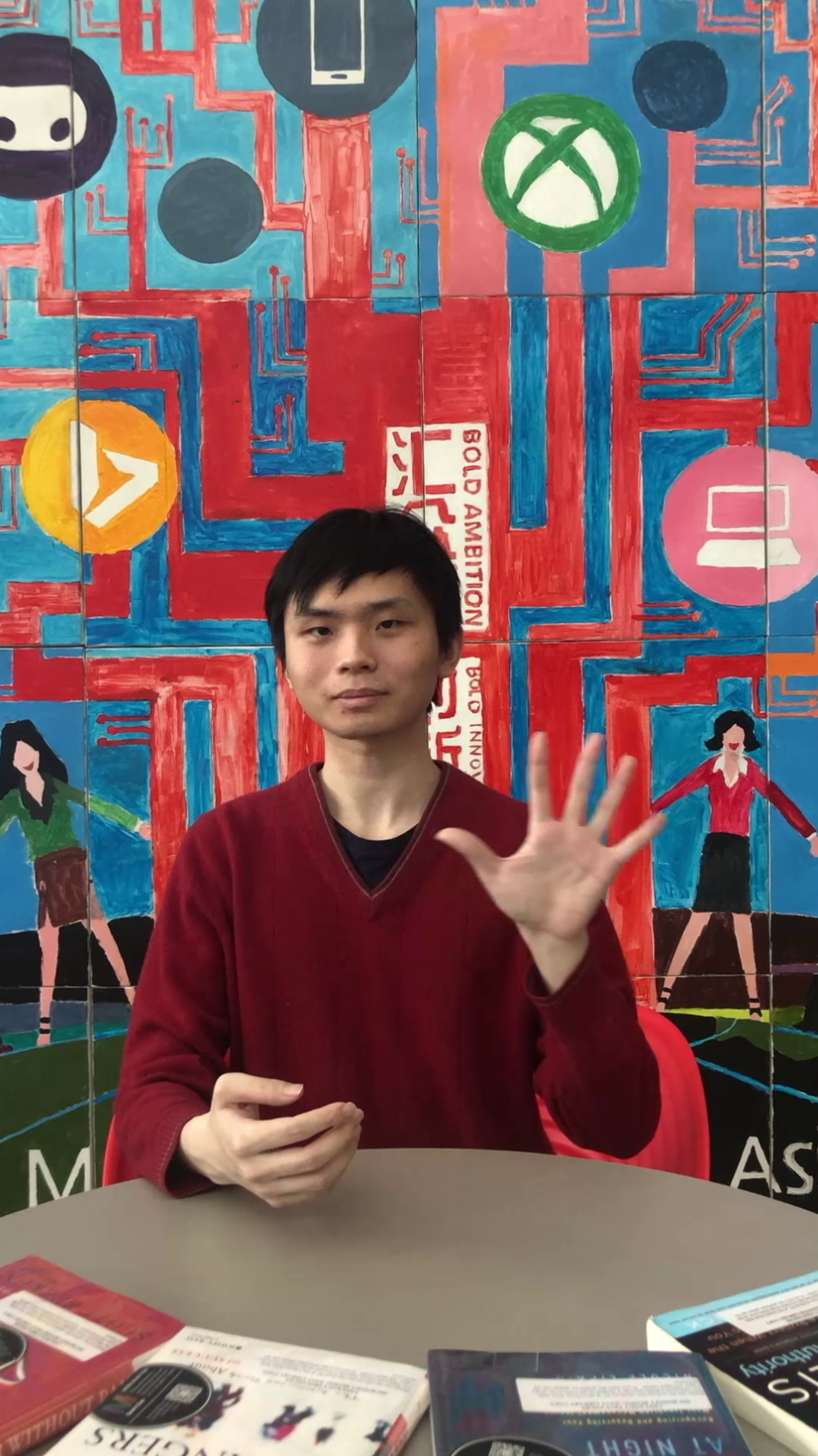}&
     \includegraphics[trim={0 0 0 3cm},clip=true,cfbox=bbox_color 9pt -9pt,width=0.15\linewidth]{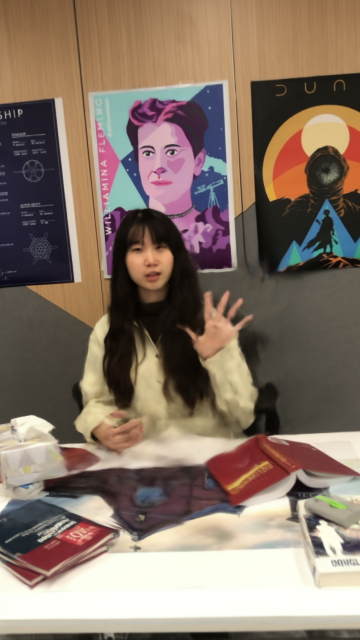} &
           \includegraphics[trim={0 0 0 9cm},clip=true,width=0.15\linewidth]{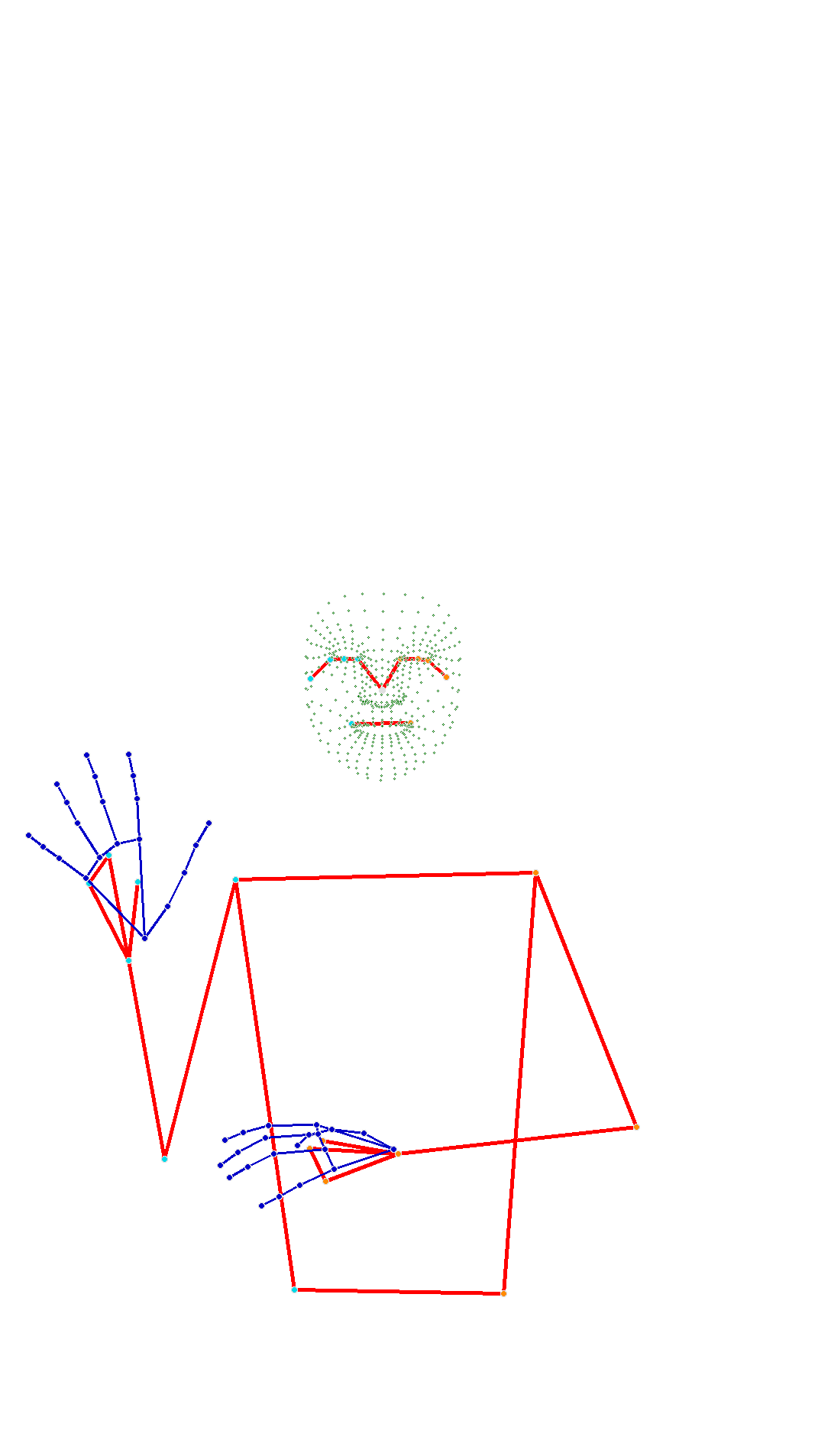}&
     \includegraphics[trim={0 0 0 9cm},clip=true,width=0.15\linewidth]{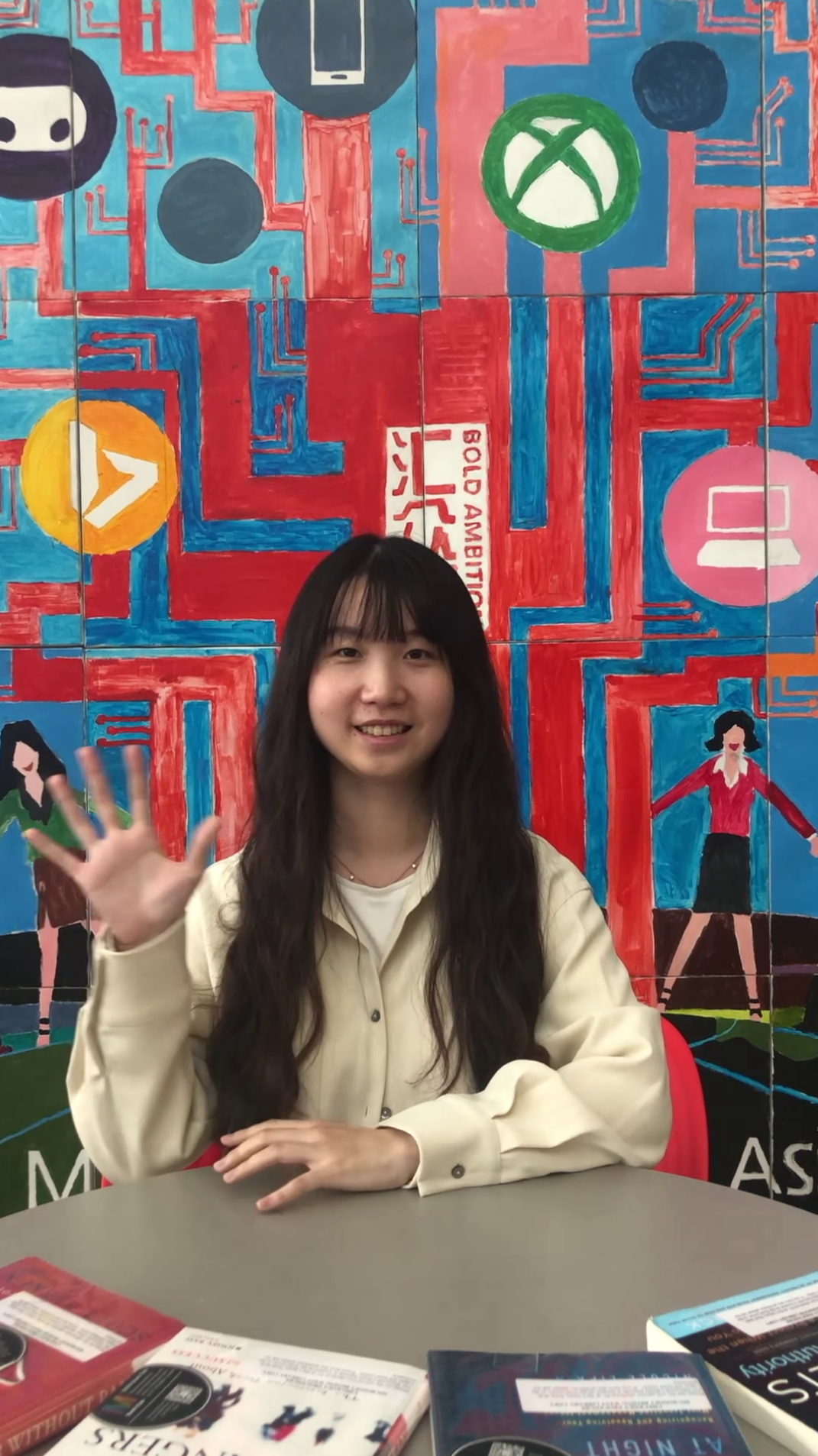}&
     \includegraphics[trim={0 1cm 0 2cm},clip=true,cfbox=bbox_color 9pt -9pt,width=0.15\linewidth]{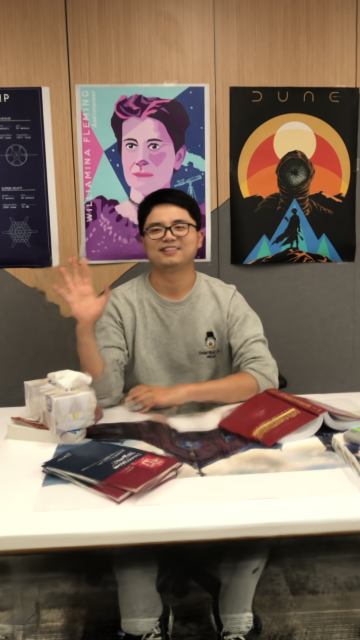}
    %  \\
    %  \includegraphics[trim={0 0 0 9cm},clip=true,width=0.15\linewidth]{figures/transfer2/junshu_0_000022_video_0_pose.png}&
    %  \includegraphics[trim={0 0 0 9cm},clip=true,width=0.15\linewidth]{figures/transfer2/junshu_0_000022_video_0_drv.png}&
    %  \includegraphics[trim={0 0 0 3cm},clip=true,cfbox=bbox_color 9pt -9pt,width=0.15\linewidth]{figures/transfer2/junshu_0_000022_video_0_out.png}&
    %  \includegraphics[trim={0 0 0 9cm},clip=true,width=0.15\linewidth]{figures/transfer2/junshu_0_000005_video_0_pose.png}&
    %  \includegraphics[trim={0 0 0 9cm},clip=true,width=0.15\linewidth]{figures/transfer2/junshu_0_000005_video_0_drv.png}&
    %  \includegraphics[trim={0 0 0 3cm},clip=true,cfbox=bbox_color 9pt -9pt,width=0.15\linewidth]{figures/transfer2/junshu_0_000005_video_0_out.png}
    
    %  \includegraphics[trim={0 1cm 0 2cm},clip=true,width=0.15\linewidth]{figures/transfer2/junshu_0_000099_video_0_out.png}
    %  \includegraphics[width=0.15\linewidth]{figures/transfer2/scene3_bo_000130_video_0_pose.png}&
    %  \includegraphics[width=0.15\linewidth]{figures/transfer2/scene3_bo_000130_video_0_drv.png}&
    %  \includegraphics[width=0.15\linewidth]{figures/transfer2/scene3_bo_000130_video_0_out.png}&
    %  \includegraphics[trim={0 0 0 9cm},clip=true,width=0.15\linewidth]{figures/transfer2/junshu_0_000099_video_0_pose.png}&
    %  \includegraphics[trim={0 0 0 9cm},clip=true,width=0.15\linewidth]{figures/transfer2/junshu_0_000099_video_0_drv.png}&
    %  \includegraphics[trim={0 1cm 0 2cm},clip=true,cfbox=bbox_color 9pt -9pt,width=0.15\linewidth]{figures/transfer2/junshu_0_000099_video_0_out.png}
    \end{tabular}\\
    \vspace{-0.2em}
    \caption{{Additional results on motion transfer.}}
    \vspace{1em}
    \label{fig:synchronized}
\end{figure*}
\end{document}